\pdfoutput=1
\documentclass[11pt]{article}

\usepackage[preprint]{acl}

\usepackage{times}
\usepackage{latexsym}

\usepackage[T1]{fontenc}
\usepackage[utf8]{inputenc}

\usepackage{microtype}

\usepackage{inconsolata}

\usepackage{graphicx}

\usepackage[table]{xcolor}
\usepackage[most]{tcolorbox}

\usepackage{lipsum}
\usepackage{geometry}

\usepackage{microtype}
\usepackage{url}
\usepackage{booktabs}

\usepackage{subfigure}
\usepackage{enumitem}
\newenvironment{itemize*}%
 {\leftmargini=20pt\begin{itemize}%
  \setlength{\itemsep}{3pt}%
  \setlength{\parskip}{0pt}%
  }%
 {\end{itemize}}
\newenvironment{enumerate*}%
 {\begin{enumerate}%
  \setlength{\itemsep}{0pt}%
  \setlength{\parskip}{0pt}}%
 {\end{enumerate}}

\usepackage{amsmath}
\usepackage{cleveref}
\usepackage{listings}
\usepackage{titlesec}
\usepackage{multirow}
\usepackage{makecell}

\usepackage{array}       % For custom column types (e.g., p{3.3cm})
\usepackage{caption}     % For better table captions
\usepackage{amssymb}
\usepackage{pdfpages}

\definecolor{lightred}{RGB}{255,163,163}
\definecolor{deepred}{RGB}{146,0,0}
\definecolor{midnightgreen}{rgb}{0.0, 0.29, 0.33}
\definecolor{deepgreen}{HTML}{0aa344}
\definecolor{deeppurple}{HTML}{7030a0}
\definecolor{deepblue}{HTML}{171d91}
\definecolor{brown}{HTML}{843c0c}
\definecolor{shadered}{HTML}{ffe5e5}
\definecolor{shadegreen}{HTML}{e5f7ed}
\definecolor{teal}{HTML}{008080}
\definecolor{brown}{HTML}{8b4513}

\definecolor{skill_green}{HTML}{5c8d40}
\definecolor{skill_red}{HTML}{b02418}
\definecolor{skill_purple}{HTML}{6e276b}
\definecolor{skill_blue}{HTML}{4fadea}
\definecolor{skill_orange}{HTML}{da7842}

\newcommand{\deepred}{\textcolor{brown}}
\newcommand{\deepgreen}{\textcolor{teal}}

\usepackage{pifont}
\usepackage[normalem]{ulem}        % For text formatting

\tcbuselibrary{listings, breakable, skins}

\newtcblisting{promptbox}[1][]{
  enhanced,
  breakable,
  listing only,
  boxrule=0.3mm,
  arc=3mm,
  auto outer arc,
  colback=gray!5!white,     % default background, can override
  colframe=blue!75!black,   % default border, can override
  fonttitle=\small\bfseries, % smaller title font
  #1,                       % allow user overrides (title, colback, colframe, ...)
  listing options={
    basicstyle=\ttfamily\scriptsize,
    breaklines=true,
    breakautoindent=false,  % <-- no auto indent for wrapped lines
    breakindent=0pt,        % <-- explicitly zero extra indent
    columns=fullflexible,
    keepspaces=true,
    showstringspaces=false
  }
}
\usepackage{amsmath, amssymb, amsfonts}

\usepackage{algorithm}
\usepackage{algpseudocode}

\title{Current Agents Fail to Leverage World Model as Tool for Foresight}

\author{
Cheng Qian$^{1}$, Emre Can Acikgoz$^{1}$, Bingxuan Li$^{1}$, Xiusi Chen$^{1}$, Yuji Zhang$^{1}$, Bingxiang He$^{2}$,\\
\textbf{Qinyu Luo$^{3}$, Dilek Hakkani-Tür$^{1}$, Gokhan Tur$^{1}$, Yunzhu Li$^{4}$, Heng Ji$^{1}$}\vspace{1.5mm}\\
$^{1}$UIUC, $^{2}$THU, $^{3}$JHU, $^{4}$Columbia\vspace{1mm}\\
\texttt{\{chengq9, hengji\}@illinois.edu}\\
}

\begin{document}
\maketitle
\begin{abstract}
Agents built on vision-language models increasingly face tasks that demand anticipating future states rather than relying on short-horizon reasoning. Generative world models offer a promising remedy: agents could use them as external simulators to foresee outcomes before acting. This paper empirically examines whether current agents can leverage such world models as tools to enhance their cognition. Across diverse agentic and visual question answering tasks, we observe that some agents rarely invoke simulation (fewer than 1\%), frequently misuse predicted rollouts (approximately 15\%), and often exhibit inconsistent or even degraded performance (up to 5\%) when simulation is available or enforced. Attribution analysis further indicates that the primary bottleneck lies in the agents’ capacity to decide when to simulate, how to interpret predicted outcomes, and how to integrate foresight into downstream reasoning. These findings underscore the need for mechanisms that foster calibrated, strategic interaction with world models, paving the way toward more reliable anticipatory cognition in future agent systems.
\end{abstract}

\section{Introduction}
Modern AI agents are increasingly expected to operate in settings where tasks unfold over long horizons, involve intricate chains of interdependent decisions, and may yield irreversible consequences. Such scenarios, ranging from multi-stage robotics manipulation \citep{huang2023voxposer, huang2024rekep, Feng2025ReflectivePlanning, Fan2025LongVLA} to complex software automation \citep{Deng2025SWEBenchPro, Wang2025RepoMaster, Agashe2025AgentS2} and real-world planning \citep{Garg2025REAL, Erdogan2025PlanAndAct, Geng2025REALMBench}, demand that agents not only reason locally but also anticipate how the environment might evolve.
Human decision-making in similar contexts is often anchored in our ability to project possible futures states, assess risks, and adjust actions accordingly~\citep{gilbert2007prospection}. Classical theories of cognition, such as mental simulation and prospective reasoning in psychology and cognitive science, emphasize that people routinely rely on internal models of how the world behaves to forecast outcomes before acting~\citep{premack1978tom_chimpanzee, johnson1983mental, schacter2007RememberingTP}. These enable us to navigate uncertainty, manage delayed rewards, and avoid catastrophic errors.

Large language model (LLM) agents, despite impressive proficiency in planning, decomposition, and interacting with diverse tools or environments, still exhibit a major limitation: they often lack a robust mechanism for foresight~\citep{chae2024webagentsworldmodels, jin2024marple}. Even when an LLM can articulate a goal-achieving plan, this does not necessarily imply that it can anticipate how future states of the environment might unfold.\footnote{As a form of reasoning, \textit{Foresight} in our scope refers to an agent’s ability to anticipate how the environment will evolve, rather than to generate plans to achieve goals from its own perspective.} For instance, a household robot may plan to open a window to vent smoke, but fail to anticipate that incoming wind will blow loose papers into a spill, spreading contamination and creating new hazards. Such error illustrates that effective action depends not only on a plausible plan, but on foresight about how the environment will evolve under interventions, especially when those dynamics introduce delayed, non-local constraints. Take a step further from this perspective, effective foresight is not a single capability but a governed process: agents must decide what future to query, how to interpret predicted outcomes, and when to act upon them. Failures at any of these stages can negate the benefits of accurate simulation.

\begin{figure}[!t]
    \centering
    \includegraphics[width=\linewidth]{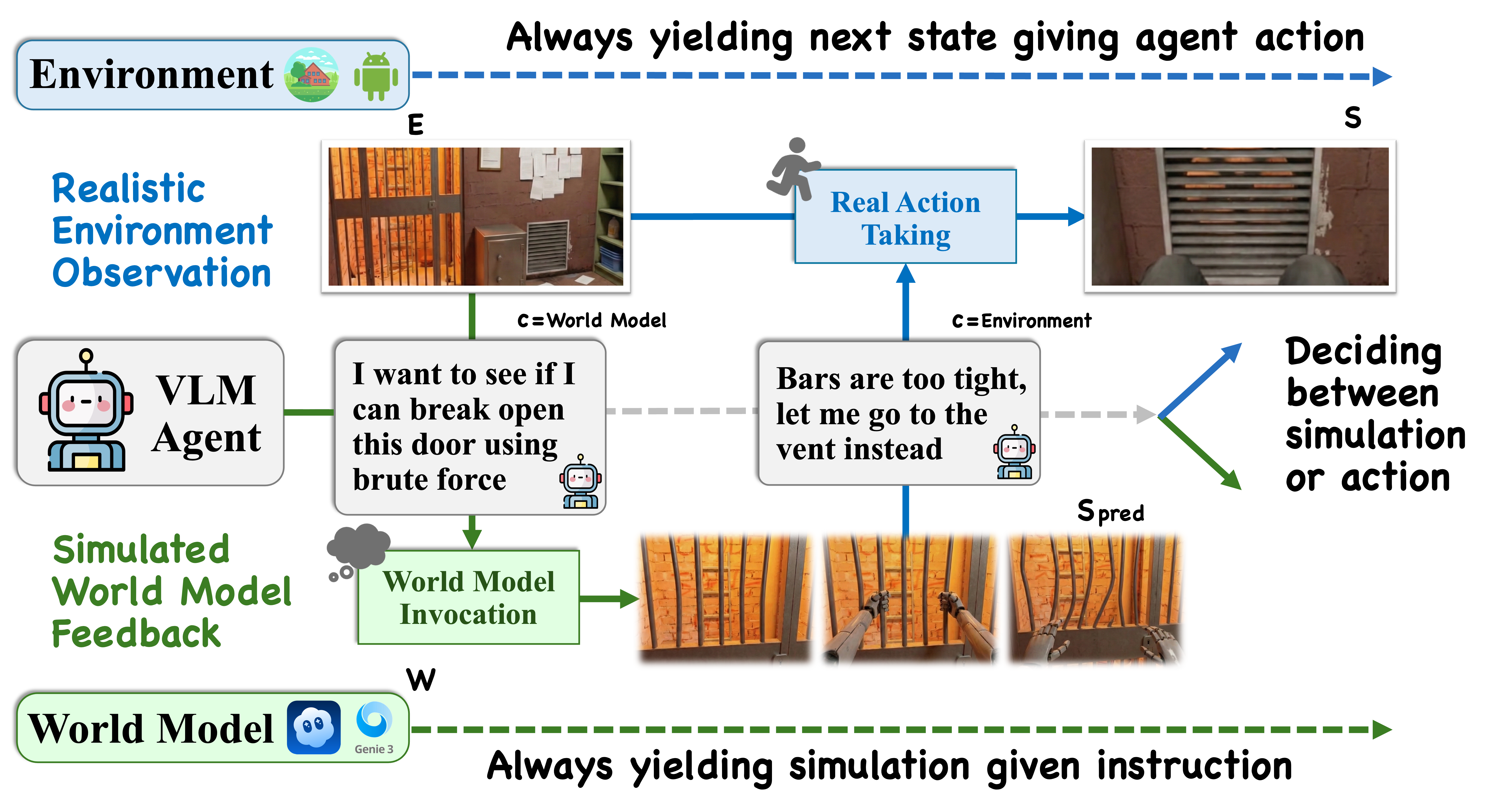}
    \caption{The world model as tool framework where the agent decides between real action taking and simulation.}
    \label{fig:framework}
    \vspace{-8pt}
\end{figure}

Recent research has begun to explore whether foresight can be explicitly supported by equipping agents with mechanisms for simulating the future. One line of work takes an training-free approach, augmenting agents with dedicated foresight modules through prompts~\citep{qian2024escapebench} or encouraging deliberate world-modeling via structured instructions~\citep{wang2025vagen}. Another line of research focuses on intrinsic solutions by training or embedding learned world models directly into the agent itself, as seen in efforts such as code world model or multimodal generative simulators~\citep{carbonneaux2025cwm}. However, simple prompting approaches often behave rigidly, failing to capture the visually diverse world states, while intrinsic methods typically require extensive retraining, substantial computational resources, and are difficult to integrate reliably across model families.

At the same time, world models---especially large-scale video and environment simulators\footnote{The terms \textit{simulator} and \textit{world model} are used interchangeably in this paper. They both yield the prediction about future environment state given an agent action.} such as Sora, WAN, and other generative dynamics models~\citep{videoworldsimulators2024, wan2025wan}---have advanced to produce reasonably coherent and temporally consistent predictions in open-ended settings. This raises a natural question: instead of building foresight inside the agent or enforcing highly structured prompting, can we \textbf{treat world models as off-the-shelf tools that an agent may optionally call upon when needed}? Doing so reframes foresight not as a fixed architectural component but as a resource that an agent might strategically leverage. This perspective opens an important direction: \emph{To what extent can current LLM agents decide when to use world models for environmental foresight, and does this foresight substantially improve their cognition?}

Our preliminary investigation reveals two concrete limitations. First, many LLM agents are reluctant to invoke world-model tools even when it improves environmental foresight, and this reluctance increases with model scale or capability. The behavior is also family-consistent: LLaMA variants show comparatively higher willingness, while GPT and Qwen variants frequently decline tool use across sizes. Second, even when world models are invoked, cognitive benefits are inconsistent because agents often misuse the tool, for example by producing only one deterministic future, overriding simulations with over-confident internal reasoning, or failing to evaluate counterfactual branches. 

Taken together, these findings underscore that progress in foresight-augmented agents is more about calibrating its behavior: models should more strategically consult external simulations when benefits are clear, less prone to over-trust inaccurate internal reasoning, and better able to incorporate simulated evidence without either dismissing it or rigidly overfitting to it. Our analysis also reveals that these calibration issues follow systematic patterns. We identify three recurrent governance breakdowns, including misguided input formulation, ambiguous interpretation of simulated outcomes, and unstable action integration, which collectively explain the majority of observed regressions. To summarize:
\begin{itemize}[topsep=2pt, partopsep=-5pt, leftmargin=8pt, itemsep=-4pt]
    \item We introduce the conceptual framing of \emph{world models as tools}, offering an analysis-focused perspective that examines how existing agents interact with external simulators.
    \item We propose an evaluation framework that enables systematic study of agents' willingness, correctness, and consistency when employing world models, providing a basis for future training.
    \item Through extensive analysis, we uncover behavioral patterns, family-specific tendencies, error taxonomy, and structural weaknesses that inform future research on agentic foresight.
\end{itemize}

Overall, we hope this study deepens our understanding of how agentic models reason about the future, how they engage with external simulation tools, and what obstacles remain before such systems can exhibit reliable anticipatory cognition. We foresee future agents as systems that can strategically invoke rich world simulations and integrate them with internal reasoning, ultimately moving toward more grounded, self-aware, and safety-aligned decision-making.

\section{Related Work}
Cognitive science argues that humans plan by mentally simulating actions and outcomes using internal world models, ranging from \emph{theory of mind} (simulating another person’s inner world to anticipate their reaction) to broader expectations about how an environment’s state will change; analogously, recent work equips agents with world modeling machinery for foresight, either by treating an LLM or VLM as a dynamics model and searching over imagined trajectories, or by embedding a learned simulator in the decision loop to mitigate short-sighted action selection~\cite{deng2025simura, hao2023reasoning}. This idea extends beyond text: in code, agents can anticipate program outcomes internally before committing to an answer~\cite{carbonneaux2025cwm}; in vision, agents benefit from explicitly separating state estimation from action-conditioned transition prediction, often by importing dynamics priors from video world models~\cite{wang2025vagen, zhang2025can} or future frame prediction~\cite{ebm2025}. In parallel, scaling generative video models yields increasingly realistic long-horizon simulators~\cite{brooks2024video}, and lightweight action conditioning further turns them into controllable rollouts for planning and reinforcement learning~\cite{he2025pre}; recent efforts also emphasize evaluating such world models by \emph{closed-loop} embodied utility rather than open-loop visual fidelity, e.g., via standardized online-planning interfaces and task-success benchmarks~\citep{zhang2025world}. Complementary evidence also shows that language itself can function as a simulator when agents are prompted to hypothesize and reflect~\cite{qian2024escapebench,Mobile-Agent-E2025}, and that closing the loop with external physics engines can enforce feasibility constraints in embodied settings~\cite{meng2025magic}. In contrast to work that primarily proposes new simulators or integration architectures, our work focuses on the agent-side question of \emph{when} to invoke world models and \emph{how} simulated futures are interpreted and used during downstream reasoning.

\section{Preliminaries}

We study an evaluation protocol where a decision-making agent may explicitly invoke a learned world model as a tool for planning. The key premise is simple: instead of acting blindly in the real environment, the agent can optionally simulate candidate actions in a predictive model before committing to them. This allows us to measure not only whether the agent can reach a goal, but whether it uses simulation to improve the quality of its decisions.

At time step $t$, the agent has already executed a sequence of actions and received corresponding observations, which we denote as its trajectory $T_t$. The real environment is in state $S_t$, and the agent's objective is to reach a task-specific goal state $G$ through interactions with the real environment $E$. The world model $W$ approximates the transition dynamics of $E$ but runs independently without affecting the real world.

We illustrate the protocol with a concrete room escape scenario in \Cref{fig:framework}. An embodied agent is trapped in a room and evaluates two irreversible options: breaking a reinforced metal door or crawling through a narrow air vent. Vision alone cannot confirm feasibility, so the agent simulates both actions using a world model. Simulation reveals that the door is too reinforced to break under realistic force limits, while the vent leads to a valid exit path. With this feedback, the agent commits to the vent-crawl action in the real environment and escapes. The example highlights the role of the world model as a tool for de-risking high-stakes decisions before acting irreversibly.

\subsection{Interaction Process}
Formally, the agent’s policy maps its current trajectory to an action $a_t$ and a choice of execution context $c_t \in \{E, W\}$:

\begin{equation}
    (c_t, a_t) = A(T_t).
\end{equation}

If the agent selects $c_t = W$, it performs a simulation step. The model predicts the next state without altering reality:

\begin{equation}
    S_{\text{pred}} = W(S_t, a_t).
\end{equation}

The simulated outcome is appended to the trajectory as a hypothetical observation, enabling future reasoning:

\begin{equation}
    T_{t+1} \leftarrow T_t \cup \{(a_t, S_{\text{pred}})\}.
\end{equation}

If the agent instead selects $c_t = E$, the action is executed in the real environment with true dynamics:

\begin{equation}
    S_{t+1} \leftarrow E(S_t, a_t),
\end{equation}

and the real outcome is recorded in the same trajectory container:

\begin{equation}
    T_{t+1} \leftarrow T_t \cup \{(a_t, S_{t+1})\}.
\end{equation}

The process continues until either the agent reaches the goal ($S_t = G$) or a maximum horizon $L$ is exceeded. Because both real and simulated outcomes are stored in the same trajectory object, we can evaluate how world model feedback influences final decisions, how often simulation is used, and how reliably it prevents costly or impossible real-world attempts.

\begin{algorithm}[t]
\caption{Agent Interaction Process}
\label{alg:interaction}
\begin{algorithmic}[1]
\Require Agent $A$, real environment $E$, world model $W$, initial real state $S_1$, horizon $L$, goal predicate $G(\cdot)$
\State $T_1 \gets \emptyset$ \Comment{Trajectory: records both real and simulated outcomes}
\For{$t = 1,2,\dots,L$}
    \State $(c_t, a_t) \gets A(T_t)$ \Comment{$c_t \in \{E,W\}$ chooses where to apply $a_t$}
    \If{$c_t = W$} \Comment{Simulate; do not change the real state}
        \State $S^{\text{pred}}_{t+1} \gets W(S_t, a_t)$
        \State $T_{t+1} \gets T_t \cup \{(c_t, a_t, S^{\text{pred}}_{t+1})\}$
        \State $S_{t+1} \gets S_t$ \Comment{Real state remains unchanged}
    \Else \Comment{$c_t = E$: execute in the real environment}
        \State $S_{t+1} \gets E(S_t, a_t)$
        \State $T_{t+1} \gets T_t \cup \{(c_t, a_t, S_{t+1})\}$
    \EndIf
    \If{$G(S_{t+1})$}
        \State \textbf{break}
    \EndIf
\EndFor
\State \Return $T_{t+1}$
\end{algorithmic}
\end{algorithm}

\subsection{Interaction Modes}
A key aspect of the framework is how the agent is informed about the world model and allowed to interact with it. We control this through the system prompt, which leads to three distinct modes:
\begin{itemize}[topsep=3pt, partopsep=0pt, leftmargin=8pt, itemsep=0pt]
    \item \textbf{Normal Mode.} The agent is informed that a world model $W$ exists and may choose to query it or directly act in $E$. This mode captures realistic decision-making in which simulation is an optional tool whose value must be inferred and used appropriately.
    \item \textbf{WM Invisible Mode.} The agent is agnostic of $W$, and therefore never queries it. All actions are executed directly in $E$, serving as the standard baseline consistent with prior evaluations.
    \item \textbf{WM Force Mode.} The agent is instructed to simulate actions in $W$ before executing them in $E$, making world model use mandatory. This mode reveals whether compulsory planning helps or hinders performance when simulation is no longer a strategic choice.
\end{itemize}
In our experiments, we examine all three, with particular focus on comparing the first two modes, which isolates the effect of optional access to a world model while keeping all other factors fixed.

\paragraph{Implications.}
By embedding world model usage directly into the decision process, we evaluate not only whether an agent can solve tasks, but whether it can \emph{use foresight intelligently}. The agent must decide when simulation is worthwhile, interpret predictive outcomes from $W$, and combine them with real observations from $E$ to guide future actions. Performance therefore reflects the agent's ability to integrate simulated and real experience, rather than simply benefiting from an external model. This allows us to study the gap between having powerful predictive tools and being able to exploit them effectively in interaction-heavy environments.

% \section{Framework}

\section{Experiments}

\subsection{Task Choices}
To understand when and how world models benefit VLM agents, we evaluate two complementary types of tasks that require VLM-based reasoning.

The first type is \textbf{agentic decision-making tasks}, which require visual grounding, perception-to-action mapping, and multi-step reasoning in interactive environments. Following VAGEN~\citep{wang2025vagen}, we select four diverse tasks: \emph{FrozenLake}, \emph{Navigation}, \emph{PrimitiveSkill}, and \emph{Sokoban}. These tasks jointly cover: (i) long-horizon interaction, (ii) reasoning in both 2D symbolic layouts and photorealistic 3D scenes, and (iii) action planning across different embodiments, including household agents, robot manipulation, and game environments. These properties make them ideal for assessing whether simulated rollouts can meaningfully support agent's cognition.

The second type is pure \textbf{VLM reasoning tasks}, where no embodied control is required. Our aim is to test the simulation paradigm's impact when the agent interacts only through perceptual or spatial reasoning without action execution. We include four representative datasets: \emph{3DSRBench}~\citep{ma20253dsrbench}, \emph{MMSI Bench}~\citep{yang2025mmsi}, \emph{SAT}~\citep{ray2024sat}, and \emph{Spatial-MM Object}~\citep{shiri2024empirical}. These benchmarks are chosen because: (i) they demand rich spatial reasoning and visual grounding, and (ii) their questions contain latent dynamics or hypothetical transformations that can theoretically benefit from simulated outcomes. Without this property, evaluating world model augmentation would be inherently meaningless, since simulation would not provide additional utility beyond static perception.

\begin{table*}[t]
    \centering
    \setlength\tabcolsep{2pt}
    \setlength\extrarowheight{2pt}
    
    \tabcolsep=0.01\linewidth
    \resizebox{\linewidth}{!}{
    \begin{tabular}{l c c c c}
        \toprule
        \textbf{Tasks} & \textbf{Goal Condition \(G\)} & \textbf{Environment State \(S\)} & \textbf{Action \(a\) in Environment} & \textbf{Action \(a\) in World Model} \\
        \midrule
        \textbf{FrozenLake} & \makecell[c]{Reach the target grid without falling} & \makecell[c]{2D grid observation of agent position} & \multicolumn{2}{c}{\makecell[c]{Discrete movement (up, down, left, right)}} \\
        \textbf{Navigation} & \makecell[c]{Arrive at a target location} & \makecell[c]{First-person RGB scene} & \multicolumn{2}{c}{\makecell[c]{Embodied moves (forward, rotate, look, etc.)}} \\
        \textbf{PrimitiveSkill} & \makecell[c]{Manipulate objects to a target pose} & \makecell[c]{Arm pose + object pose} & \multicolumn{2}{c}{\makecell[c]{Manipulation command (pick, place, move with coordinates)}} \\
        \textbf{Sokoban} & \makecell[c]{Push boxes to goal cells} & \makecell[c]{2D game state with box locations} & \multicolumn{2}{c}{\makecell[c]{Discrete movement (up, down, left, right)}} \\
        \midrule
        \textbf{All VQA Tasks} & \makecell[c]{Output the correct answer} & \makecell[c]{Provided problem images} & \makecell[c]{Direct textual answer} & \makecell[c]{Simulation-oriented query on an image} \\
        \bottomrule
    \end{tabular}
    }
    \caption{Task families and how world model queries integrate with the agent–environment loop.}
    \label{tab:dataset_interaction_illustration}
\end{table*}

Together, these two categories allow us to examine world model usage both in interactive control and in non-embodied visual reasoning, revealing whether simulation is effective or not. \Cref{tab:dataset_interaction_illustration} summarizes the task families and how world model queries fit into the agent–environment loop.

\subsection{Experiment Settings}

\paragraph{Test Models.} 
We evaluate nine vision-language models across both open- and closed-source families, including GPT~\citep{openai2025gpt5system}, Llama~\citep{meta2025llama4}, and Qwen~\citep{bai2025qwen2}. For each family, we select multiple model sizes to investigate how world model augmentation affects models with different capacities.
All tested models possess basic visual reasoning abilities so that potential improvements or degradation can be attributed to world model usage rather than insufficient perception.

\paragraph{World Model.} 
For agentic tasks, we construct the world model by cloning the current environment state and simulating the effects of hypothetical actions directly in the cloned environment. This setup provides accurate simulated states without altering the real environment, which is sufficient for our purpose since the focus of this work is not the fidelity of world model learning, but rather whether agents can \emph{use} a world model effectively.  
For VQA reasoning tasks, simulation requires perceptual imagination rather than state transitions, so we use Wan2.1~\citep{wan2025wan} as the world model to generate visual states based on textual simulation instructions given by the agent. Full configurations are provided in \Cref{apdx:experiment_settings}.

\paragraph{Modes and Metrics.} 
Our main evaluations compare \emph{Normal Mode} (world model optional) with \emph{World Model Invisible Mode} (world model unavailable), providing a clean measure of the value of optional simulation. \emph{World Model Force Mode} is additionally applied to agentic tasks to explore whether mandatory simulation helps or harms planning.  
Across all agentic tasks, the performance is measured by the final task's success rate, consistent with prior work. In VQA tasks, we evaluate answer correctness on multiple-choice questions, as simulation here theoretically aids perceptual reasoning rather than sequential control.

%%%%%%%%%%%%%%%%%%%%%%%%%%%%%%%%%%%%%%%%%%%%%%%%%%%%%%
%%%%%%%%%%%%%%%% Results and Analysis %%%%%%%%%%%%%%%%
%%%%%%%%%%%%%%%%%%%%%%%%%%%%%%%%%%%%%%%%%%%%%%%%%%%%%%

\subsection{Experiment Results}

\begin{table}[t]
    \centering
    \setlength\tabcolsep{2pt}
    \setlength\extrarowheight{2pt}
    
    \tabcolsep=0.01\linewidth
    \resizebox{\linewidth}{!}{
    \begin{tabular}{l c c c c c}
        \toprule
        \textbf{Model} & \textbf{\makecell{Frozen\\Lake}} & \textbf{Navigate} & \textbf{\makecell{Primitive\\Skill}} & \textbf{Sokoban} & \textbf{Avg.} \\
        \addlinespace[2pt]
        \midrule
        \addlinespace[2pt]
        \multicolumn{6}{c}{\cellcolor[HTML]{EFEFEF} \textit{Without World Model Access}} \\
        \midrule
        GPT-4o-mini & 0.36 & 0.26 & 0.32 & 0.00 & 0.27 \\
        GPT-4o & 0.58 & 0.35 & 0.51 & 0.02 & 0.40 \\
        GPT-5-mini & 0.86 & 0.66 & 0.11 & 0.03 & 0.41 \\
        GPT-5 & 0.77 & 0.74 & 0.19 & 0.06 & 0.47 \\
        Llama-4-Maverick & 0.70 & 0.31 & 0.40 & 0.00 & 0.35 \\
        Llama-4-Scout & 0.59 & 0.55 & 0.32 & 0.02 & 0.42 \\
        Qwen2.5-VL-7B & 0.36 & 0.26 & 0.13 & 0.02 & 0.20 \\
        Qwen2.5-VL-32B & 0.59 & 0.36 & 0.49 & 0.00 & 0.40 \\
        Qwen2.5-VL-72B & 0.61 & 0.38 & 0.41 & 0.00 & 0.37 \\
        \addlinespace[2pt]
        \midrule
        \addlinespace[2pt]
        \multicolumn{6}{c}{\cellcolor[HTML]{EFEFEF} \textit{With World Model Access}} \\
        \midrule
        GPT-4o-mini & 0.39\deepgreen{$_{\uparrow0.03}$} & 0.24\deepred{$_{\downarrow0.02}$} & 0.22\deepred{$_{\downarrow0.10}$} & 0.00 & 0.22\deepred{$_{\downarrow0.05}$} \\
        GPT-4o & 0.58 & 0.31\deepred{$_{\downarrow0.04}$} & 0.46\deepred{$_{\downarrow0.05}$} & 0.02 & 0.36\deepred{$_{\downarrow0.04}$} \\
        GPT-5-mini & 0.89\deepgreen{$_{\uparrow0.03}$} & 0.63\deepred{$_{\downarrow0.03}$} & 0.19\deepgreen{$_{\uparrow0.08}$} & 0.00\deepred{$_{\downarrow0.03}$} & 0.43\deepgreen{$_{\uparrow0.02}$} \\
        GPT-5 & 0.89\deepgreen{$_{\uparrow0.12}$} & 0.71\deepred{$_{\downarrow0.03}$} & 0.20\deepgreen{$_{\uparrow0.01}$} & 0.14\deepgreen{$_{\uparrow0.08}$} & 0.48\deepgreen{$_{\uparrow0.01}$} \\
        Llama-4-Maverick & 0.66\deepred{$_{\downarrow0.04}$} & 0.20\deepred{$_{\downarrow0.11}$} & 0.32\deepred{$_{\downarrow0.08}$} & 0.03\deepgreen{$_{\uparrow0.03}$} & 0.27\deepred{$_{\downarrow0.08}$} \\
        Llama-4-Scout & 0.55\deepred{$_{\downarrow0.04}$} & 0.54\deepred{$_{\downarrow0.01}$} & 0.24\deepred{$_{\downarrow0.08}$} & 0.03\deepgreen{$_{\uparrow0.01}$} & 0.38\deepred{$_{\downarrow0.04}$} \\
        Qwen2.5-VL-7B & 0.45\deepgreen{$_{\uparrow0.09}$} & 0.26 & 0.12\deepred{$_{\downarrow0.01}$} & 0.00\deepred{$_{\downarrow0.02}$} & 0.20 \\
        Qwen2.5-VL-32B & 0.58\deepred{$_{\downarrow0.01}$} & 0.32\deepred{$_{\downarrow0.04}$} & 0.39\deepred{$_{\downarrow0.10}$} & 0.02\deepgreen{$_{\uparrow0.02}$} & 0.34\deepred{$_{\downarrow0.06}$} \\
        Qwen2.5-VL-72B & 0.45\deepred{$_{\downarrow0.16}$} & 0.37\deepred{$_{\downarrow0.01}$} & 0.32\deepred{$_{\downarrow0.09}$} & 0.00 & 0.33\deepred{$_{\downarrow0.04}$} \\
        \bottomrule

    \end{tabular}
    }
    \caption{Agent Task Success Rate: Comparison between w/wo WM access. Almost all the models in all the agent tasks fail to reach higher performance with WM access.}
    \label{tab:agent_success_rate}
\end{table}
\begin{table}[t]
    \centering
    \setlength\tabcolsep{2pt}
    \setlength\extrarowheight{2pt}
    
    \tabcolsep=0.018\linewidth
    \resizebox{\linewidth}{!}{
    \begin{tabular}{l c c c c c}
        \toprule
        \textbf{Model} & \textbf{3DSRBench} & \textbf{MMSI} & \textbf{SAT} & \textbf{Spatial} & \textbf{Avg.} \\
        \addlinespace[2pt]
        \midrule
        \addlinespace[2pt]
        \multicolumn{6}{c}{\cellcolor[HTML]{EFEFEF} \textit{Without World Model Access}} \\
        \midrule
        GPT-4o-mini & 0.58 & 0.28 & 0.52 & 0.65 & 0.56 \\
        GPT-4o & 0.66 & 0.31 & 0.71 & 0.72 & 0.63 \\
        GPT-5-mini & 0.67 & 0.35 & 0.85 & 0.78 & 0.66 \\
        GPT-5 & 0.69 & 0.38 & 0.86 & 0.80 & 0.68 \\
        Llama-4-Maverick & 0.61 & 0.27 & 0.52 & 0.74 & 0.60 \\
        Llama-4-Scout & 0.59 & 0.27 & 0.41 & 0.74 & 0.58 \\
        Qwen2.5-VL-7B & 0.53 & 0.24 & 0.59 & 0.62 & 0.52 \\
        Qwen2.5-VL-32B & 0.59 & 0.30 & 0.47 & 0.67 & 0.57 \\
        Qwen2.5-VL-72B & 0.61 & 0.29 & 0.47 & 0.71 & 0.59 \\
        % \addlinespace[2pt]
        % \midrule
        % \addlinespace[2pt]
        % \multicolumn{6}{c}{\cellcolor[HTML]{EFEFEF} \textit{With World Model Access}} \\
        % \midrule
        % GPT-4o-mini & 0.59 & 0.27 & 0.57 & 0.66 & 0.56 \\
        % GPT-4o & 0.66 & 0.30 & 0.73 & 0.72 & 0.63 \\
        % GPT-5-mini & 0.68 & 0.36 & 0.83 & 0.79 & 0.67 \\
        % GPT-5 & 0.70 & 0.37 & 0.85 & 0.79 & 0.68 \\
        % Llama-4-Maverick & 0.62 & 0.28 & 0.47 & 0.75 & 0.60 \\
        % Llama-4-Scout & 0.59 & 0.28 & 0.36 & 0.73 & 0.58 \\
        % Qwen2.5-VL-7B & 0.54 & 0.24 & 0.66 & 0.63 & 0.52 \\
        % Qwen2.5-VL-32B & 0.58 & 0.28 & 0.47 & 0.67 & 0.56 \\
        % Qwen2.5-VL-72B & 0.61 & 0.29 & 0.48 & 0.73 & 0.59 \\
        % \bottomrule
        \addlinespace[2pt]
        \midrule
        \addlinespace[2pt]
        \multicolumn{6}{c}{\cellcolor[HTML]{EFEFEF} \textit{With World Model Access}} \\
        \midrule
        GPT-4o-mini & 0.59\deepgreen{$_{\uparrow0.01}$} & 0.27\deepred{$_{\downarrow0.01}$} & 0.57\deepgreen{$_{\uparrow0.05}$} & 0.66\deepgreen{$_{\uparrow0.01}$} & 0.56 \\
        GPT-4o & 0.66 & 0.30\deepred{$_{\downarrow0.01}$} & 0.73\deepgreen{$_{\uparrow0.02}$} & 0.72 & 0.63 \\
        GPT-5-mini & 0.68\deepgreen{$_{\uparrow0.01}$} & 0.36\deepgreen{$_{\uparrow0.01}$} & 0.83\deepred{$_{\downarrow0.02}$} & 0.79\deepgreen{$_{\uparrow0.01}$} & 0.67\deepgreen{$_{\uparrow0.01}$} \\
        GPT-5 & 0.70\deepgreen{$_{\uparrow0.01}$} & 0.37\deepred{$_{\downarrow0.01}$} & 0.85\deepred{$_{\downarrow0.01}$} & 0.79\deepred{$_{\downarrow0.01}$} & 0.68 \\
        Llama-4-Maverick & 0.62\deepgreen{$_{\uparrow0.01}$} & 0.28\deepgreen{$_{\uparrow0.01}$} & 0.47\deepred{$_{\downarrow0.05}$} & 0.75\deepgreen{$_{\uparrow0.01}$} & 0.60 \\
        Llama-4-Scout & 0.59 & 0.28\deepgreen{$_{\uparrow0.01}$} & 0.36\deepred{$_{\downarrow0.05}$} & 0.73\deepred{$_{\downarrow0.01}$} & 0.58 \\
        Qwen2.5-VL-7B & 0.54\deepgreen{$_{\uparrow0.01}$} & 0.24 & 0.66\deepgreen{$_{\uparrow0.07}$} & 0.63\deepgreen{$_{\uparrow0.01}$} & 0.52 \\
        Qwen2.5-VL-32B & 0.58\deepred{$_{\downarrow0.01}$} & 0.28\deepred{$_{\downarrow0.02}$} & 0.47 & 0.67 & 0.56\deepred{$_{\downarrow0.01}$} \\
        Qwen2.5-VL-72B & 0.61 & 0.29 & 0.48\deepgreen{$_{\uparrow0.01}$} & 0.73\deepgreen{$_{\uparrow0.02}$} & 0.59 \\
        \bottomrule
    \end{tabular}
    }
    \vspace{-2mm}
    \caption{VQA Task Accuracy: Comparison between w/wo WM access across VQA benchmarks. All model performances are nearly the same for benchmarks.}
    \label{tab:vqa_accuracy}
    \vspace{-4mm}
\end{table}

\paragraph{Finding 1: World Models Do Not Reliably Improve Performance..} 
\Cref{tab:agent_success_rate} and \Cref{tab:vqa_accuracy} compare base models with their world model–augmented variants. Across both agent and VQA settings, the anticipated advantages of explicit foresight fail to materialize. In agent tasks, additional world model signals often introduce noise rather than guidance, leading most models (except the GPT-5 family) to perform \emph{worse} on average. In VQA, the effect is milder but still underwhelming: gains are marginal, and performance with or without world model access is nearly indistinguishable. These findings challenge the assumption that off-policy world model rollouts inherently strengthen downstream reasoning and instead suggest that current models struggle to leverage such information in a meaningful or stable way.

\begin{table}[t]
    \centering
    \setlength\tabcolsep{2pt}
    \setlength\extrarowheight{2pt}
    
    \tabcolsep=0.01\linewidth
    \resizebox{\linewidth}{!}{
    \begin{tabular}{l c c c c c}
        \toprule
        \textbf{Model} & \textbf{\makecell{Frozen\\Lake}} & \textbf{Navigate} & \textbf{\makecell{Primitive\\Skill}} & \textbf{Sokoban} & \textbf{Avg.} \\
        \addlinespace[2pt]
        \midrule
        \addlinespace[2pt]
        \multicolumn{6}{c}{\cellcolor[HTML]{EFEFEF} \textit{Agent Task World Model Usage Rate}} \\
        \midrule
        GPT-4o-mini & 0.3438 & 0.1367 & 0.3906 & 0.3906 & 0.2749 \\
        GPT-4o & 0.4531 & 0.1967 & 0.1289 & 0.0469 & 0.1813 \\
        GPT-5-mini & 0.7188 & 0.6100 & 0.5195 & 0.8750 & 0.6111 \\
        GPT-5 & 0.2031 & 0.3367 & 0.0742 & 0.1406 & 0.2076 \\
        Llama-4-Maverick & 0.9844 & 0.9933 & 1.0000 & 1.0000 & 0.9956 \\
        Llama-4-Scout & 0.8125 & 0.3300 & 0.9414 & 0.9375 & 0.6608 \\
        Qwen2.5-VL-7B & 0.1406 & 0.0167 & 0.1211 & 0.0000 & 0.0658 \\
        Qwen2.5-VL-32B & 0.6875 & 0.5433 & 0.5781 & 0.1406 & 0.5322 \\
        Qwen2.5-VL-72B & 0.6562 & 0.0033 & 0.4180 & 0.0156 & 0.2208 \\
        \bottomrule
        \addlinespace[2pt]
        \textbf{Model} & \textbf{3DSRBench} & \textbf{MMSI} & \textbf{SAT} & \textbf{Spatial} & \textbf{Avg.} \\
        \addlinespace[2pt]
        \midrule
        \addlinespace[2pt]
        \multicolumn{6}{c}{\cellcolor[HTML]{EFEFEF} \textit{VQA Task World Model Usage Rate}} \\
        \midrule
        GPT-4o-mini & 0.0818 & 0.3012 & 0.2533 & 0.0122 & 0.0972 \\
        GPT-4o & 0.0014 & 0.0087 & 0.0133 & 0.0000 & 0.0022 \\
        GPT-5-mini & 0.0103 & 0.0087 & 0.0133 & 0.0032 & 0.0087 \\
        GPT-5 & 0.0000 & 0.0000 & 0.0000 & 0.0000 & 0.0000 \\
        Llama-4-Maverick & 0.3863 & 0.7465 & 0.4533 & 0.0733 & 0.3678 \\
        Llama-4-Scout & 0.3923 & 0.5320 & 0.3733 & 0.1690 & 0.3639 \\
        Qwen2.5-VL-7B & 0.0047 & 0.0054 & 0.0467 & 0.0039 & 0.0054 \\
        Qwen2.5-VL-32B & 0.0080 & 0.0033 & 0.0067 & 0.0013 & 0.0060 \\
        Qwen2.5-VL-72B & 0.0105 & 0.0358 & 0.0467 & 0.0000 & 0.0121 \\
        \bottomrule
    \end{tabular}
    }
    \caption{World Model Usage Rate Statistics: Generally models are not that willing to use the world model as tool, especially for VQA tasks.}
    \label{tab:wm_use_stats}
\end{table}

\paragraph{Finding 2: Models Rarely Choose to Invoke the World Model.}
\Cref{tab:wm_use_stats} reports each model’s world model usage rate, defined as the fraction of cases in which at least one world model tool invocation is made. Usage is consistently low, particularly for VQA, where rates remain below 0.1 for all but the Llama family. This reluctance indicates that models generally do not recognize the world model as a valuable computational tool, even when demonstrations are provided. The hesitation also reflects a deeper issue: models lack a clear internal strategy for when and why world-model rollouts could meaningfully improve their predictions.

\paragraph{Finding 3: Different Model Families Exhibit Distinct Calling Behaviors.}
Despite the overall low usage rates, the patterns vary substantially across families. Llama models are the most proactive in querying the world model, but they show little measurable benefit. Within the GPT series, smaller models query more frequently, seemingly compensating for weaker internal reasoning, while larger models display high self-confidence and thus bypass external help. Qwen models follow a similar trend for agent tasks, except for Qwen2.5-VL-7B, which is unusually unwilling to call the world model despite being the smallest and least capable model among all that are tested. This suggests a form of potential misplaced confidence. Collectively, these patterns reveal that invocation habits are shaped more by a model’s perception than by the actual utility of the world model.

\begin{figure}[t]
    \centering
    \includegraphics[width=\linewidth]{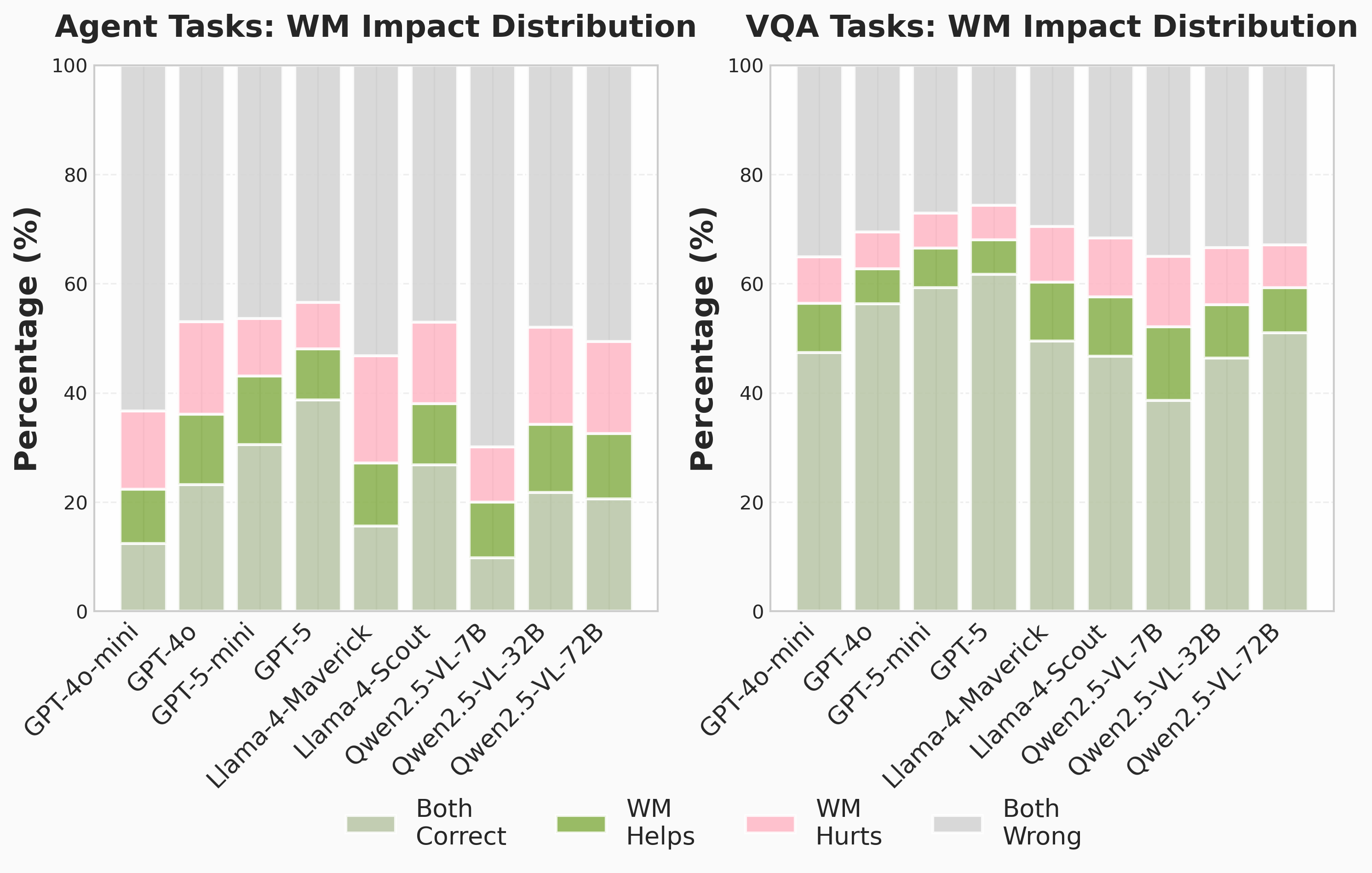}
    \caption{Percentage breakdown of the world model's impact. ``WM Helps'' is generally less frequent than ``WM Hurts'' for agent tasks, while VQA shows a more balanced distribution.}
    \label{fig:task_benefit_distribution}
\end{figure}

\paragraph{Finding 4: Fine-Grained Analysis Shows a Net Neutral Impact.}
To understand not just whether world models are invoked but whether they \emph{help}, we analyze case-level differences in \Cref{fig:task_benefit_distribution}. The distribution reveals that: for VQA tasks, ``WM Helps'' and ``WM Hurts'' occur at nearly equal rates, indicating that models cannot robustly leverage the foresight provided, and such guidance might be treated as noise. For agent tasks, harmful rollouts occur even more frequently than helpful ones. Overall, the detailed statistics reinforce a consistent conclusion: under current interfaces and model behaviors, world-model access offers limited practical advantage and can even be counterproductive. Understanding \emph{why} this happens requires a deeper examination of success and failure modes, which we analyze in the next section.

%%%%%%%%%%%%%%%%%%%%%%%%%%%%%%%%%%%%%%%%%%%%%%%%%%%%%%
%%%%%%%%%%%%%%%% Results and Analysis %%%%%%%%%%%%%%%%
%%%%%%%%%%%%%%%%%%%%%%%%%%%%%%%%%%%%%%%%%%%%%%%%%%%%%%

\section{Analysis}

\subsection{Attribution Analysis}

\begin{figure}[t]
    \centering
    \begin{minipage}{\linewidth}
        \centering
        \includegraphics[width=\linewidth]{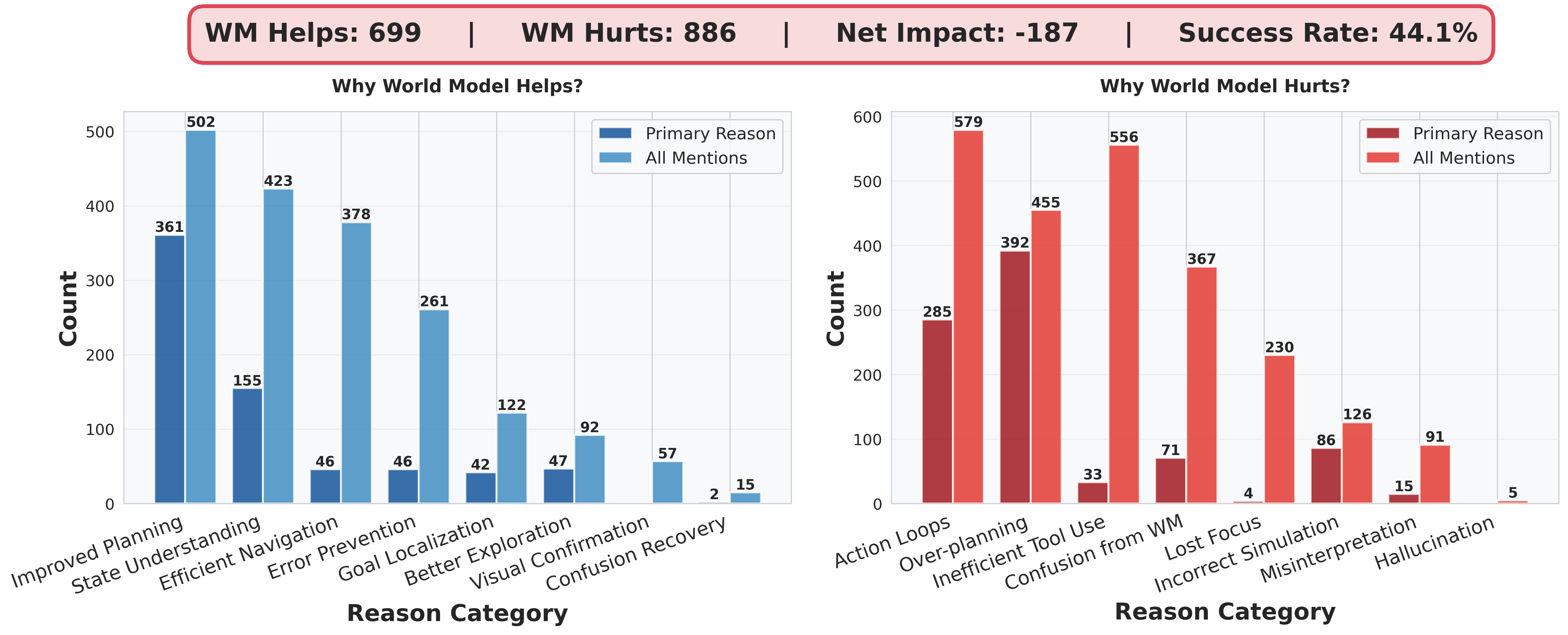}
    \end{minipage}
    \vspace{8pt}
    \begin{minipage}{\linewidth}
        \centering
        \includegraphics[width=\linewidth]{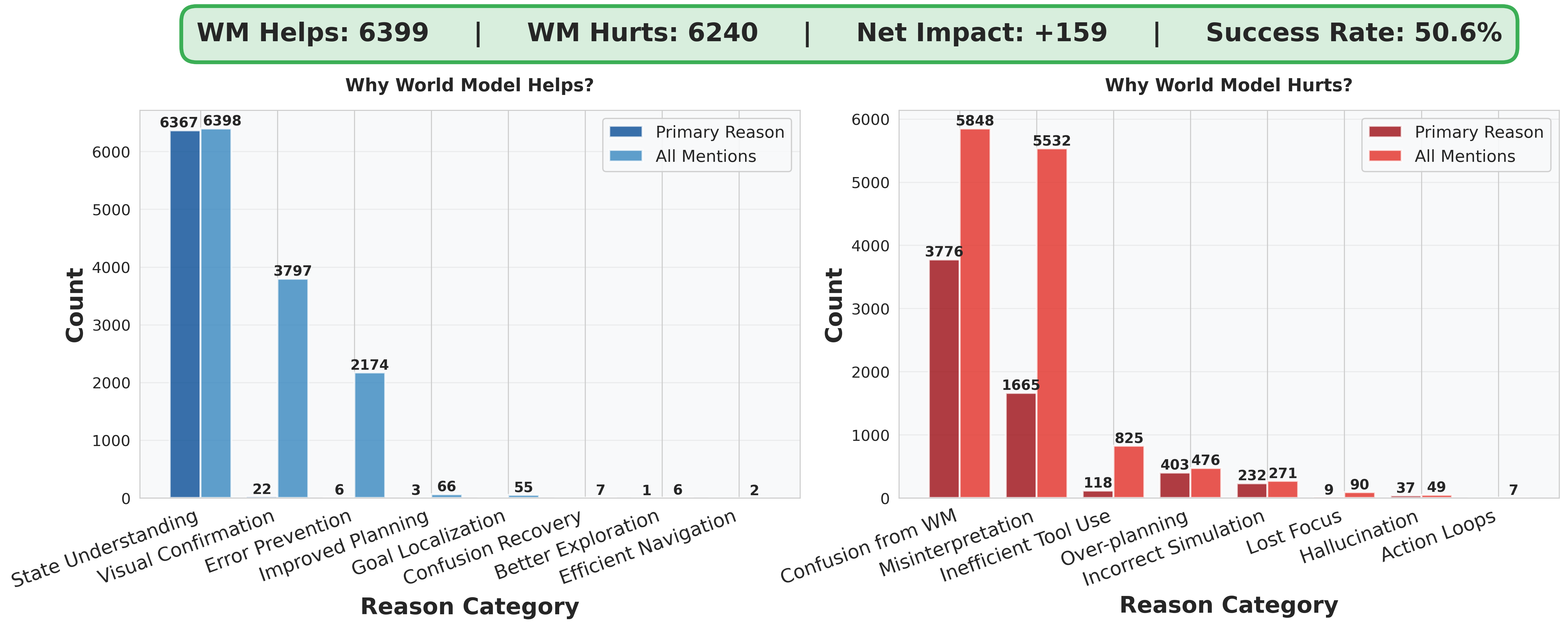}
    \end{minipage}
    \caption{Attributions for when world model use helps or hurts across Agent (top) and VQA (bottom) tasks.}
    \label{fig:wm_deep_analysis}
\end{figure}

\begin{figure}[t]
    \centering
    \includegraphics[width=\linewidth]{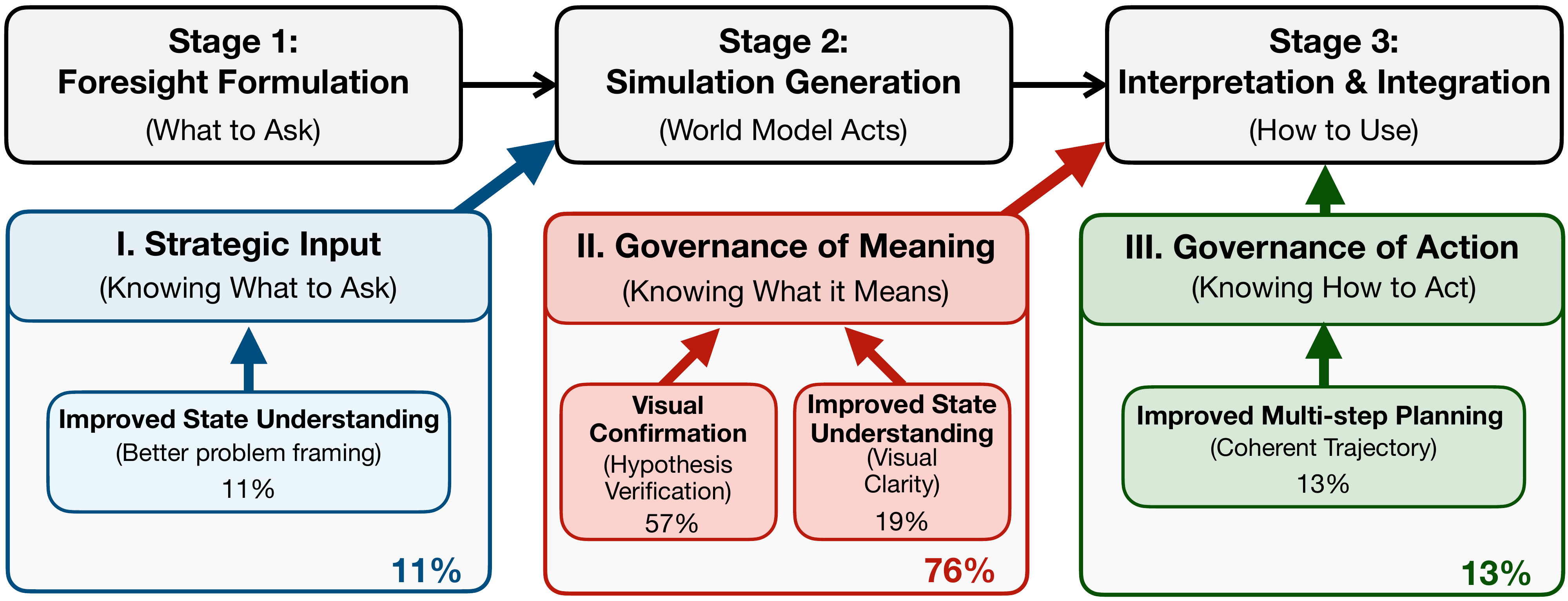}
    \caption{\textbf{A Taxonomy of World Model Governance Successes}: Correctly functioning governance follows a three-stage cognitive pipeline enabled by \emph{Strategic Input (I)}, \emph{Clear Interpretation (II)}, and \emph{Grounded Action (III)}. Success arises from calibrated queries (I), unambiguous verification (II), and stable integration of simulations into actionable plans (III).}
    \label{fig:wm_governance_taxonomy_success}
\end{figure}

\begin{figure}[t]
    \centering
    \includegraphics[width=\linewidth]{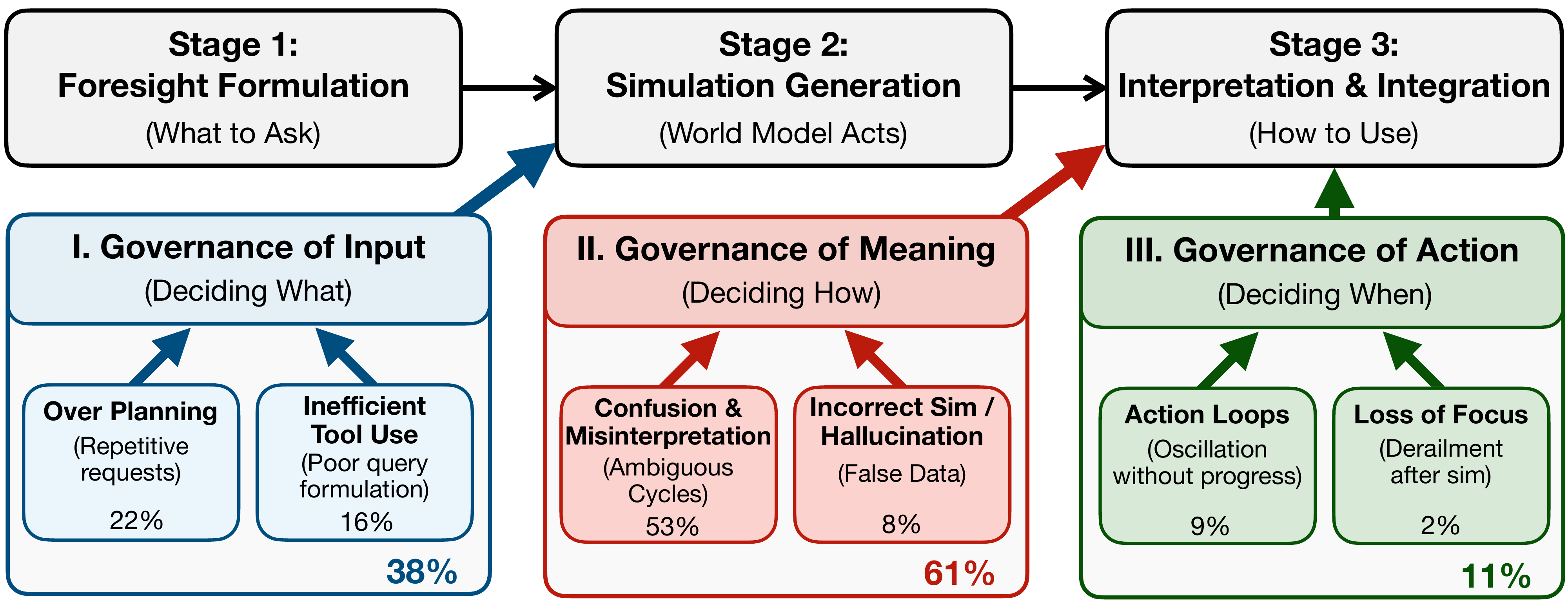}
    \caption{\textbf{A Taxonomy of World Model Governance Failures}: Pipeline breakdowns map to three disruptive pillars: \emph{Calibration Failures (I)} causing unnecessary or missed simulation, \emph{Interpretation Ambiguity (II)} corrupting signal-to-decision alignment, and \emph{Unstable Integration Policy (III)} preventing foresight from becoming sustained progress. The dominant zones indicate the key bottleneck is governance stability, instead of foresight generation.}
    \label{fig:wm_governance_taxonomy_fail}
\end{figure}

We categorize eight major ways in which world model simulations influence outcomes and annotate both primary and contributing reasons for each case with the help of GPT-4o (see \Cref{apdx:attribution_details} for details).
The aggregated statistics in \Cref{fig:wm_deep_analysis} uncover a consistent pattern: models can extract value from foresight, but their use of it is poorly calibrated, leading to fragile gains and frequent regressions. These patterns further motivate a structured taxonomy of world model governance successes (\Cref{fig:wm_governance_taxonomy_success}) and failures (\Cref{fig:wm_governance_taxonomy_fail}) across the cognitive pipeline.

\paragraph{Finding 5: Agent Tasks Provide Useful Planning Signals, but Instability in Execution.}
In agent tasks, rollouts help mainly by improving planning and state understanding, confirming that simulated futures \emph{can} guide multi-step decisions. However, failures are more frequent and more varied. The prevalence of action loops and over-planning shows that models tend to overreact to simulated information, repeatedly re-planning without making meaningful progress. Additional failure modes, including inefficient tool use, misinterpreting simulations, and loss of focus, all point to a deeper issue: models lack a stable policy for \emph{how} to integrate world-model feedback. Instead of grounding simulations in a coherent trajectory, they oscillate between cautious over-analysis and unfocused repetition. This makes foresight beneficial in isolated cases but detrimental over long horizons.
In \Cref{fig:wm_governance_taxonomy_fail}, these behavior mostly corresponds to failures in \emph{governance of action}, where agents lack a stable policy for grounding simulated foresight into coherent, forward progress.

\paragraph{Finding 6: VQA Tasks Can Provide Targeted Benefits but Amplified Ambiguity.}
For VQA, the influence of world models is more concentrated. Rollouts help primarily through improved state understanding and visual confirmation, which acts as a safeguard against committing an incorrect answer. Yet the dominant failure modes, confusion and misinterpretation, stem from a feedback loop: the model formulates unclear simulation requests, the world model returns ambiguous scenarios, and these further mislead the VLM. Thus, instead of correcting uncertainty, simulations often magnify it when the initial instruction is under-specified.
In \Cref{fig:wm_governance_taxonomy_fail}, these errors reflect failures in \emph{governance of meaning}, where ambiguity in simulation requests and outcomes propagates rather than resolves uncertainty.

\paragraph{Implications: The Core Bottleneck Is Foresight Governance.}
Instead of struggling with future simulation, current agent often fails in governing foresight: deciding what to simulate (input governance), how to interpret it (meaning governance), and when to act upon it (action governance).
In agent tasks this leads to unstable execution, while in VQA it amplifies ambiguity. Thus, improving the world model alone is insufficient: what is missing is an internal mechanism for principled, calibrated use of foresight. Designing such mechanisms is crucial for realizing the potential of world model–augmented systems.
\Cref{fig:wm_governance_taxonomy_success} and \Cref{fig:wm_governance_taxonomy_fail} summarize these patterns as complementary taxonomies of world model governance failures and successes, revealing that stable foresight hinges more on governance rather than simulation.

\subsection{Further Studies}

\begin{figure}[t]
    \centering
    \includegraphics[width=\linewidth]{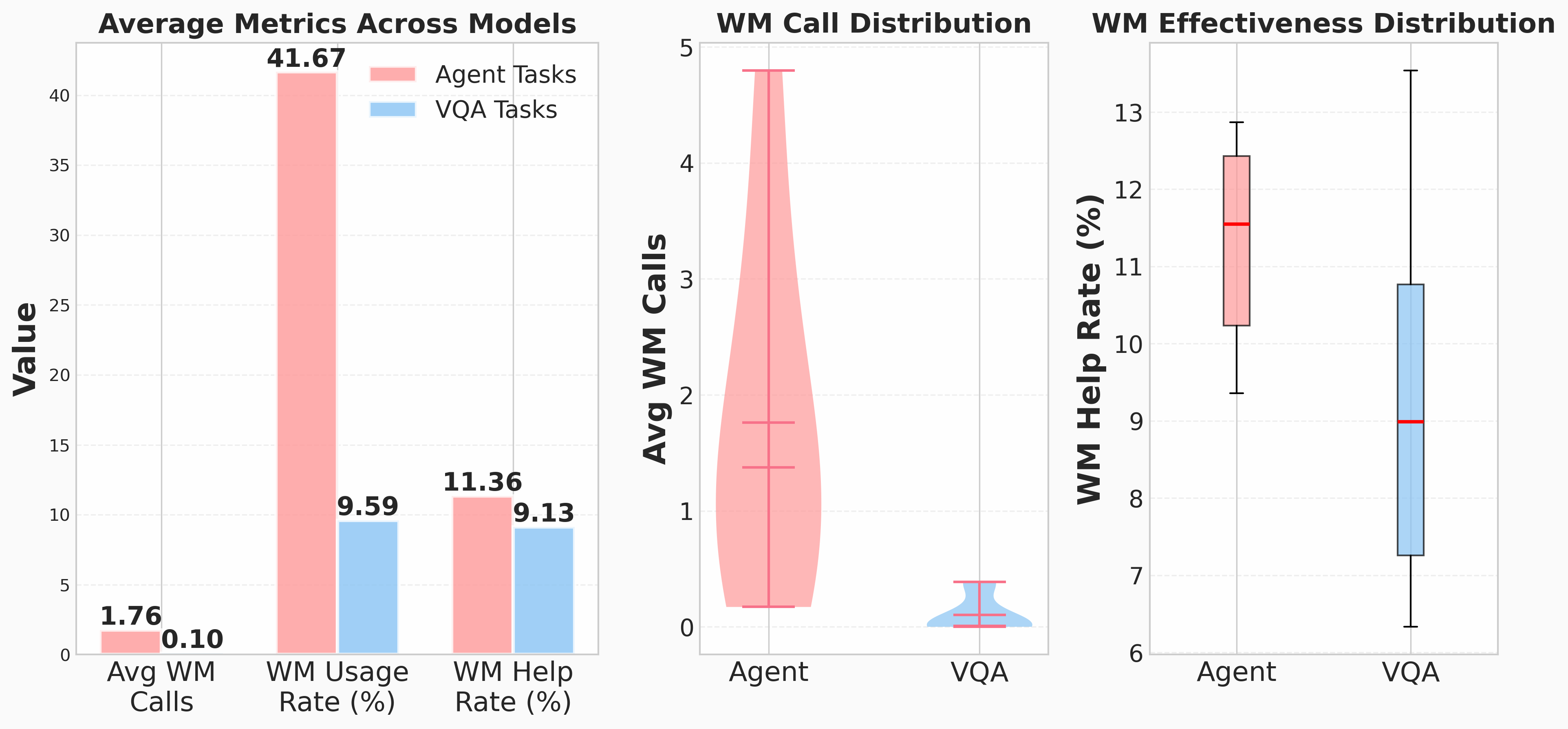}
    \caption{Comparison of Agent and VQA tasks. Former shows higher world model usage and higher help rate.}
    \label{fig:task_aggregate_comparison}
\end{figure}

Beyond attribution-level patterns, we further analyze how world model usage behaves across tasks, call frequencies, and beyond.

\paragraph{Task Perspective: Agent Tasks Leverage Foresight More Effectively.}
As shown in \Cref{fig:task_aggregate_comparison}, agent tasks consistently exhibit higher world-model usage and a higher probability that rollouts provide actual benefit. This aligns with the structure of agentic problems, where state transitions and sequential decisions naturally create opportunities for foresight to guide behavior. In contrast, VQA demands highly targeted disambiguation; without precise simulation requests, world model information is more likely to distract than assist. The gap between the two tasks suggests that current invocation patterns are better suited for dynamic, stateful environments than for static perception queries, though the later can also benefit from the world model as tool paradigm.

\begin{figure}[t]
    \centering
    \includegraphics[width=\linewidth]{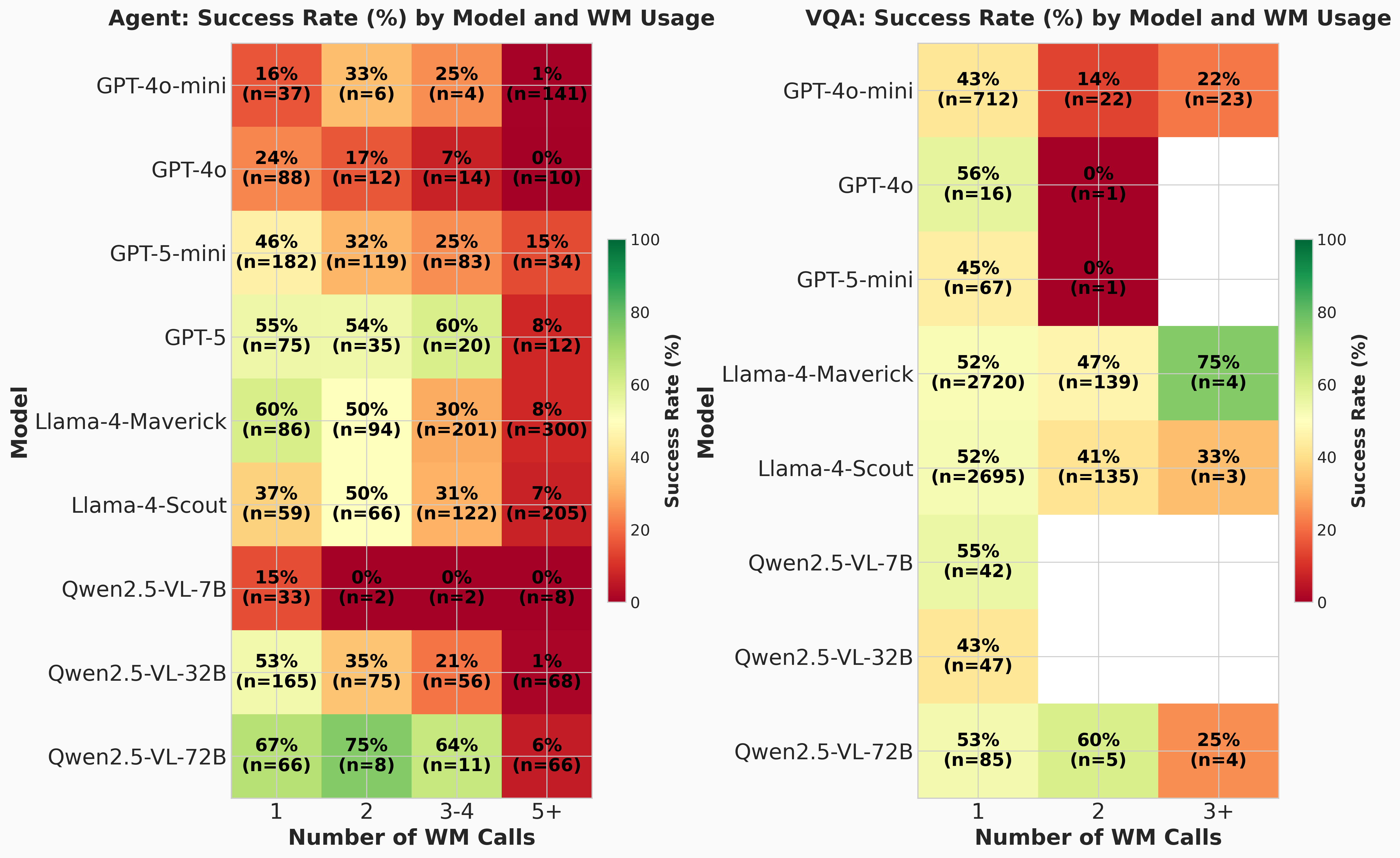}
    \caption{Success rate as a function of world model call count. More calls often correlate with worse outcomes.}
    \label{fig:wm_effectiveness_heatmap}
\end{figure}

\paragraph{Call Perspective: More Calls Reflect Uncertainty, Not Better Reasoning.}
\Cref{fig:wm_effectiveness_heatmap} shows a clear negative correlation between the number of world model calls and task success. Rather than accumulating useful evidence, repeated calls often signal unresolved confusion instead of scaling effect: models re-query instead of integrating prior rollouts into a stable plan. This is especially visible in agent tasks, where excessive calls frequently coincide with action loops and degraded execution. The pattern reinforces a central conclusion of our empirical evidence: the challenge is not generating simulations, but knowing when to stop, how to interpret them, and how to integrate them.

\begin{figure}[t]
    \centering
    \includegraphics[width=\linewidth]{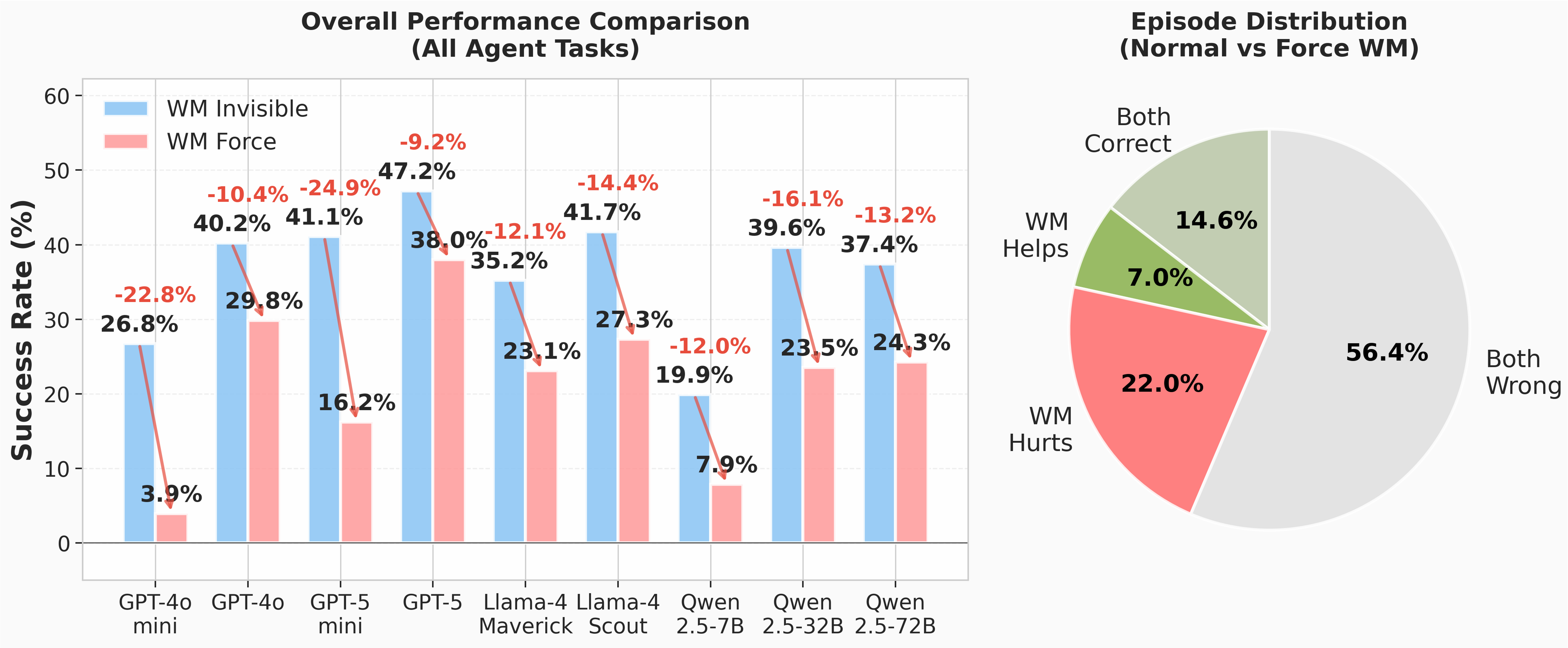}
    \caption{Comparison of agent task performance between world model invisible and forced world model use conditions.}
    \label{fig:force_wm_comparison_agent}
\end{figure}

\paragraph{Usage Perspective: Forcing World Model Invocation Degrades Performance.}
Across all agent tasks, we also evaluate a setting where the model is \emph{forced} to consult its world model before producing an action. As shown in \Cref{fig:force_wm_comparison_agent}, this intervention leads to an even steeper drop in performance than the normal mode with world model access. This outcome aligns with earlier findings: when optional world model access already introduces systematic failure modes, mandating its use amplifies those weaknesses. Both the frequency and severity of cases in which the world model hurts increase, and average performance across models declines for every task. Taken together, these results indicate that, under current designs, forcing world model usage is not effective.

\section{Discussions}
\label{sec:discussion}

\paragraph{World Model for Agent Confirmation.}
Attribution analysis suggests world models mostly help as \textbf{visual confirmation} in VQA: the tested model forms a hypothesis and asks the world model to verify it, which can filter out implausible actions. However, when the initial hypothesis is wrong, a confirmation-style query can lock in the error by producing a plausible simulation aligned with that wrong belief, empirically matching the cases where world models hurt. This reflects a broader limitation: current agents treat the world model as a verifier of a single guess rather than a tool to surface alternatives the agent has not yet considered. A more corrective protocol is to promote \textbf{discrimination} rather than affirmation. Specifically, the agent may propose several plausible hypotheses, simulate each with the world model, and pick the one whose predicted cues best match the observation. This turns simulation into structured hypothesis testing, thus increasing the chance of correcting an initial error rather than reinforcing it.

\paragraph{Dedicated Modules for Integration of Foresight.}
Foresight is currently integrated by simply expanding the context over multiple rounds, with demonstrations in the system prompt. However in practice, agents often discount simulator observations and instead become more confident in their initial (incorrect) beliefs. A more reliable approach is to introduce dedicated modules that enforce a structured interaction loop between reasoning and foresight: (i) \textbf{Decider}: determines whether to invoke the world model or execute a real action, and boldly proposes candidate actions to test; (ii) \textbf{Reflector}: evaluates observations from real execution or simulation, checks whether outcomes match the expectations, and feeds back what to adjust next; (iii) \textbf{Memory}: maintains long and short-term task objectives by storing observations and selectively releasing relevant context to the Decider and Reflector. Overall, a dedicated mechanism can make foresight usage explicit and organized, reducing sprawling, brittle chains of multimodal reasoning where the agent may get lost in the middle.

\paragraph{Agent Training for Better World Model Interactions.}
Still, without fine-tuning it is difficult to reshape an agent's default behavior: prompting rarely teaches \emph{when} to invoke a world model or \emph{how} to query it effectively. To intrinsically improve this paradigm, training is necessary. One practical route is RL with online, multi-turn rollouts where the world model is an explicit tool, using rewards that go beyond final answer correctness to encourage: (i) \textbf{appropriate} world model invocation, and (ii) \textbf{diverse, distinctive} queries (e.g., measured semantically). The challenge is credit assignment: we want to reward useful interaction without inducing indiscriminate calling. In addition to penalties for excessive or repeated invocations, two complementary remedies are: (i) constructing high-quality supervised fine-tuning data to set strong cold start interaction habits; and (ii) using indirect objectives such as rewarding information gain (e.g., reductions in hypothesis entropy). These incentives directly target the attribution-identified failure modes and should yield more strategic and exploratory world model use.

\section{Conclusion and Future Work}
Our investigation reveals that giving agents access to a world model reshapes their behavior in unexpected ways. Rather than serving as a straightforward enhancement, simulation introduces new cognitive pressures: agents must manage hypothetical branches and maintain coherent reasoning across mixed real and imagined experience. The difficulties we observe, including hesitation, over-analysis, and misaligned interpretation, suggest that effective foresight requires more fine-grained governance. Looking ahead, promising directions include learning architectures that maintain stable internal state across simulated and real trajectories, exploration-driven simulation policies, and interfaces that encourage agents to articulate and refine their hypotheses before querying a world model. Beyond performance, studying how agents build and anchor beliefs across possible futures may deepen anticipatory cognition and integrate world model simulation directly into decision-making.

\section*{Limitations}
Our study covers a limited set of model families; we currently exclude Gemini and Claude due to cost constraints. While this narrows breadth, the models we evaluate are widely used and suffice to probe general trends in the willingness to employ world models and the effectiveness of doing so.
Our evaluation also uses different simulators across task types: agentic tasks rely on a ground-truth simulator, whereas VQA uses WAN2.1 as the simulator. This design choice keeps the focus on agent-side behavior rather than simulator quality, but it may introduce mild cross-task incomparability. In practice, we observe occasional quality artifacts in WAN2.1-generated frames that can confuse agents. Notably, many failure cases trace back to underspecified or ambiguous agent instructions, which degrade simulation quality and in turn affect downstream performance.
We view these as actionable future directions. Future work will expand coverage to additional model families and employ stronger, more consistent simulators to better isolate agent effectiveness.

\section*{Ethical Statement}
This paper investigates ``world models as tools'' for VLM agents in controlled experiments (simulating in a cloned environment state or via generative rollouts) rather than real-world deployment. A key ethical risk is that simulated futures can be over-trusted or misread, leading to unsafe decisions if transferred to high-stakes domains; in fact, our results show that forcing simulation can degrade performance and create new failure modes, underscoring that naive use is not reliably beneficial. To reduce potential harm, we recommend treating simulations as uncertain hypotheses instead of ground truth, encouraging multi-hypothesis/counterfactual checks rather than ``confirmation'' rollouts, and avoiding safety-critical deployment without robust oversight, evaluation, and privacy protections when user data is involved.

\bibliography{custom}

\clearpage
\appendix

\section*{Appendix}
\label{sec:appendix}

\section{Evaluation Settings Details}
\label{apdx:experiment_settings}
In this section, we describe the evaluation settings and hyperparameters used throughout our experiments. Unless otherwise specified, all tested models are evaluated with an inference temperature of 0.0 and a maximum output length of 2048 tokens per turn. Each episode is constrained to a maximum of 15 turns across all tasks.

For agentic tasks, we directly leverage the real environment simulation as the tested model's foresight mechanism. Specifically, we use all frames produced by the environment when executing the model's predicted simulation actions, and we cap the number of incorporated frames to the reasoning trajectory at three for each step.

For VQA tasks, we adopt WAN2.1 as the world model and use Qwen2.5-3B-VL for prompt augmentation, following official recommendations. The hyperparameters for generating world model frames are provided in the following:
\begin{table}[h!]
\vspace{0pt}
\renewcommand{\arraystretch}{1.2}
\centering
\resizebox{0.75\columnwidth}{!}{
\rowcolors{2}{gray!10}{white}
\begin{tabular*}{\linewidth}{p{0.65\linewidth} p{0.28\linewidth}}
\toprule
\textbf{Hyperparameter} & \textbf{Value} \\
\midrule
Frame Num      & 21 \\
Steps           & 15 \\
Shift           & 16.0 \\
Guide Scale    & 5.0 \\
Sample Solver  & unipc \\
Seed            & -1 \\
Timeout         & 1200 \\
\bottomrule
\end{tabular*}
}
\vspace{-2pt}
\caption{Hyperparameters used in WAN2.1 as world model.}
\vspace{-5pt}
\end{table}

In both categories of tasks, we limit the number of frames incorporated into the model’s reasoning trajectory to three. To reduce redundancy, we first filter frames so that the similarity between any two consecutive frames remains below 0.95. From the filtered set, we then uniformly interpolate the sequence to obtain three representative frames. The final frame generated by the world model is always included in the returned set.

For each data point, we perform a single evaluation. Task accuracy is then computed by averaging the accuracy across all data points, where accuracy corresponds to the success rate for agent tasks and the multiple choice accuracy for VQA tasks.

\section{Additional Experiment Results}
\label{apdx:additional_experiment}

\begin{table}[t]
    \centering
    \setlength\tabcolsep{2pt}
    \setlength\extrarowheight{2pt}
    
    \tabcolsep=0.01\linewidth
    \resizebox{\linewidth}{!}{
    \begin{tabular}{l c c c c c}
        \toprule
        \textbf{Model} & \textbf{\makecell{Frozen\\Lake}} & \textbf{Navigate} & \textbf{\makecell{Primitive\\Skill}} & \textbf{Sokoban} & \textbf{Avg.} \\
        \addlinespace[2pt]
        \midrule
        \addlinespace[2pt]
        \multicolumn{6}{c}{\cellcolor[HTML]{EFEFEF} \textit{Without World Model Access}} \\
        \midrule
        GPT-4o-mini & 0.36 & 0.26 & 0.32 & 0.00 & 0.27 \\
        GPT-4o & 0.58 & 0.35 & 0.51 & 0.02 & 0.40 \\
        GPT-5-mini & 0.86 & 0.66 & 0.11 & 0.03 & 0.41 \\
        GPT-5 & 0.77 & 0.74 & 0.19 & 0.06 & 0.47 \\
        Llama-4-Maverick & 0.70 & 0.31 & 0.40 & 0.00 & 0.35 \\
        Llama-4-Scout & 0.59 & 0.55 & 0.32 & 0.02 & 0.42 \\
        Qwen2.5-VL-7B & 0.36 & 0.26 & 0.13 & 0.02 & 0.20 \\
        Qwen2.5-VL-32B & 0.59 & 0.36 & 0.49 & 0.00 & 0.40 \\
        Qwen2.5-VL-72B & 0.61 & 0.38 & 0.41 & 0.00 & 0.37 \\
        \addlinespace[2pt]
        \midrule
        \addlinespace[2pt]
        \multicolumn{6}{c}{\cellcolor[HTML]{EFEFEF} \textit{With World Model Access}} \\
        \midrule
        GPT-4o-mini & 0.39 & 0.24 & 0.22 & 0.00 & 0.22 \\
        GPT-4o & 0.58 & 0.31 & 0.46 & 0.02 & 0.36 \\
        GPT-5-mini & 0.89 & 0.63 & 0.19 & 0.00 & 0.43 \\
        GPT-5 & 0.89 & 0.71 & 0.20 & 0.14 & 0.48 \\
        Llama-4-Maverick & 0.66 & 0.20 & 0.32 & 0.03 & 0.27 \\
        Llama-4-Scout & 0.55 & 0.54 & 0.24 & 0.03 & 0.38 \\
        Qwen2.5-VL-7B & 0.45 & 0.26 & 0.12 & 0.00 & 0.20 \\
        Qwen2.5-VL-32B & 0.58 & 0.32 & 0.39 & 0.02 & 0.34 \\
        Qwen2.5-VL-72B & 0.45 & 0.37 & 0.32 & 0.00 & 0.33 \\
        \addlinespace[2pt]
        \midrule
        \addlinespace[2pt]
        \multicolumn{6}{c}{\cellcolor[HTML]{EFEFEF} \textit{Force WM Use}} \\
        \midrule
        GPT-4o-mini & 0.06 & 0.05 & 0.03 & 0.00 & 0.04 \\
        GPT-4o & 0.58 & 0.21 & 0.40 & 0.00 & 0.30 \\
        GPT-5-mini & 0.67 & 0.21 & 0.02 & 0.02 & 0.16 \\
        GPT-5 & 0.91 & 0.58 & 0.09 & 0.06 & 0.38 \\
        Llama-4-Maverick & 0.64 & 0.12 & 0.31 & 0.00 & 0.23 \\
        Llama-4-Scout & 0.50 & 0.36 & 0.18 & 0.00 & 0.27 \\
        Qwen2.5-VL-7B & 0.25 & 0.12 & 0.01 & 0.00 & 0.08 \\
        Qwen2.5-VL-32B & 0.39 & 0.25 & 0.24 & 0.00 & 0.24 \\
        Qwen2.5-VL-72B & 0.53 & 0.27 & 0.20 & 0.00 & 0.24 \\
        \bottomrule
    \end{tabular}
    }
    \caption{Agent Task Success Rate: Comprehensive comparison between without WM access, with WM access, and force WM use.}
    \label{tab:apdx_agent_success_rate}
\end{table}

\begin{table}[t]
    \centering
    \setlength\tabcolsep{2pt}
    \setlength\extrarowheight{2pt}
    
    \tabcolsep=0.018\linewidth
    \resizebox{\linewidth}{!}{
    \begin{tabular}{l c c c c c}
        \toprule
        \textbf{Model} & \textbf{3DSRBench} & \textbf{MMSI} & \textbf{SAT} & \textbf{Spatial} & \textbf{Avg.} \\
        \addlinespace[2pt]
        \midrule
        \addlinespace[2pt]
        \multicolumn{6}{c}{\cellcolor[HTML]{EFEFEF} \textit{Without World Model Access}} \\
        \midrule
        GPT-4o-mini & 0.58 & 0.28 & 0.52 & 0.65 & 0.56 \\
        GPT-4o & 0.66 & 0.31 & 0.71 & 0.72 & 0.63 \\
        GPT-5-mini & 0.67 & 0.35 & 0.85 & 0.78 & 0.66 \\
        GPT-5 & 0.69 & 0.38 & 0.86 & 0.80 & 0.68 \\
        Llama-4-Maverick & 0.61 & 0.27 & 0.52 & 0.74 & 0.60 \\
        Llama-4-Scout & 0.59 & 0.27 & 0.41 & 0.74 & 0.58 \\
        Qwen2.5-VL-7B & 0.53 & 0.24 & 0.59 & 0.62 & 0.52 \\
        Qwen2.5-VL-32B & 0.59 & 0.30 & 0.47 & 0.67 & 0.57 \\
        Qwen2.5-VL-72B & 0.61 & 0.29 & 0.47 & 0.71 & 0.59 \\
        \addlinespace[2pt]
        \midrule
        \addlinespace[2pt]
        \multicolumn{6}{c}{\cellcolor[HTML]{EFEFEF} \textit{With World Model Access}} \\
        \midrule
        GPT-4o-mini & 0.59 & 0.27 & 0.57 & 0.66 & 0.56 \\
        GPT-4o & 0.66 & 0.30 & 0.73 & 0.72 & 0.63 \\
        GPT-5-mini & 0.68 & 0.36 & 0.83 & 0.79 & 0.67 \\
        GPT-5 & 0.70 & 0.37 & 0.85 & 0.79 & 0.68 \\
        Llama-4-Maverick & 0.62 & 0.28 & 0.47 & 0.75 & 0.60 \\
        Llama-4-Scout & 0.59 & 0.28 & 0.36 & 0.73 & 0.58 \\
        Qwen2.5-VL-7B & 0.54 & 0.24 & 0.66 & 0.63 & 0.52 \\
        Qwen2.5-VL-32B & 0.58 & 0.28 & 0.47 & 0.67 & 0.56 \\
        Qwen2.5-VL-72B & 0.61 & 0.29 & 0.48 & 0.73 & 0.59 \\
        \addlinespace[2pt]
        \midrule
        \addlinespace[2pt]
        \multicolumn{6}{c}{\cellcolor[HTML]{EFEFEF} \textit{Force WM Use}} \\
        \midrule
        GPT-4o-mini & 0.59 & 0.28 & 0.61 & 0.64 & 0.56 \\
        GPT-4o & 0.66 & 0.29 & 0.67 & 0.72 & 0.63 \\
        GPT-5-mini & 0.67 & 0.34 & 0.81 & 0.80 & 0.66 \\
        GPT-5 & 0.69 & 0.33 & 0.85 & 0.79 & 0.67 \\
        Llama-4-Maverick & 0.61 & 0.30 & 0.50 & 0.73 & 0.59 \\
        Llama-4-Scout & 0.59 & 0.28 & 0.41 & 0.73 & 0.58 \\
        Qwen2.5-VL-7B & 0.51 & 0.22 & 0.37 & 0.55 & 0.48 \\
        Qwen2.5-VL-32B & 0.57 & 0.27 & 0.38 & 0.66 & 0.55 \\
        Qwen2.5-VL-72B & 0.59 & 0.28 & 0.49 & 0.71 & 0.58 \\
        \bottomrule
    \end{tabular}
    }
    \caption{VQA Task Accuracy: Comprehensive comparison between without WM access, with WM access, and force WM use.}
    \label{tab:apdx_vqa_accuracy}
\end{table}

\paragraph{Additional Main Experiment Results.}
In \Cref{tab:apdx_agent_success_rate} and \Cref{tab:apdx_vqa_accuracy}, we extend our original results by reporting success rates and accuracies under the forced world model usage setting. The results indicate that mandating the use of the world model does not improve, but often degrades, the effectiveness of world model utilization by the tested models. In fact, forcing world model usage generally yields worse performance than allowing the model to access the world model as an optional tool. These findings provide a more direct comparison and further support our main claim that compulsory world model usage is not a scalable or effective strategy for current VLM-based agents.

\begin{table}[t]
    \centering
    \setlength\tabcolsep{2pt}
    \setlength\extrarowheight{2pt}
    
    \tabcolsep=0.01\linewidth
    \resizebox{\linewidth}{!}{
    \begin{tabular}{l c c c c c}
        \toprule
        \textbf{Model} & \textbf{FrozenLake} & \textbf{Navigation} & \textbf{PrimitiveSkill} & \textbf{Sokoban} & \textbf{Overall} \\
        \addlinespace[2pt]
        \midrule
        \addlinespace[2pt]
        \multicolumn{6}{c}{\cellcolor[HTML]{EFEFEF} \textit{Average WM Calls}} \\
        \midrule
        GPT-4o-mini & 1.3281 & 1.2467 & 4.1484 & 3.3906 & 2.5409 \\
        GPT-4o & 0.6250 & 0.2100 & 0.5000 & 0.1250 & 0.3494 \\
        GPT-5-mini & 1.5625 & 1.2833 & 0.9375 & 2.9375 & 1.3348 \\
        GPT-5 & 0.2188 & 0.8067 & 0.0742 & 0.1875 & 0.4196 \\
        Llama-4-Maverick & 3.0000 & 5.6533 & 4.0352 & 5.6250 & 4.7968 \\
        Llama-4-Scout & 4.1719 & 1.0933 & 4.5352 & 7.7812 & 3.2953 \\
        Qwen2.5-VL-7B & 0.1406 & 0.0900 & 0.3164 & 0.0000 & 0.1711 \\
        Qwen2.5-VL-32B & 1.0312 & 1.7167 & 1.7969 & 0.4688 & 1.5658 \\
        Qwen2.5-VL-72B & 1.6406 & 0.0267 & 3.2305 & 0.0156 & 1.3757 \\
        \addlinespace[2pt]
        \midrule
        \addlinespace[2pt]
        \multicolumn{6}{c}{\cellcolor[HTML]{EFEFEF} \textit{WM Usage Rate}} \\
        \midrule
        GPT-4o-mini & 0.3438 & 0.1367 & 0.3906 & 0.3906 & 0.2749 \\
        GPT-4o & 0.4531 & 0.1967 & 0.1289 & 0.0469 & 0.1813 \\
        GPT-5-mini & 0.7188 & 0.6100 & 0.5195 & 0.8750 & 0.6111 \\
        GPT-5 & 0.2031 & 0.3367 & 0.0742 & 0.1406 & 0.2076 \\
        Llama-4-Maverick & 0.9844 & 0.9933 & 1.0000 & 1.0000 & 0.9956 \\
        Llama-4-Scout & 0.8125 & 0.3300 & 0.9414 & 0.9375 & 0.6608 \\
        Qwen2.5-VL-7B & 0.1406 & 0.0167 & 0.1211 & 0.0000 & 0.0658 \\
        Qwen2.5-VL-32B & 0.6875 & 0.5433 & 0.5781 & 0.1406 & 0.5322 \\
        Qwen2.5-VL-72B & 0.6562 & 0.0033 & 0.4180 & 0.0156 & 0.2208 \\
        \bottomrule
    \end{tabular}
    }
    \caption{Agent Task WM Statistics (With WM Access): Upper section shows average WM calls per episode. Lower section shows WM usage rate (percentage of episodes that used WM at least once).}
    \label{tab:apdx_agent_wm_stats}
\end{table}

\begin{table}[t]
    \centering
    \setlength\tabcolsep{2pt}
    \setlength\extrarowheight{2pt}
    
    \tabcolsep=0.01\linewidth
    \resizebox{\linewidth}{!}{
    \begin{tabular}{l c c c c c}
        \toprule
        \textbf{Model} & \textbf{FrozenLake} & \textbf{Navigation} & \textbf{PrimitiveSkill} & \textbf{Sokoban} & \textbf{Overall} \\
        \addlinespace[2pt]
        \midrule
        \addlinespace[2pt]
        \multicolumn{6}{c}{\cellcolor[HTML]{EFEFEF} \textit{Average WM Calls}} \\
        \midrule
        GPT-4o-mini & 5.9062 & 13.9467 & 12.2969 & 10.5781 & 12.2617 \\
        GPT-4o & 2.5938 & 7.9900 & 6.9102 & 3.2656 & 6.6389 \\
        GPT-5-mini & 5.9062 & 12.1667 & 8.0664 & 8.7500 & 9.7266 \\
        GPT-5 & 1.6719 & 6.7300 & 3.0273 & 1.8594 & 4.4152 \\
        Llama-4-Maverick & 2.6094 & 9.7300 & 6.9805 & 7.7500 & 7.8494 \\
        Llama-4-Scout & 7.7500 & 9.1033 & 8.1367 & 10.6562 & 8.7602 \\
        Qwen2.5-VL-7B & 4.5156 & 7.1867 & 6.8164 & 8.0000 & 6.8743 \\
        Qwen2.5-VL-32B & 2.9531 & 8.6767 & 6.8594 & 8.0938 & 7.4064 \\
        Qwen2.5-VL-72B & 3.6562 & 9.1633 & 8.7891 & 9.7188 & 8.5599 \\
        \addlinespace[2pt]
        \midrule
        \addlinespace[2pt]
        \multicolumn{6}{c}{\cellcolor[HTML]{EFEFEF} \textit{WM Usage Rate}} \\
        \midrule
        GPT-4o-mini & 1.0000 & 1.0000 & 1.0000 & 1.0000 & 1.0000 \\
        GPT-4o & 1.0000 & 1.0000 & 1.0000 & 1.0000 & 1.0000 \\
        GPT-5-mini & 1.0000 & 1.0000 & 0.9492 & 1.0000 & 0.9810 \\
        GPT-5 & 0.9375 & 0.9967 & 0.7930 & 0.7812 & 0.8947 \\
        Llama-4-Maverick & 1.0000 & 1.0000 & 1.0000 & 1.0000 & 1.0000 \\
        Llama-4-Scout & 1.0000 & 1.0000 & 1.0000 & 1.0000 & 1.0000 \\
        Qwen2.5-VL-7B & 1.0000 & 0.9000 & 1.0000 & 1.0000 & 0.9561 \\
        Qwen2.5-VL-32B & 1.0000 & 1.0000 & 1.0000 & 1.0000 & 1.0000 \\
        Qwen2.5-VL-72B & 1.0000 & 1.0000 & 1.0000 & 1.0000 & 1.0000 \\
        \bottomrule
    \end{tabular}
    }
    \caption{Agent Task WM Statistics (Force WM Use): Upper section shows average WM calls per episode when forced to use WM. Lower section shows WM usage rate (expected to be 100\% in forced mode).}
    \label{tab:apdx_agent_wm_stats_force}
\end{table}

\begin{table}[t]
    \centering
    \setlength\tabcolsep{2pt}
    \setlength\extrarowheight{2pt}
    
    \tabcolsep=0.01\linewidth
    \resizebox{\linewidth}{!}{
    \begin{tabular}{l c c c c c}
        \toprule
        \textbf{Model} & \textbf{FrozenLake} & \textbf{Navigation} & \textbf{PrimitiveSkill} & \textbf{Sokoban} & \textbf{Overall} \\
        \addlinespace[2pt]
        \midrule
        \addlinespace[2pt]
        \multicolumn{6}{c}{\cellcolor[HTML]{EFEFEF} \textit{Without WM Access}} \\
        \midrule
        GPT-4o-mini & 7.81 & 12.91 & 4.12 & 15.00 & 9.34 \\
        GPT-4o & 3.75 & 11.27 & 4.36 & 14.78 & 8.31 \\
        GPT-5-mini & 1.55 & 7.95 & 5.91 & 14.12 & 7.17 \\
        GPT-5 & 4.75 & 7.52 & 6.92 & 14.02 & 7.65 \\
        Llama-4-Maverick & 2.27 & 12.73 & 6.52 & 15.00 & 9.64 \\
        Llama-4-Scout & 3.12 & 9.86 & 6.88 & 14.78 & 8.58 \\
        Qwen2.5-VL-7B & 6.09 & 12.16 & 8.12 & 14.80 & 10.33 \\
        Qwen2.5-VL-32B & 2.03 & 11.48 & 4.98 & 15.00 & 8.49 \\
        Qwen2.5-VL-72B & 2.44 & 10.96 & 4.58 & 15.00 & 8.15 \\
        \addlinespace[2pt]
        \midrule
        \addlinespace[2pt]
        \multicolumn{6}{c}{\cellcolor[HTML]{EFEFEF} \textit{With WM Access}} \\
        \midrule
        GPT-4o-mini & 7.89 & 12.94 & 8.58 & 15.00 & 11.03 \\
        GPT-4o & 5.08 & 11.75 & 5.43 & 14.78 & 9.05 \\
        GPT-5-mini & 3.19 & 8.68 & 7.12 & 14.95 & 8.17 \\
        GPT-5 & 3.58 & 8.02 & 8.24 & 14.17 & 8.26 \\
        Llama-4-Maverick & 6.88 & 13.80 & 9.99 & 14.75 & 11.81 \\
        Llama-4-Scout & 7.56 & 10.01 & 11.13 & 14.69 & 10.64 \\
        Qwen2.5-VL-7B & 2.05 & 12.30 & 8.19 & 15.00 & 10.05 \\
        Qwen2.5-VL-32B & 5.12 & 11.66 & 6.75 & 14.86 & 9.51 \\
        Qwen2.5-VL-72B & 4.47 & 11.01 & 8.42 & 15.00 & 9.80 \\
        \addlinespace[2pt]
        \midrule
        \addlinespace[2pt]
        \multicolumn{6}{c}{\cellcolor[HTML]{EFEFEF} \textit{Force WM Use}} \\
        \midrule
        GPT-4o-mini & 12.52 & 14.65 & 14.52 & 15.00 & 14.43 \\
        GPT-4o & 6.73 & 13.45 & 10.17 & 15.00 & 11.74 \\
        GPT-5-mini & 7.44 & 13.47 & 14.23 & 14.77 & 13.31 \\
        GPT-5 & 5.62 & 10.76 & 12.60 & 14.77 & 11.34 \\
        Llama-4-Maverick & 4.83 & 14.52 & 11.95 & 15.00 & 12.70 \\
        Llama-4-Scout & 10.52 & 12.81 & 12.97 & 15.00 & 12.86 \\
        Qwen2.5-VL-7B & 8.97 & 14.08 & 13.73 & 15.00 & 13.56 \\
        Qwen2.5-VL-32B & 6.08 & 12.86 & 11.05 & 15.00 & 11.75 \\
        Qwen2.5-VL-72B & 6.00 & 12.85 & 12.34 & 15.00 & 12.22 \\
        \bottomrule
    \end{tabular}
    }
    \caption{Agent Task Average Steps: Comparison of average trajectory length between without WM access, with WM access, and forced WM use.}
    \label{tab:apdx_agent_avg_steps}
\end{table}

\begin{table}[t]
    \centering
    \setlength\tabcolsep{2pt}
    \setlength\extrarowheight{2pt}
    
    \tabcolsep=0.01\linewidth
    \resizebox{\linewidth}{!}{
    \begin{tabular}{l c c c c c}
        \toprule
        \textbf{Model} & \textbf{FrozenLake} & \textbf{Navigation} & \textbf{PrimitiveSkill} & \textbf{Sokoban} & \textbf{Overall} \\
        \addlinespace[2pt]
        \midrule
        \addlinespace[2pt]
        \multicolumn{6}{c}{\cellcolor[HTML]{EFEFEF} \textit{Without WM Access}} \\
        \midrule
        GPT-4o-mini & 0.6100 & 0.8807 & 0.9213 & 0.8323 & 0.8590 \\
        GPT-4o & 0.9458 & 0.5408 & 0.7103 & 0.5063 & 0.5854 \\
        GPT-5-mini & 0.9293 & 0.9824 & 0.6567 & 0.6038 & 0.8111 \\
        GPT-5 & 0.3487 & 0.9077 & 0.4955 & 0.1542 & 0.6064 \\
        Llama-4-Maverick & 0.7379 & 0.7346 & 0.4886 & 0.6406 & 0.6587 \\
        Llama-4-Scout & 0.8200 & 0.8296 & 0.5350 & 0.7822 & 0.7334 \\
        Qwen2.5-VL-7B & 0.9641 & 0.9989 & 0.7008 & 0.9472 & 0.9023 \\
        Qwen2.5-VL-32B & 0.8615 & 0.8606 & 0.7969 & 0.7469 & 0.8278 \\
        Qwen2.5-VL-72B & 0.9167 & 0.9951 & 0.7261 & 0.7906 & 0.9012 \\
        \addlinespace[2pt]
        \midrule
        \addlinespace[2pt]
        \multicolumn{6}{c}{\cellcolor[HTML]{EFEFEF} \textit{With WM Access}} \\
        \midrule
        GPT-4o-mini & 0.9604 & 0.8830 & 0.9681 & 0.9437 & 0.9207 \\
        GPT-4o & 0.9600 & 0.5135 & 0.7793 & 0.4101 & 0.5809 \\
        GPT-5-mini & 0.9804 & 0.9831 & 0.6414 & 0.5795 & 0.8025 \\
        GPT-5 & 0.5240 & 0.9176 & 0.4060 & 0.2086 & 0.5969 \\
        Llama-4-Maverick & 0.7136 & 0.8198 & 0.7567 & 0.6547 & 0.7748 \\
        Llama-4-Scout & 0.8822 & 0.8588 & 0.7393 & 0.8245 & 0.8091 \\
        Qwen2.5-VL-7B & 0.9771 & 0.9932 & 0.7586 & 0.8917 & 0.9072 \\
        Qwen2.5-VL-32B & 0.9238 & 0.8544 & 0.8617 & 0.7518 & 0.8448 \\
        Qwen2.5-VL-72B & 0.9336 & 0.9867 & 0.6967 & 0.8510 & 0.8717 \\
        \addlinespace[2pt]
        \midrule
        \addlinespace[2pt]
        \multicolumn{6}{c}{\cellcolor[HTML]{EFEFEF} \textit{Force WM Use}} \\
        \midrule
        GPT-4o-mini & 0.7853 & 1.0000 & 0.9962 & 0.9354 & 0.9749 \\
        GPT-4o & 0.7610 & 0.8193 & 0.9777 & 0.4229 & 0.8201 \\
        GPT-5-mini & 0.9643 & 0.9948 & 0.6930 & 0.8157 & 0.8539 \\
        GPT-5 & 0.5389 & 0.9392 & 0.4224 & 0.2286 & 0.6193 \\
        Llama-4-Maverick & 0.8285 & 0.9224 & 0.8437 & 0.7625 & 0.8737 \\
        Llama-4-Scout & 0.9212 & 0.9472 & 0.8639 & 0.9104 & 0.9097 \\
        Qwen2.5-VL-7B & 0.9634 & 0.8935 & 0.8583 & 0.9792 & 0.8934 \\
        Qwen2.5-VL-32B & 0.7918 & 0.9181 & 0.9233 & 0.8000 & 0.8997 \\
        Qwen2.5-VL-72B & 0.9792 & 0.9933 & 0.9693 & 0.9333 & 0.9767 \\
        \bottomrule
    \end{tabular}
    }
    \caption{Agent Task Action Validity Rate: Percentage of valid actions across all steps in different modes.}
    \label{tab:apdx_agent_action_validity}
\end{table}

\begin{table}[t]
    \centering
    \setlength\tabcolsep{2pt}
    \setlength\extrarowheight{2pt}
    
    \tabcolsep=0.018\linewidth
    \resizebox{\linewidth}{!}{
    \begin{tabular}{l c c c c c}
        \toprule
        \textbf{Model} & \textbf{3DSRBench} & \textbf{MMSI} & \textbf{SAT} & \textbf{Spatial} & \textbf{Overall} \\
        \addlinespace[2pt]
        \midrule
        \addlinespace[2pt]
        \multicolumn{6}{c}{\cellcolor[HTML]{EFEFEF} \textit{Average WM Calls}} \\
        \midrule
        GPT-4o-mini & 0.0939 & 0.4280 & 0.2600 & 0.0122 & 0.1203 \\
        GPT-4o & 0.0014 & 0.0098 & 0.0133 & 0.0000 & 0.0023 \\
        GPT-5-mini & 0.0105 & 0.0087 & 0.0133 & 0.0032 & 0.0089 \\
        GPT-5 & 0.0000 & 0.0000 & 0.0000 & 0.0000 & 0.0000 \\
        Llama-4-Maverick & 0.4007 & 0.8234 & 0.4600 & 0.0746 & 0.3868 \\
        Llama-4-Scout & 0.4144 & 0.5612 & 0.3733 & 0.1690 & 0.3820 \\
        Qwen2.5-VL-7B & 0.0047 & 0.0054 & 0.0467 & 0.0039 & 0.0054 \\
        Qwen2.5-VL-32B & 0.0080 & 0.0033 & 0.0067 & 0.0013 & 0.0060 \\
        Qwen2.5-VL-72B & 0.0128 & 0.0618 & 0.0467 & 0.0000 & 0.0167 \\
        \addlinespace[2pt]
        \midrule
        \addlinespace[2pt]
        \multicolumn{6}{c}{\cellcolor[HTML]{EFEFEF} \textit{WM Usage Rate}} \\
        \midrule
        GPT-4o-mini & 0.0818 & 0.3012 & 0.2533 & 0.0122 & 0.0972 \\
        GPT-4o & 0.0014 & 0.0087 & 0.0133 & 0.0000 & 0.0022 \\
        GPT-5-mini & 0.0103 & 0.0087 & 0.0133 & 0.0032 & 0.0087 \\
        GPT-5 & 0.0000 & 0.0000 & 0.0000 & 0.0000 & 0.0000 \\
        Llama-4-Maverick & 0.3863 & 0.7465 & 0.4533 & 0.0733 & 0.3678 \\
        Llama-4-Scout & 0.3923 & 0.5320 & 0.3733 & 0.1690 & 0.3639 \\
        Qwen2.5-VL-7B & 0.0047 & 0.0054 & 0.0467 & 0.0039 & 0.0054 \\
        Qwen2.5-VL-32B & 0.0080 & 0.0033 & 0.0067 & 0.0013 & 0.0060 \\
        Qwen2.5-VL-72B & 0.0105 & 0.0358 & 0.0467 & 0.0000 & 0.0121 \\
        \bottomrule
    \end{tabular}
    }
    \caption{VQA Task WM Statistics (With WM Access): Upper section shows average WM calls per question. Lower section shows WM usage rate (percentage of questions that used WM at least once).}
    \label{tab:apdx_vqa_wm_stats}
\end{table}

\begin{table}[t]
    \centering
    \setlength\tabcolsep{2pt}
    \setlength\extrarowheight{2pt}
    
    \tabcolsep=0.018\linewidth
    \resizebox{\linewidth}{!}{
    \begin{tabular}{l c c c c c}
        \toprule
        \textbf{Model} & \textbf{3DSRBench} & \textbf{MMSI} & \textbf{SAT} & \textbf{Spatial} & \textbf{Overall} \\
        \addlinespace[2pt]
        \midrule
        \addlinespace[2pt]
        \multicolumn{6}{c}{\cellcolor[HTML]{EFEFEF} \textit{Average WM Calls}} \\
        \midrule
        GPT-4o-mini & 1.0112 & 1.3066 & 1.0000 & 1.0096 & 1.0457 \\
        GPT-4o & 0.9998 & 1.1051 & 1.0000 & 1.0000 & 1.0123 \\
        GPT-5-mini & 1.0522 & 1.1408 & 1.0933 & 1.0617 & 1.0654 \\
        GPT-5 & 1.0017 & 0.8407 & 1.0000 & 1.0006 & 0.9824 \\
        Llama-4-Maverick & 1.1388 & 1.4605 & 0.8400 & 1.0315 & 1.1498 \\
        Llama-4-Scout & 1.0281 & 1.1105 & 0.7867 & 1.0019 & 1.0280 \\
        Qwen2.5-VL-7B & 1.0731 & 1.2568 & 1.0067 & 1.0263 & 1.0843 \\
        Qwen2.5-VL-32B & 1.0037 & 1.0303 & 0.6933 & 1.0019 & 1.0005 \\
        Qwen2.5-VL-72B & 1.4188 & 1.5764 & 0.7133 & 1.0405 & 1.3483 \\
        \addlinespace[2pt]
        \midrule
        \addlinespace[2pt]
        \multicolumn{6}{c}{\cellcolor[HTML]{EFEFEF} \textit{WM Usage Rate}} \\
        \midrule
        GPT-4o-mini & 0.9998 & 1.0000 & 1.0000 & 1.0000 & 0.9999 \\
        GPT-4o & 0.9998 & 1.0000 & 1.0000 & 1.0000 & 0.9999 \\
        GPT-5-mini & 0.9996 & 0.9902 & 1.0000 & 1.0000 & 0.9986 \\
        GPT-5 & 0.9998 & 0.8267 & 1.0000 & 1.0000 & 0.9793 \\
        Llama-4-Maverick & 0.9981 & 1.0000 & 0.7600 & 0.9968 & 0.9934 \\
        Llama-4-Scout & 0.9994 & 1.0000 & 0.7600 & 0.9974 & 0.9945 \\
        Qwen2.5-VL-7B & 0.9961 & 0.9772 & 0.9933 & 0.9955 & 0.9937 \\
        Qwen2.5-VL-32B & 1.0000 & 1.0000 & 0.6933 & 1.0000 & 0.9941 \\
        Qwen2.5-VL-72B & 1.0000 & 0.9946 & 0.6933 & 0.9974 & 0.9929 \\
        \bottomrule
    \end{tabular}
    }
    \caption{VQA Task WM Statistics (Force WM Use): Upper section shows average WM calls per question when forced to use WM. Lower section shows WM usage rate (expected to be 100\% in forced mode).}
    \label{tab:apdx_vqa_wm_stats_force}
\end{table}

\paragraph{Additional Main Experiment Statistics.}
In \Cref{tab:apdx_agent_wm_stats} and \Cref{tab:apdx_agent_wm_stats_force}, we report the frequency and rate of world model usage for each agent task across all evaluated models, under both the normal setting and the forced world model usage setting. We observe that explicitly forcing world model usage in the system prompt leads to a higher absolute number of world model invocations per episode. However, even under this enforced setting, some models do not consistently comply. Notably, for GPT-5, the world model usage rate still fails to reach 1.0, indicating that the model does not use the world model at least once in every episode, despite explicit instructions to do so before each action.

We present analogous statistics for VQA tasks in \Cref{tab:apdx_vqa_wm_stats} and \Cref{tab:apdx_vqa_wm_stats_force}, where we observe similar overall trends. An exception arises in SAT tasks, for which models appear particularly reluctant to use the world model as a tool, even under the forced usage setting.

Additionally, in \Cref{tab:apdx_agent_avg_steps} and \Cref{tab:apdx_agent_action_validity}, we report statistics on the average number of steps per episode and the action validity rate for agent tasks. Given that the maximum number of steps per episode is capped at 15, we find that many models frequently reach this limit across all data points for certain tasks. This behavior may lead to repetitive actions, a phenomenon we analyze in detail in the attribution analysis presented in the main text. Further, we observe no meaningful differences in either step count or action validity rate across different evaluation modes. This suggests that these factors are not primary contributors to the observed performance differences, but instead serve to further corroborate the conclusions drawn in our attribution analysis.

\begin{table*}[t]
    \centering
    \small
    \setlength\tabcolsep{2pt}
    \setlength\extrarowheight{2pt}
    
    \begin{tabular}{l c c c c | c c c c | c c c c | c c c c}
        \toprule
        \textbf{Model} & \multicolumn{4}{c}{\textbf{FrozenLake}} & \multicolumn{4}{c}{\textbf{Navigation}} & \multicolumn{4}{c}{\textbf{PrimitiveSkill}} & \multicolumn{4}{c}{\textbf{Sokoban}} \\
         & \rotatebox{90}{\scriptsize Both Correct} & \rotatebox{90}{\scriptsize WM Helps} & \rotatebox{90}{\scriptsize WM Hurts} & \rotatebox{90}{\scriptsize Both Wrong} & \rotatebox{90}{\scriptsize Both Correct} & \rotatebox{90}{\scriptsize WM Helps} & \rotatebox{90}{\scriptsize WM Hurts} & \rotatebox{90}{\scriptsize Both Wrong} & \rotatebox{90}{\scriptsize Both Correct} & \rotatebox{90}{\scriptsize WM Helps} & \rotatebox{90}{\scriptsize WM Hurts} & \rotatebox{90}{\scriptsize Both Wrong} & \rotatebox{90}{\scriptsize Both Correct} & \rotatebox{90}{\scriptsize WM Helps} & \rotatebox{90}{\scriptsize WM Hurts} & \rotatebox{90}{\scriptsize Both Wrong} \\
        \midrule
        GPT-4o-mini & 17.2 & 21.9 & 18.8 & 42.2 & 11.7 & 12.3 & 14.7 & 61.3 & 15.2 & 6.6 & 16.4 & 61.7 & 0.0 & 0.0 & 0.0 & 100.0 \\
        GPT-4o & 32.8 & 25.0 & 25.0 & 17.2 & 18.0 & 12.7 & 17.3 & 52.0 & 32.8 & 12.9 & 18.4 & 35.9 & 0.0 & 1.6 & 1.6 & 96.9 \\
        GPT-5-mini & 76.6 & 12.5 & 9.4 & 1.6 & 50.3 & 13.0 & 15.3 & 21.3 & 3.5 & 15.2 & 7.0 & 74.2 & 0.0 & 0.0 & 3.1 & 96.9 \\
        GPT-5 & 70.3 & 18.8 & 6.2 & 4.7 & 61.7 & 9.3 & 12.3 & 16.7 & 13.3 & 6.2 & 5.5 & 75.0 & 1.6 & 12.5 & 4.7 & 81.2 \\
        Llama-4-Maverick & 50.0 & 15.6 & 20.3 & 14.1 & 10.7 & 9.3 & 20.7 & 59.3 & 16.8 & 15.2 & 23.0 & 44.9 & 0.0 & 3.1 & 0.0 & 96.9 \\
        Llama-4-Scout & 31.2 & 23.4 & 28.1 & 17.2 & 40.0 & 14.0 & 14.7 & 31.3 & 16.9 & 7.1 & 15.3 & 60.8 & 0.0 & 3.1 & 1.6 & 95.3 \\
        Qwen2.5-VL-7B & 17.2 & 28.1 & 18.8 & 35.9 & 12.7 & 13.0 & 13.7 & 60.7 & 7.0 & 5.1 & 5.9 & 82.0 & 0.0 & 0.0 & 1.6 & 98.4 \\
        Qwen2.5-VL-32B & 35.9 & 21.9 & 23.4 & 18.8 & 17.7 & 14.7 & 18.3 & 49.3 & 28.5 & 10.2 & 20.3 & 41.0 & 0.0 & 1.6 & 0.0 & 98.4 \\
        Qwen2.5-VL-72B & 26.6 & 18.8 & 34.4 & 20.3 & 22.0 & 15.3 & 15.7 & 47.0 & 22.7 & 9.4 & 18.0 & 50.0 & 0.0 & 0.0 & 0.0 & 100.0 \\
        \bottomrule
    \end{tabular}
    \caption{Agent Task Performance Comparison (WM Invisible vs Normal): For each model and task, percentage of episodes in four categories: Both Correct (correct in both without and with WM modes), WM Helps (wrong without WM, correct with WM), WM Hurts (correct without WM, wrong with WM), and Both Wrong (wrong in both modes). Percentages sum to 100\% for each model-task combination.}
    \label{tab:apdx_agent_comparative}
\end{table*}

\begin{table*}[t]
    \centering
    \small
    \setlength\tabcolsep{2pt}
    \setlength\extrarowheight{2pt}
    
    \begin{tabular}{l c c c c | c c c c | c c c c | c c c c}
        \toprule
        \textbf{Model} & \multicolumn{4}{c}{\textbf{3DSRBench}} & \multicolumn{4}{c}{\textbf{MMSI}} & \multicolumn{4}{c}{\textbf{SAT}} & \multicolumn{4}{c}{\textbf{Spatial}} \\
         & \rotatebox{90}{\scriptsize Both Correct} & \rotatebox{90}{\scriptsize WM Helps} & \rotatebox{90}{\scriptsize WM Hurts} & \rotatebox{90}{\scriptsize Both Wrong} & \rotatebox{90}{\scriptsize Both Correct} & \rotatebox{90}{\scriptsize WM Helps} & \rotatebox{90}{\scriptsize WM Hurts} & \rotatebox{90}{\scriptsize Both Wrong} & \rotatebox{90}{\scriptsize Both Correct} & \rotatebox{90}{\scriptsize WM Helps} & \rotatebox{90}{\scriptsize WM Hurts} & \rotatebox{90}{\scriptsize Both Wrong} & \rotatebox{90}{\scriptsize Both Correct} & \rotatebox{90}{\scriptsize WM Helps} & \rotatebox{90}{\scriptsize WM Hurts} & \rotatebox{90}{\scriptsize Both Wrong} \\
        \midrule
        GPT-4o-mini & 49.8 & 9.1 & 8.5 & 32.6 & 17.7 & 8.9 & 10.2 & 63.3 & 42.0 & 14.7 & 10.0 & 33.3 & 57.7 & 8.0 & 7.7 & 26.5 \\
        GPT-4o & 59.1 & 6.4 & 6.9 & 27.5 & 20.9 & 8.9 & 9.8 & 60.5 & 64.7 & 8.7 & 6.0 & 20.7 & 67.3 & 4.8 & 4.2 & 23.7 \\
        GPT-5-mini & 60.1 & 7.5 & 6.9 & 25.4 & 25.9 & 10.5 & 8.7 & 54.9 & 80.0 & 3.3 & 4.7 & 12.0 & 74.2 & 4.8 & 3.7 & 17.3 \\
        GPT-5 & 62.9 & 6.9 & 6.5 & 23.7 & 29.5 & 7.4 & 8.9 & 54.3 & 82.0 & 3.3 & 4.0 & 10.7 & 75.1 & 4.0 & 4.4 & 16.4 \\
        Llama-4-Maverick & 50.5 & 11.6 & 10.8 & 27.1 & 15.7 & 12.8 & 11.4 & 60.1 & 38.7 & 8.0 & 13.3 & 40.0 & 67.4 & 7.2 & 7.1 & 18.3 \\
        Llama-4-Scout & 47.3 & 11.6 & 11.3 & 29.8 & 13.8 & 14.6 & 13.1 & 58.5 & 27.3 & 8.7 & 13.3 & 50.7 & 66.5 & 6.2 & 7.7 & 19.7 \\
        Qwen2.5-VL-7B & 39.5 & 14.1 & 13.6 & 32.8 & 11.3 & 12.8 & 13.0 & 62.9 & 48.7 & 17.3 & 10.7 & 23.3 & 50.8 & 11.8 & 10.7 & 26.7 \\
        Qwen2.5-VL-32B & 47.9 & 10.3 & 11.2 & 30.7 & 18.1 & 10.1 & 11.6 & 60.2 & 39.3 & 7.3 & 8.0 & 45.3 & 58.9 & 8.3 & 7.6 & 25.1 \\
        Qwen2.5-VL-72B & 52.8 & 8.1 & 8.0 & 31.0 & 18.3 & 11.2 & 10.8 & 59.7 & 43.3 & 4.7 & 3.3 & 48.7 & 65.4 & 7.5 & 5.7 & 21.5 \\
        \bottomrule
    \end{tabular}
    \caption{VQA Task Performance Comparison (WM Invisible vs Normal): For each model and task, percentage of questions in four categories: Both Correct (correct in both without and with WM modes), WM Helps (wrong without WM, correct with WM), WM Hurts (correct without WM, wrong with WM), and Both Wrong (wrong in both modes). Percentages sum to 100\% for each model-task combination.}
    \label{tab:apdx_vqa_comparative}
\end{table*}

\begin{table*}[t]
    \centering
    \small
    \setlength\tabcolsep{2pt}
    \setlength\extrarowheight{2pt}
    
    \begin{tabular}{l c c c c | c c c c | c c c c | c c c c}
        \toprule
        \textbf{Model} & \multicolumn{4}{c}{\textbf{FrozenLake}} & \multicolumn{4}{c}{\textbf{Navigation}} & \multicolumn{4}{c}{\textbf{PrimitiveSkill}} & \multicolumn{4}{c}{\textbf{Sokoban}} \\
         & \rotatebox{90}{\scriptsize Both Correct} & \rotatebox{90}{\scriptsize WM Helps} & \rotatebox{90}{\scriptsize WM Hurts} & \rotatebox{90}{\scriptsize Both Wrong} & \rotatebox{90}{\scriptsize Both Correct} & \rotatebox{90}{\scriptsize WM Helps} & \rotatebox{90}{\scriptsize WM Hurts} & \rotatebox{90}{\scriptsize Both Wrong} & \rotatebox{90}{\scriptsize Both Correct} & \rotatebox{90}{\scriptsize WM Helps} & \rotatebox{90}{\scriptsize WM Hurts} & \rotatebox{90}{\scriptsize Both Wrong} & \rotatebox{90}{\scriptsize Both Correct} & \rotatebox{90}{\scriptsize WM Helps} & \rotatebox{90}{\scriptsize WM Hurts} & \rotatebox{90}{\scriptsize Both Wrong} \\
        \midrule
        GPT-4o-mini & 3.1 & 3.1 & 32.8 & 60.9 & 3.0 & 2.0 & 23.3 & 71.7 & 2.3 & 0.8 & 29.3 & 67.6 & 0.0 & 0.0 & 0.0 & 100.0 \\
        GPT-4o & 34.4 & 23.4 & 23.4 & 18.8 & 12.3 & 9.0 & 23.0 & 55.7 & 30.1 & 10.2 & 21.1 & 38.7 & 0.0 & 0.0 & 1.6 & 98.4 \\
        GPT-5-mini & 57.8 & 9.4 & 28.1 & 4.7 & 14.0 & 6.7 & 51.7 & 27.7 & 0.0 & 2.0 & 10.5 & 87.5 & 0.0 & 1.6 & 3.1 & 95.3 \\
        GPT-5 & 71.9 & 18.8 & 4.7 & 4.7 & 50.7 & 7.3 & 23.3 & 18.7 & 7.4 & 2.0 & 11.3 & 79.3 & 1.6 & 4.7 & 4.7 & 89.1 \\
        Llama-4-Maverick & 42.2 & 21.9 & 28.1 & 7.8 & 4.0 & 8.3 & 27.3 & 60.3 & 15.6 & 15.6 & 24.2 & 44.5 & 0.0 & 0.0 & 0.0 & 100.0 \\
        Llama-4-Scout & 29.7 & 20.3 & 29.7 & 20.3 & 27.7 & 8.7 & 27.0 & 36.7 & 12.2 & 5.9 & 20.0 & 62.0 & 0.0 & 0.0 & 1.6 & 98.4 \\
        Qwen2.5-VL-7B & 7.8 & 17.2 & 28.1 & 46.9 & 4.7 & 7.0 & 21.7 & 66.7 & 0.4 & 0.8 & 12.5 & 86.3 & 0.0 & 0.0 & 1.6 & 98.4 \\
        Qwen2.5-VL-32B & 20.3 & 18.8 & 39.1 & 21.9 & 11.3 & 13.7 & 24.7 & 50.3 & 19.1 & 4.7 & 29.7 & 46.5 & 0.0 & 0.0 & 0.0 & 100.0 \\
        Qwen2.5-VL-72B & 34.4 & 18.8 & 26.6 & 20.3 & 18.7 & 8.7 & 19.0 & 53.7 & 15.6 & 3.9 & 25.0 & 55.5 & 0.0 & 0.0 & 0.0 & 100.0 \\
        \bottomrule
    \end{tabular}
    \caption{Agent Task Performance Comparison (WM Invisible vs WM Force): For each model and task, percentage of episodes in four categories when comparing without WM access to forced WM use: Both Correct (correct in both modes), WM Helps (wrong without WM, correct with forced WM), WM Hurts (correct without WM, wrong with forced WM), and Both Wrong (wrong in both modes). Percentages sum to 100\% for each model-task combination.}
    \label{tab:apdx_agent_comparative_wmforce}
\end{table*}

\begin{table*}[t]
    \centering
    \small
    \setlength\tabcolsep{2pt}
    \setlength\extrarowheight{2pt}
    
    \begin{tabular}{l c c c c | c c c c | c c c c | c c c c}
        \toprule
        \textbf{Model} & \multicolumn{4}{c}{\textbf{3DSRBench}} & \multicolumn{4}{c}{\textbf{MMSI}} & \multicolumn{4}{c}{\textbf{SAT}} & \multicolumn{4}{c}{\textbf{Spatial}} \\
         & \rotatebox{90}{\scriptsize Both Correct} & \rotatebox{90}{\scriptsize WM Helps} & \rotatebox{90}{\scriptsize WM Hurts} & \rotatebox{90}{\scriptsize Both Wrong} & \rotatebox{90}{\scriptsize Both Correct} & \rotatebox{90}{\scriptsize WM Helps} & \rotatebox{90}{\scriptsize WM Hurts} & \rotatebox{90}{\scriptsize Both Wrong} & \rotatebox{90}{\scriptsize Both Correct} & \rotatebox{90}{\scriptsize WM Helps} & \rotatebox{90}{\scriptsize WM Hurts} & \rotatebox{90}{\scriptsize Both Wrong} & \rotatebox{90}{\scriptsize Both Correct} & \rotatebox{90}{\scriptsize WM Helps} & \rotatebox{90}{\scriptsize WM Hurts} & \rotatebox{90}{\scriptsize Both Wrong} \\
        \midrule
        GPT-4o-mini & 48.4 & 10.8 & 9.9 & 31.0 & 16.6 & 11.2 & 11.3 & 61.0 & 39.3 & 21.3 & 12.7 & 26.7 & 54.9 & 9.0 & 10.5 & 25.6 \\
        GPT-4o & 57.1 & 8.8 & 9.0 & 25.2 & 18.1 & 10.9 & 12.6 & 58.4 & 56.0 & 11.3 & 14.7 & 18.0 & 65.2 & 7.1 & 6.4 & 21.3 \\
        GPT-5-mini & 60.1 & 7.3 & 6.9 & 25.6 & 24.1 & 10.4 & 10.5 & 55.0 & 76.0 & 5.3 & 8.7 & 10.0 & 74.2 & 5.7 & 3.7 & 16.5 \\
        GPT-5 & 62.9 & 6.5 & 6.4 & 24.1 & 25.9 & 6.6 & 12.5 & 55.0 & 81.3 & 4.0 & 4.7 & 10.0 & 75.1 & 4.2 & 4.4 & 16.2 \\
        Llama-4-Maverick & 49.6 & 11.5 & 11.8 & 27.2 & 14.6 & 15.7 & 12.5 & 57.2 & 38.7 & 11.3 & 13.3 & 36.7 & 65.3 & 7.3 & 9.2 & 18.3 \\
        Llama-4-Scout & 46.3 & 12.5 & 12.3 & 28.9 & 15.3 & 12.6 & 11.6 & 60.6 & 30.0 & 11.3 & 10.7 & 48.0 & 66.5 & 6.9 & 7.6 & 19.0 \\
        Qwen2.5-VL-7B & 34.0 & 16.7 & 19.2 & 30.2 & 9.4 & 12.2 & 14.8 & 63.5 & 25.3 & 11.3 & 34.0 & 29.3 & 40.9 & 13.8 & 20.6 & 24.7 \\
        Qwen2.5-VL-32B & 46.3 & 10.5 & 12.7 & 30.5 & 16.7 & 10.6 & 13.0 & 59.7 & 32.7 & 5.3 & 14.7 & 47.3 & 57.0 & 9.2 & 9.6 & 24.2 \\
        Qwen2.5-VL-72B & 49.5 & 9.6 & 11.3 & 29.6 & 14.0 & 13.7 & 15.2 & 57.2 & 41.3 & 7.3 & 5.3 & 46.0 & 62.8 & 8.2 & 8.2 & 20.8 \\
        \bottomrule
    \end{tabular}
    \caption{VQA Task Performance Comparison (WM Invisible vs WM Force): For each model and task, percentage of questions in four categories when comparing without WM access to forced WM use: Both Correct (correct in both modes), WM Helps (wrong without WM, correct with forced WM), WM Hurts (correct without WM, wrong with forced WM), and Both Wrong (wrong in both modes). Percentages sum to 100\% for each model-task combination.}
    \label{tab:apdx_vqa_comparative_wmforce}
\end{table*}

\paragraph{Additional Comparative Analysis.}
In \Cref{tab:apdx_agent_comparative}, \Cref{tab:apdx_vqa_comparative}, \Cref{tab:apdx_agent_comparative_wmforce}, and \Cref{tab:apdx_vqa_comparative_wmforce}, we present a detailed breakdown of the percentage of cases in which the world model either improves or degrades performance relative to the world model invisible setting. These comparisons are reported separately for agentic and VQA tasks, under both the normal world model access and forced world model usage configurations.

Across all evaluated tasks and models, we observe that the proportion of cases in which the world model improves performance does not consistently exceed the proportion of cases in which it degrades performance. This indicates that the effectiveness of world models is not reliably demonstrated across models or task settings. These statistics further support our main conclusions regarding the limited effectiveness of world models for current VLM-based agents and motivate our subsequent investigations into the specific conditions under which the world model helps or hurts performance, which we analyze through detailed attribution studies.

\section{Attribution Analysis Details}
\label{apdx:attribution_details}

We begin by manually analyzing 60 cases in which the world model either improves or degrades performance. Through detailed inspection, we identify and summarize the core categories that explain when and why the world model is effective or fails to assist the tested model. Based on this manual analysis, we curate a set of eight attribution categories respectively that capture the primary failure and success modes.

Following this initial human analysis, we employ GPT-4o to assist with large-scale attribution. For each case, GPT-4o is tasked with identifying one primary attribution reason and any number of additional contributing reasons. To perform this analysis, we provide GPT-4o with the interaction trajectory of the tested model when the world model is invisible, along with the corresponding trajectory for the same query when the world model is available (or enforced to use). GPT-4o is then instructed to compare these trajectories and assign attribution labels accordingly.

To assess the reliability of the automated annotations, we manually review all 60 cases that were previously examined and annotated by humans, comparing GPT-4o’s primary attribution labels with human judgments. The agreement rate for the primary attribution reaches 81.67\%, indicating strong alignment and suggesting that the model’s case-by-case annotations are reliable. In addition, we randomly sample 20 further cases for manual verification, finding that all attribution reasons produced by GPT-4o are reasonable. Based on these validation results, we report attribution statistics for all cases using the automated annotations.

\section{Additional Analysis Results}

In this section, we present more analysis results in addition to the analysis in the main text.

\begin{figure*}[t]
    \centering
    \includegraphics[width=0.90\linewidth]{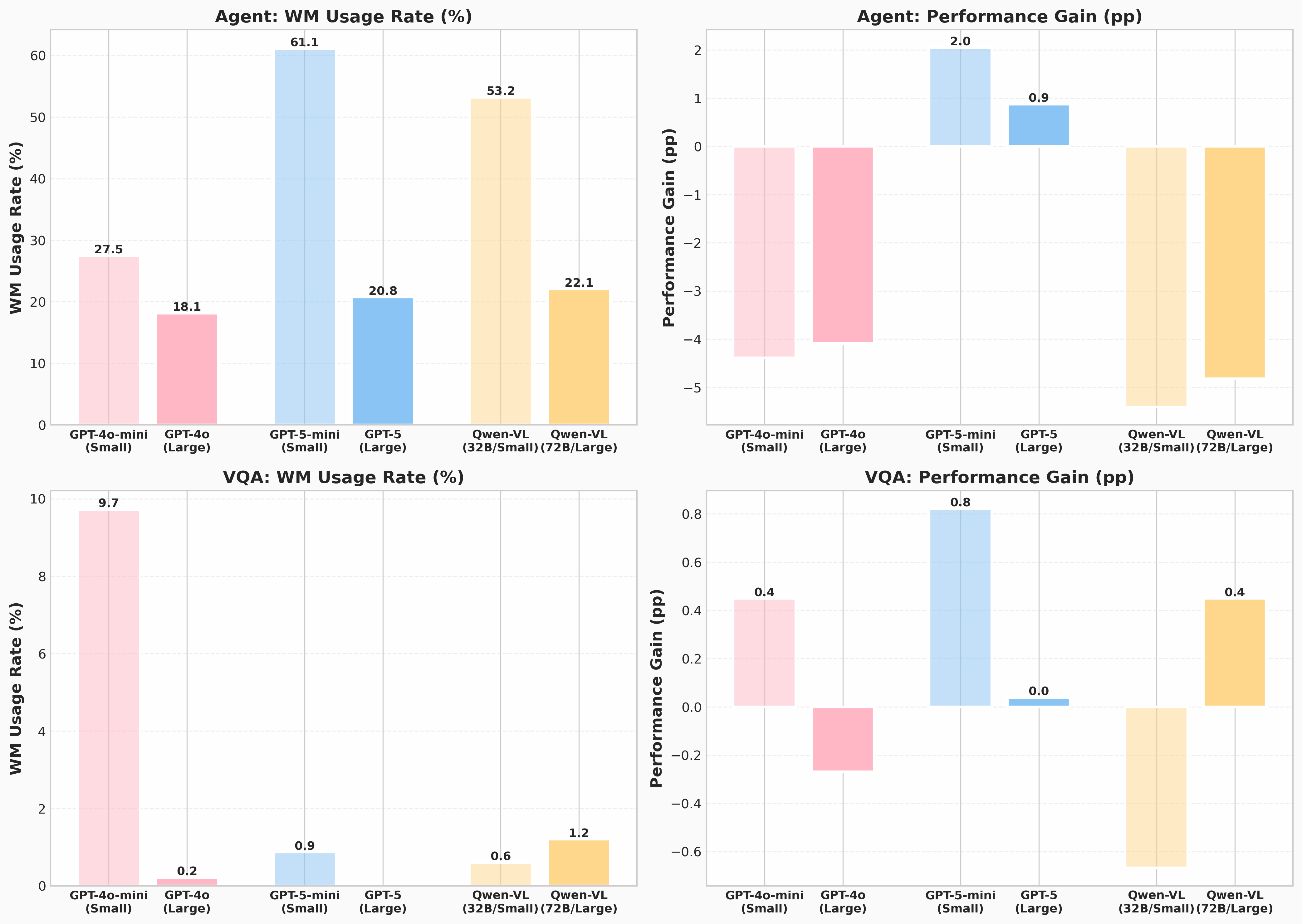}
    \caption{Comparison of world model usage rate and corresponding performance gain across model families and scales, for both agentic tasks (top row) and VQA tasks (bottom row). Results are aggregated by model family (GPT, Qwen) and size, highlighting systematic differences in invocation behavior and the limited, sometimes negative, returns of increased world model usage.}
    \label{fig:apdx_scaling_family_comparison}
\end{figure*}

In \Cref{fig:apdx_scaling_family_comparison}, we examine how world model usage and its effectiveness vary jointly with model family and scale, across both agent tasks and VQA settings. A clear pattern emerges that increased world model invocation does not consistently yield positive returns. For agent tasks, smaller models tend to rely more heavily on the world model and occasionally obtain modest gains, whereas larger models reduce their usage and often experience neutral or negative effects. Qwen models further illustrate that high usage alone is insufficient, as frequent but poorly calibrated world model calls correlate with pronounced performance degradation. In VQA tasks, usage rates remain uniformly low regardless of scale, and performance changes are marginal, reinforcing the conclusion that current models lack a principled mechanism for deciding when foresight is beneficial and how to integrate it effectively.

\begin{figure*}[t]
    \centering
    \includegraphics[width=0.90\linewidth]{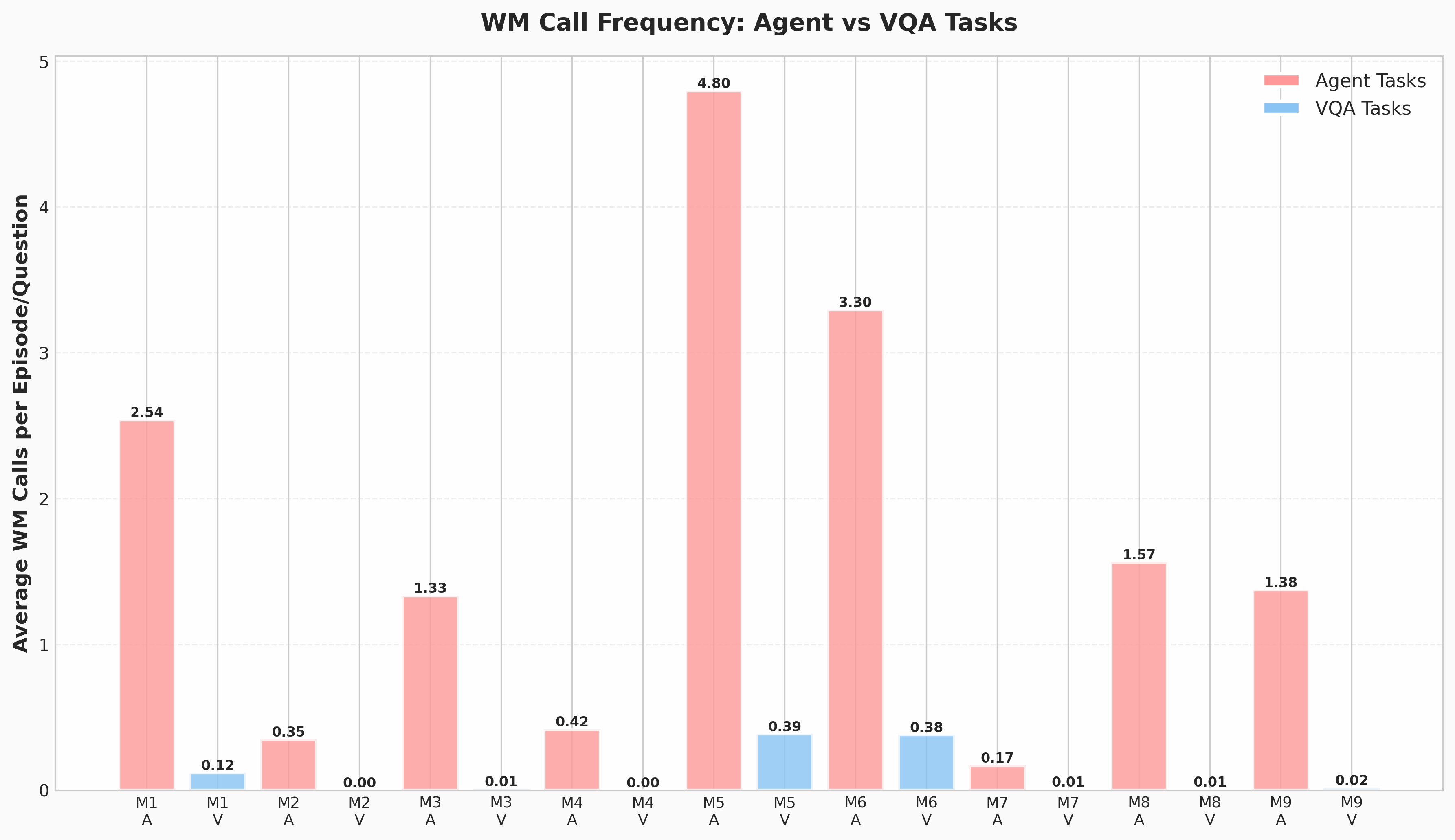}
    \caption{Average number of world model calls per episode or question across agent tasks and VQA tasks, aggregated by model. Each pair compares the same backbone evaluated in agent settings (A) versus VQA settings (V), illustrating systematic differences in how frequently models rely on simulation under interactive control versus purely perceptual reasoning.}
    \label{fig:apdx_task_wm_calls_comparison}
\end{figure*}

In \Cref{fig:apdx_task_wm_calls_comparison}, we compare how often models invoke the world model when solving agent tasks versus VQA tasks. Across all backbones, agentic settings induce substantially higher call frequencies, often by an order of magnitude, reflecting the natural role of foresight in sequential decision-making. In contrast, VQA tasks trigger extremely sparse usage, with most models making fewer than one call per question on average. This disparity aligns with the main text’s analysis that simulation is more naturally integrated into dynamic, stateful environments, whereas in VQA it requires highly targeted and well-specified queries that current models rarely generate. Importantly, the elevated call frequency in agent tasks does not imply better outcomes, reinforcing our broader finding that repeated world model invocation often signals uncertainty rather than effective integration of foresight.

\begin{figure*}[t]
    \centering
    \includegraphics[width=0.85\linewidth]{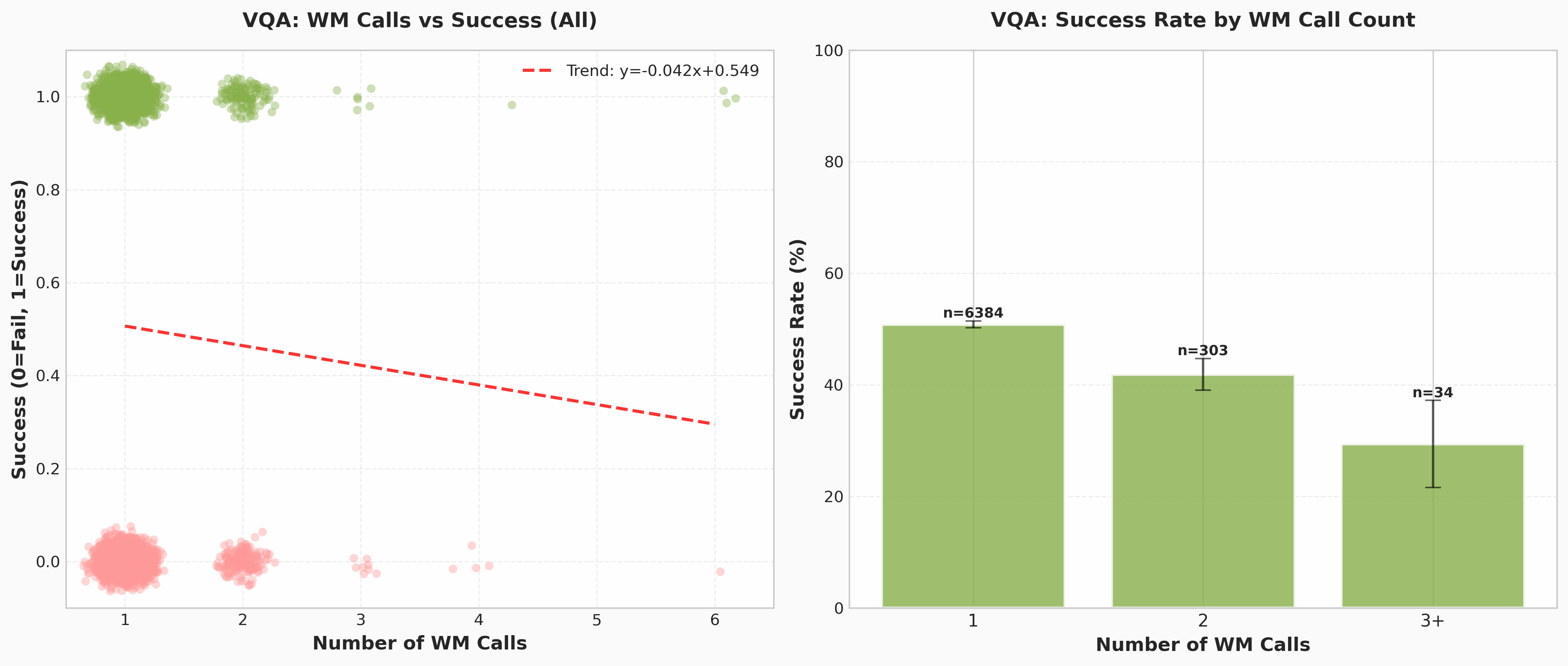}
    \caption{Relationship between world model call frequency and answer correctness for VQA tasks. \emph{Left}: per-question scatter plot of success (1) or failure (0) versus the number of world model calls, with a fitted linear trend. \emph{Right}: success rate grouped by world model call count, with error bars indicating standard error and $n$ denoting the number of questions in each bin.}
    \label{fig:apdx_wm_effectiveness_vqa_all}
\end{figure*}

\begin{figure*}[t]
    \centering
    \includegraphics[width=0.85\linewidth]{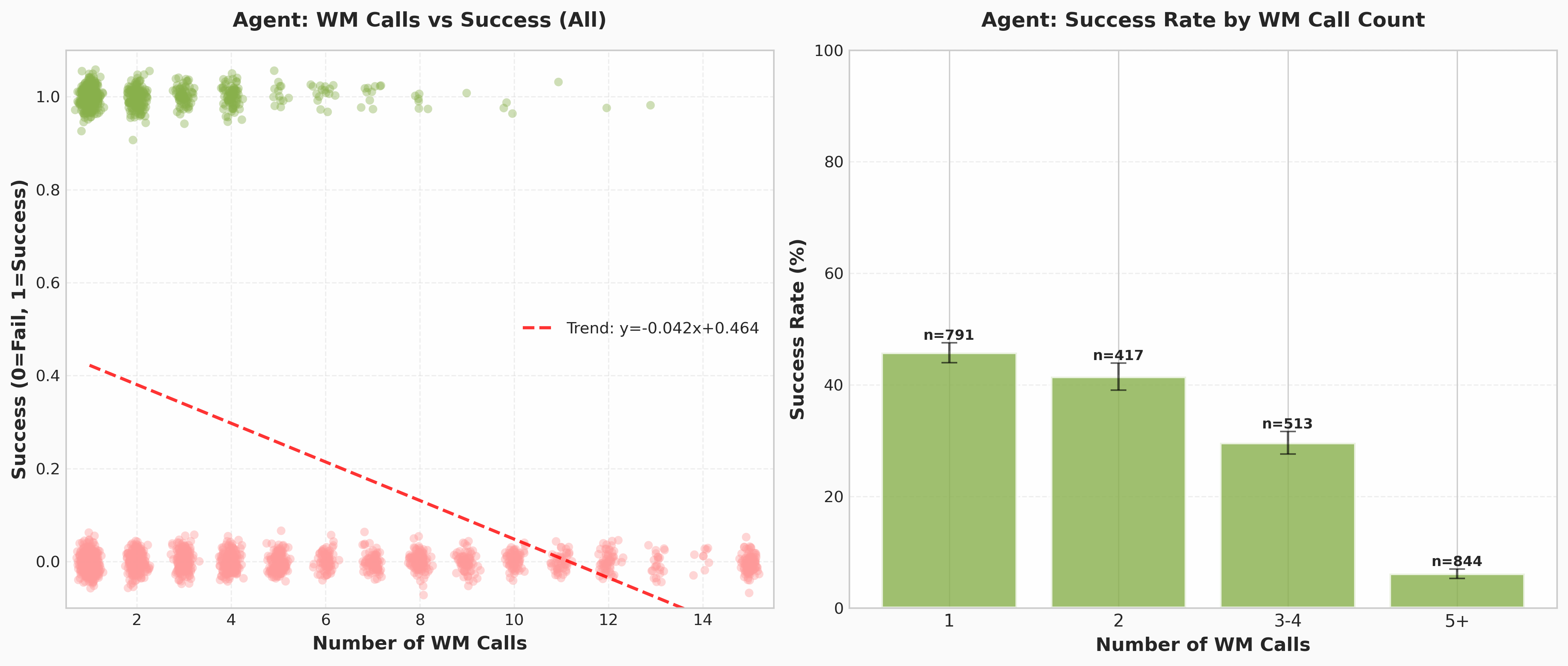}
    \caption{Relationship between world model call frequency and task success for agentic tasks. \emph{Left}: per-episode success as a function of world model call count with a fitted trend line. \emph{Right}: success rate aggregated by world model call frequency bins, showing a monotonic decline as the number of calls increases.}
    \label{fig:apdx_wm_effectiveness_agent_all}
\end{figure*}

In \Cref{fig:apdx_wm_effectiveness_agent_all} and \Cref{fig:apdx_wm_effectiveness_vqa_all}, we jointly analyze how the frequency of world model invocation correlates with performance in agent and VQA tasks. In both settings, higher world model call counts are associated with lower success rates, as evidenced by negative linear trends and consistent degradation in the binned statistics. The effect is more pronounced in agent tasks, where excessive calls coincide with sharp performance drops, reflecting action loops and over-planning behaviors discussed in the main text. In VQA, although overall call counts are much lower, repeated invocations still correspond to reduced accuracy, suggesting that additional simulations often amplify ambiguity rather than resolve it. Together, these results reinforce the central conclusion that world model usage in current systems lacks effective governance: repeated querying signals uncertainty and poor integration of foresight rather than accumulating reliable evidence.

\begin{figure*}[t]
    \centering
    \includegraphics[width=0.85\linewidth]{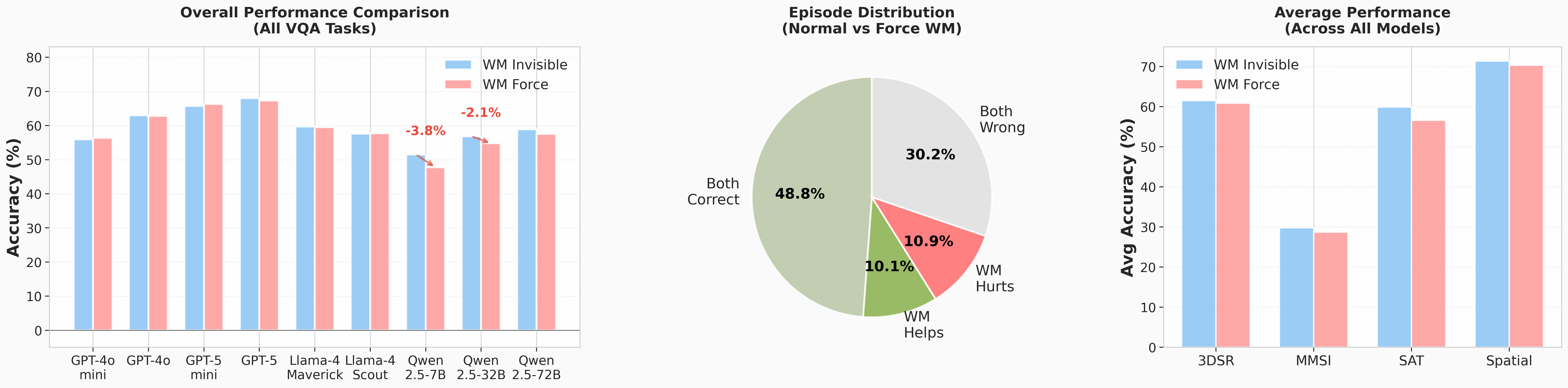}
    \caption{Effect of forcing world model usage on VQA performance. \emph{Left}: per-model accuracy comparison between WM Invisible and WM Force settings across all VQA tasks. \emph{Middle}: episode-level outcome distribution categorized into cases where world model helps, hurts, both correct, or both wrong. \emph{Right}: average accuracy aggregated by VQA benchmark, showing the net impact of mandatory world model invocation across datasets.}
    \label{fig:apdx_force_wm_vqa}
\end{figure*}

In \Cref{fig:apdx_force_wm_vqa}, we analyze the impact of forcing world model usage in VQA tasks, mirroring the forced-usage study in the main text for agent task settings. Across models and benchmarks, mandatory world model invocation consistently fails to improve performance and often leads to modest but systematic degradation. The episode-level breakdown shows that cases where forcing world model use actively hurts outnumber those where it helps, while nearly half of the questions remain unaffected, indicating limited corrective value. Dataset-level averages further confirm that forcing simulation does not benefit VQA reasoning, reinforcing the main text’s conclusion that, without a principled mechanism for deciding when and how to simulate, compulsory world model usage amplifies confusion rather than resolving ambiguity.

\section{Use of LLMs}
In this work, LLMs are used strictly for research support rather than as sources of substantive content. Their use falls into: (i) serving as the tested model or world model, and (ii) assisting with language refinement during paper writing. For writing support, we used GPT-5 solely to polish text (improving coherence and grammar) while all ideas, logic, results, and technical contributions originate from the authors. To safeguard rigor, we have carefully reviewed all LLM-refined texts to confirm that no hallucinated content was introduced and that the original arguments, findings, and perspectives were faithfully preserved.

\clearpage
\section{Evaluation Instruction}
In this section, we present all prompts used for the evaluation. Under each mode and for each task, the prompt set consists of three templates: the system prompt, the initial task prompt, and the feedback prompt.

Within these templates, several placeholders are defined. The \textit{max actions} placeholder specifies the maximum number of actions the agent may take in a single run; for all agentic tasks, this value is fixed to~3. The \textit{action separation mark} is consistently set to the comma symbol ``',''. The \textit{state} placeholder corresponds to positions where images are inserted in the prompt, including interpolated images produced during simulation as well as the task’s initial state image. 

All remaining placeholders are associated with individual data points of a task. For example, the \textit{instruction} field describes the specific final goal for a given instance, while in VQA tasks, the \textit{question} and \textit{options} fields are instantiated uniquely for each data point. These fields are therefore populated independently according to the underlying dataset.

In the following, we detail the complete prompt sets used for the three modes of our evaluation.

\subsection{Normal Mode}
\begin{promptbox}[title=FrozenLake Task System Instruction,colback=orange!5,colframe=orange!80!black]
## Goal
You are a FrozenLake solver. Your objective is to **reach the gift box goal** while **avoiding holes hidden in the ice**. At the start of every turn you will receive a top-down image of the lake. You must decide what actions to take or whether to simulate actions using the world model tool.

---

## Environment Layout
- Agent: green mini-figure.
- Goal: wrapped gift box tile.
- Hole: blue cracked tile that ends the episode if you fall in.
- Safe ice: white tiles you can step on.

---

## Action Rules
- Allowed actions: `Up`, `Down`, `Left`, `Right`.
- You may output up to **{max_actions}** action(s) per turn, separated by **`{action_sep}`**.
- Actions execute sequentially in the order you provide them.
- Plan for slip: the agent may continue sliding past the intended tile.

---

## Tool Usage: World Model Simulation
You can call the world model tool to preview how a sequence of moves might unfold.
- Input: current state image + the actions you want to simulate.
- Output: predicted frames showing how the agent moves.
- Use the simulation to validate hypotheses; it is predictive, not authoritative.

**When calling the tool, describe clearly** what you observe in the current image (agent position, goal position, holes) and what sequence of moves you want to simulate. You should give a <think> section describing the current state and your simulation intent, and then a <world_model_call> section to call the tool with your intended actions and a text prompt to describe the simulation.

### Format when using the world model:
```
<think> <observation>Describe the agent, goal, and holes as you currently see them.</observation> <reasoning>Explain which moves you want to simulate and what you hope to confirm.</reasoning> </think>

<world_model_call>
Action: The actions you want to simulate, separated by {action_sep}.
Prompt: Simulated FrozenLake style. The agent performs: [describe actions clearly]. Keep the camera and the background fixed.
</world_model_call>
```

**Example:**
```
<think> <observation>The agent starts at the top-left. A hole is one tile right; the goal is two tiles right and one down.</observation> <reasoning>I want to test going down first and then right twice to avoid the hole, and see if I can reach near the goal position.</reasoning> </think>

<world_model_call>
Action: Down{action_sep}Right{action_sep}Right
Prompt: Simulated FrozenLake style. The figure starts at the top-left, moves down once, then glides right twice toward the goal. Keep the camera and the background fixed.
</world_model_call>
```

---

## Output When Taking Real Actions
After reasoning (and optionally simulating), decide on your actual move sequence. You should give a <think> section to describe the current state and the reasonings behind your intended actions, and then an <answer> section to output the actions you want to take.

### Format when taking real actions:
```
<think> <observation>Summarize the current layout, and any updates from simulations.</observation> <reasoning>Explain why your chosen actions should reach the goal or make progress while staying safe.</reasoning> </think>

<answer>[Your actions separated by {action_sep}]</answer>
```

**Example:**
```
<think> <observation>The agent is one tile left of the goal with no adjacent holes.</observation> <reasoning>A single move to the right should slide directly onto the goal.</reasoning> </think>

<answer>Right</answer>
```

---

## Guidelines for Your Reasoning
- Track relative positions of yourself, holes, and the goal before each move.
- World model may help you disambiguate risky moves or multi-step plans.
- Use the world model when uncertain about a move's effect.
- The world model's output is predictive, not authoritative.
- When giving actions and using the world model, **output only** in the required structured format.
\end{promptbox}

\begin{promptbox}[title=FrozenLake Task Initial Instruction,colback=orange!5,colframe=orange!80!black]
The initial FrozenLake state is:
{state}
You can provide up to {max_actions} action(s) separated by '{action_sep}', or you can use the world model tool to predict the outcome of your actions before you actually perform them. Please output in the required structured format.
\end{promptbox}

\begin{promptbox}[title=FrozenLake Task Feedback Instruction,colback=orange!5,colframe=orange!80!black]
Previous valid action(s) that you have taken: {actions}. This led to the current FrozenLake state:
{state}
You can provide up to {max_actions} action(s) separated by '{action_sep}', or you can use the world model tool to predict the outcome of your actions before you actually perform them. Please output in the required structured format.
\end{promptbox}

% ==================================================== %
% ==================================================== %

\begin{promptbox}[title=Navigation Task System Instruction,colback=green!5,colframe=green!80!black]
## Goal
You are a 3D navigation agent. Your objective is to **reach the goal location** (or **approach it as closely as possible**) while **avoiding obstacles**. You must decide what actions to take or whether to simulate actions using the world model tool.

---

## 3D Environment
- First-person viewpoint inside an indoor scene.
- **Goal**: You will be given an instruction about your goal in your initial observation, which is a target somewhere in the environment.
- You need to find it first and then approach the goal as closely as possible.
- **Obstacles**: walls, furniture, or other impassable geometry.
- You can move in all four directions, rotate your view, or look up and down.

---

## Action Rules
- Allowed actions: `moveahead`, `moveback`, `moveright`, `moveleft`, `rotateright`, `rotateleft`, `lookup`, `lookdown`.
- Movement actions translate by 0.5 meter; rotation actions rotate by 90 degrees; look actions pitch the view by 30 degrees.
- Each turn you may output up to **{max_actions}** action(s), separated by **`{action_sep}`**.
- Actions execute sequentially in the order provided.

---

## Tool Usage: World Model Simulation
Use the world model tool to simulate how a sequence of actions changes your view.
- Input: current camera frame + the action sequence you want to test.
- Output: interpolated frames predicting the resulting viewpoint.
- Treat the prediction as guidance; reality may differ.

**When calling the tool, describe clearly** what you observe in the current image and what action sequence you want to simulate. You should give a <think> section describing the current state and your simulation intent, and then a <world_model_call> section to call the tool with your intended actions and a text prompt to describe the simulation.

### Format when using the world model:
```
<think> <observation>Describe what you currently see, especially goal cues and obstacles.</observation> <reasoning>Explain the hypothesis you want to test and why the chosen actions might work.</reasoning> </think>

<world_model_call>
Action: Simulated actions separated by {action_sep}.
Prompt: Simulated indoor 3D environment style. Describe the intended motion and keep the camera stable.
</world_model_call>
```

**Example:**
```
<think> <observation>I am standing in the room facing a kitchen table. There's a microwave on the table and a fridge in the corner on my left. There are also some chairs beside the table. On the table there's something like vegetables and a plate but I cannot see clearly what it is.</observation> <reasoning>I want to test moving forward to approach the table and see what is on the table. I am given the instruction to reach close to the bread. Usually it should appear on the table.</reasoning> </think>

<world_model_call>
Action: moveahead{action_sep}moveahead
Prompt: Simulated indoor 3D environment style. You walk forward twice along the corridor. Your camera shows the first person view of the environment. You should keep the camera stable and smooth.
</world_model_call>
```

---

## Output When Taking Real Actions
After reasoning (and optionally simulating) decide on your actual actions. You should give a <think> section to describe the current state and the reasonings behind your intended actions, and then an <answer> section to output the actions you want to take.

### Format when taking real actions:
```
<think> <observation>Summarize the scene and any updates from simulations.</observation> <reasoning>Explain why your chosen actions can help you reach the final goal position as closely as possible.</reasoning> </think>

<answer>[Your actions separated by {action_sep}]</answer>
```

**Example:**
```
<think> <observation>I am currently facing a fridge and is very close to it. It nearly takes half of my view. The only thing on the left I can see is a half of a television screen.</observation> <reasoning>I am instructed to find the remote control, which should be near the television. However, from the current view I cannot see any other objects. Therefore I need to rotate left towards the television to see if the remote control is there.</reasoning> </think>

<answer>rotateleft{action_sep}moveahead</answer>
```

---

## Guidelines for Your Reasoning
- Track your orientation after every rotation to avoid disorientation.
- Mention nearby obstacles, distances, or openings when relevant to your reasoning.
- The world model can help you validate blind corners or multi-move plans.
- Use the world model when uncertain about a move's effect.
- The world model's output is predictive, not authoritative.
- When giving actions and using the world model, **output only** in the required structured format.
\end{promptbox}

\begin{promptbox}[title=Navigation Task Initial Instruction,colback=green!5,colframe=green!80!black]
Your initial observation in the environment is:
{state}
This is the goal of your task: {instruction}
You can provide up to {max_actions} action(s) separated by '{action_sep}', or you can use the world model tool to predict the outcome of your actions before you actually perform them. Please output in the required structured format.
\end{promptbox}

\begin{promptbox}[title=Navigation Task Feedback Instruction,colback=green!5,colframe=green!80!black]
Previous valid action(s) that you have taken: {actions}. This led to the current navigation state:
{state}
This is the goal of your task: {instruction}
You can provide up to {max_actions} action(s) separated by '{action_sep}', or you can use the world model tool to predict the outcome of your actions before you actually perform them. Please output in the required structured format.
\end{promptbox}

% ==================================================== %
% ==================================================== %

\begin{promptbox}[title=Primitive Skill Task System Instruction,colback=blue!5,colframe=blue!80!black]
## Goal
You are controlling a Franka Emika robot arm. Your objective is to manipulate objects (cubes, triangles, etc.) to complete tasks based on human instructions.
At each turn, you will receive an **image** of the robot workspace, an **instruction**, **object positions**, and **workspace limits**. You must decide what actions to take or whether to simulate actions using the world model tool.

---

## Workspace and Coordinate System
- **Viewing direction**: You're facing toward the negative x-axis.
- **Coordinate frame**: 
  - Negative x-axis is to your front
  - Positive x-axis is to your back
  - Negative y-axis is to your left
  - Positive y-axis is to your right
  - Positive z-axis is up
- **Units**: All coordinates (x, y, z) are in **millimeters** and are **integers**.
- **Workspace limits**: Will be provided (x_workspace_limit, y_workspace_limit, z_workspace_limit).

---

## Action Rules
Allowed actions:
1. **pick(x, y, z)** - Grasp an object at position (x, y, z)
2. **place(x, y, z)** - Place the held object at position (x, y, z)
3. **push(x1, y1, z1, x2, y2, z2)** - Push an object from (x1, y1, z1) to (x2, y2, z2)

Important notes:
- Coordinates must be within workspace limits
- Positions refer to the **center** of objects
- When placing, consider object volume - it's safe to set z **much higher** to avoid collisions
- Object positions will be provided, but you must **match them to objects in the image** (which might need world model simulation to help you)
- Each turn, you may output up to **{max_actions}** actions, separated by **`{action_sep}`**

---

## Tool Usage: World Model Simulation
You can call a **world model tool** to predict what happens when you take certain actions before committing to them.
- Input: current workspace image + your described robot manipulation actions
- Output: predicted frames showing simulated outcomes of the actions
- Use this to verify object positions, test action sequences, or explore alternatives

**When calling the tool, describe clearly** what you observe in the current image (robot gripper state, object positions) and what manipulation sequence you want to simulate. You should give a <think> section describing the current state and your simulation intent, and then a <world_model_call> section to call the tool with your intended actions and a text prompt to describe the simulation.

### Format when using the world model:
```
<think> <observation>Describe the robot arm position, gripper state, and object positions as seen in the image.</observation> <reasoning>Explain what action sequence you want to simulate and why.</reasoning> </think>

<world_model_call>
Action: The actions you want to simulate, separated by {action_sep}.
Prompt: Simulated robot arm manipulation style. The robot arm performs: [describe actions clearly]. Show the gripper and objects moving accordingly.
</world_model_call>
```

**Example:**
```
<think> <observation>The robot gripper is empty and positioned near the workspace center. I see two cubes on the desk, one green and one red.</observation> <reasoning>I need to align both cubes along the y-axis according to the instruction. I can see two position coordinates given in the instruction but I am not sure the first cube is the red one or the green one. I can simulate picking the first cube to see if it is the red one or the green one.</reasoning> </think>

<world_model_call>
Action: pick(62,-55,20)
Prompt: Simulated robot arm manipulation style. The robot arm picks up the cube from position (62,-55,20). The camera and background are fixed.
</world_model_call>
```

---

## Output When Taking Real Actions
After reasoning (and optionally simulating), decide what actions to actually execute. You should give a <think> section to describe the current state and the reasonings behind your intended actions, and then an <answer> section to output the actions you want to take.

### Format when taking real actions:
```
<think> <observation>Describe object positions, robot state, and task goal based on the image, and any updates from simulations.</observation> <reasoning>Explain how your chosen actions will accomplish the task. Include spatial reasoning about coordinates and object matching.</reasoning> </think>

<answer>[Your actions separated by {action_sep}]</answer>
```

**Example:**
```
<think> <observation>There are two cubes on the desk and according to previous simulation I can see the first position coordinates (62,-55,20) is the red cube, so the second position coordinates (75,33,20) shuold be the green cube. In the current state the red cube is placed on the left of the green cube.</observation> <reasoning>In order to align both cubes along the y-axis according to the instruction, I need to pick and then place the red cube to a position where the x-coordinate is 0. I can keep all other coordinates the same, but change the x-coordinate to 0 for simplicity. This will lead the red cube to be placed at (0,-55,20). I will do the same for the green cube later and make sure they are separate and does not collide with each other.</reasoning> </think>

<answer>pick(62,-55,20){action_sep}place(0,-55,20)</answer>
```

---

## Guidelines for Your Reasoning
- Match provided positions to objects in the image using the coordinate system, visual cues and world model simulation
- Be explicit about spatial reasoning (which object is where, based on coordinates and image)
- When stacking or placing, set z higher to avoid collisions
- Ensure all coordinates are within workspace limits
- Use the world model when uncertain about action outcomes or object positions
- The world model's output is predictive, not authoritative
- When giving actions and using the world model, **output only** in the required structured format
\end{promptbox}

\begin{promptbox}[title=Primitive Skill Task Initial Instruction,colback=blue!5,colframe=blue!80!black]
The initial robot workspace state is:
{state}
Human Instruction: {instruction}
x_workspace_limit: {x_workspace}
y_workspace_limit: {y_workspace}
z_workspace_limit: {z_workspace}
Object positions:
{object_positions}
Other information:
{other_information}
You can provide up to {max_actions} action(s) separated by '{action_sep}', or you can use the world model tool to predict the outcome of your actions before you actually perform them. Please output in the required structured format.
\end{promptbox}

\begin{promptbox}[title=Primitive Skill Task Feedback Instruction,colback=blue!5,colframe=blue!80!black]
Previous valid action(s) that you have taken: {actions}. This lead to the current robot workspace state:
{state}
Human Instruction: {instruction}
x_workspace_limit: {x_workspace}
y_workspace_limit: {y_workspace}
z_workspace_limit: {z_workspace}
Object positions:
{object_positions}
Other information:
{other_information}
You can provide up to {max_actions} action(s) separated by '{action_sep}', or you can use the world model tool to predict the outcome of your actions before you actually perform them. Please output in the required structured format.
\end{promptbox}

% ==================================================== %
% ==================================================== %

\begin{promptbox}[title=Sokoban Task System Instruction,colback=red!5,colframe=red!80!black]
## Goal
You are a Sokoban solver. Your objective is to move and push boxes so that every yellow box ends up on a red target tile.  
At each turn, you will receive a **current game state** (image). You must decide what actions to take or whether to simulate actions using the world model tool.

---

## Game Elements (as seen in the image)
- You: green mini-figure
- Box: yellow tile marked with a red cross
- Target: red box position marked with a red dot
- Wall: solid blocks (impassable)
- Empty space: black or open area (walkable)

---

## Action Rules
- Allowed actions: `Up`, `Down`, `Left`, `Right`.
- You can move freely on open tiles.
- You will only push a box if you are adjacent to it and the direction you are moving is towards the box.
- Each turn, you may output up to **{max_actions}** actions, separated by **`{action_sep}`**.

---

## Tool Usage: World Model Simulation
You can call a **world model tool** to predict what happens when you take certain actions before committing to them.
- Input: current state image + your described intended actions.  
- Output: predicted frames showing simulated outcomes.  
- Use this to verify hypotheses, not as ground truth.

**When calling the tool, describe clearly** what the current state looks like and what sequence of moves to simulate. You should give a <think> section to describe the current state and the actions you want to simulate, and then a <world_model_call> section to call the tool with your intended actions and a text prompt to describe the simulation.

### Format when using the world model:
```
<think> <observation>Describe the player, boxes, and target positions as seen in the image.</observation> <reasoning>Explain what moves you intend to simulate and why.</reasoning> </think>

<world_model_call>
Action: The actions you want to simulate, separated by {action_sep}.
Prompt: Please generate in Sokoban style. The green player performs the following moves: [describe actions clearly]. Keep camera and background fixed.
</world_model_call>
```

**Example:**
```
<think> <observation>The player stands left of a box. The box is two tiles away from the target on the right.</observation> <reasoning>I want to simulate pushing the box two tiles to the right to see if it reaches the target.</reasoning> </think>

<world_model_call>
Action: Right{action_sep}Right
Prompt: Please generate in Sokoban style. The green player pushes the yellow box two tiles to the right. Keep camera and background fixed.
</world_model_call>
```

---

## Output When Taking Real Actions
After reasoning (and optionally simulating), decide what actions to actually take. You should give a <think> section to describe the current state and the reasonings behind your intended actions, and then a <answer> section to output the actions you want to take.

### Format when taking real actions:
```
<think> <observation>Summarize the player, box, and target positions, and any updates from simulations.</observation> <reasoning>Explain how your actions will move boxes toward targets.</reasoning> </think>

<answer>[Your actions separated by {action_sep}]</answer>
```

**Example:**
```
<think> <observation>The player is left of a box; the box is one step above the target.</observation> <reasoning>I can push the box down to place it on the target. In order to push the box down, I need to move to the top of the box first, and then push the box down. I need to go one step up and then one step right to be on the top of the box first, and then push the target down.</reasoning> </think>

<answer>Up{action_sep}Right{action_sep}Down</answer>
```

---

## Guidelines for Your Reasoning
- Be concise but spatially explicit when describing positions.  
- Always ensure planned moves obey Sokoban's physical rules.  
- Use the world model when uncertain about a move's effect.  
- The world model's output is predictive, not authoritative.  
- When giving actions and using the world model, **output only** in the required structured format.
\end{promptbox}

\begin{promptbox}[title=Sokoban Task Initial Instruction,colback=red!5,colframe=red!80!black]
The initial Sokoban board state is:
{state}
You can provide up to {max_actions} action(s) separated by '{action_sep}', or you can use the world model tool to predict the outcome of your actions before you actually perform them. Please output in the required structured format.
\end{promptbox}

\begin{promptbox}[title=Sokoban Task Feedback Instruction,colback=red!5,colframe=red!80!black]
Previous valid action(s) that you have taken: {actions}. This lead to the current Sokoban board state:
{state}
You can provide up to {max_actions} action(s) separated by '{action_sep}', or you can use the world model tool to predict the outcome of your actions before you actually perform them. Please output in the required structured format.
\end{promptbox}

% ==================================================== %
% ==================================================== %

\begin{promptbox}[title=VQA Task System Instruction,colback=purple!5,colframe=purple!80!black]
## Goal
You are a Visual Question Answering (VQA) assistant. Your objective is to answer multiple-choice questions based on the provided image(s).
You will initially receive **one or more images** along with a **question** and **multiple choice options**. You must analyze the images and decide whether to use the world model tool to simulate visual transformations or commit your answer.

---

## Question Format
- You will see a series of images followed by a question.
- The images given to you each has a label (e.g. Image 1, Image 2, etc.).
- Each question has **multiple choice options** (A, B, C, D, etc.).
- **You can only answer ONCE per question**. Once you provide an answer, the question is complete.

---

## Tool Usage: World Model Simulation
You can call a **world model tool** to simulate visual transformations or explore "what if" scenarios before committing to an answer.
- Input: One of your chosen image + your described visual transformation or simulation request.
- Output: predicted image(s) showing simulated outcomes.
- Use this to verify hypotheses about spatial relationships, object properties, or visual patterns.

**When calling the tool, describe clearly** what you observe in the image and what visual transformation or analysis you want to simulate. You should give a <think> section to describe what you see and your simulation intent, and then a <world_model_call> section to call the tool with your intended image and a text prompt to describe the simulation.

### Format when using the world model:
```
<think> <observation>Describe what you see in the image(s): objects, colors, positions, counts, etc.</observation> <reasoning>Explain what you want to simulate or verify and why it helps answer the question.</reasoning> </think>

<world_model_call>
Action: Give the label of the image you want to simulate on, in the format of "Image <number>".
Prompt: Describe the visual simulation or transformation you want to see. Be specific about what should change or what to analyze.
</world_model_call>
```

**Example:**
```
<think> <observation> I can see in the first image the chair is in the middle of the image, and in the second image the chair is moved to the left corner of the image. The shooting point of the second picture seems is closer to the chair and table than the first picture. </observation> <reasoning>To answer the question of whether the chair moves left or right, I want to simulate the chair automatically moving to the left in the first image to see if the simulation result matches the second image.</reasoning> </think>

<world_model_call>
Action: Image 1
Prompt: Please generate in simulated environment style. The chair moves slightly left away from the table. All other objects in the image should be kept the same and not moved. The camera goes a little bit closer to the chair and table.
</world_model_call>
```

---

## Output When Providing Your Answer
After reasoning (and optionally simulating), decide what your final answer is. You should give a <think> section to describe your observations and the reasonings behind your intended answer, and then an <answer> section to output the letter of your chosen option.

### Format:
```
<think> <observation>Summarize what you see in the image(s) relevant to the question, and any updates from simulations.</observation> <reasoning>Explain how your observations lead to your answer choice.</reasoning> </think>

<answer>[The letter of your chosen option (A, B, C, D, etc.)]</answer>
```

**Example:**
```
<think> <observation>There is one giraffe sitting on the ground and two zebras standing next to it. Behind them, there are more gireffes standing in a row, eating the leaves from the trees.</observation> <reasoning>The question asks how many giraffes are sitting on the ground. Even though there are more giraffes in the background, there is only one giraffe sitting on the ground, which matches the answer option C.</reasoning> </think>

<answer>C</answer>
```

---

## Guidelines for Your Reasoning
- Be precise when describing what you observe in each image.
- Count carefully when dealing with quantity questions.
- Compare images systematically when looking for patterns or differences.
- Use the world model when uncertain about visual details or counts.
- The world model's output is predictive, not authoritative.
- When giving answers and using the world model, **output only** in the required structured format.
\end{promptbox}

\begin{promptbox}[title=VQA Task Initial Instruction,colback=purple!5,colframe=purple!80!black]
The question is:
{question}

Options:
{options}

Images:
{images}

You can use the world model tool to simulate visual transformations or verify your observations before answering, or you can directly provide your answer. Please output in the required structured format.
\end{promptbox}

\begin{promptbox}[title=VQA Task Feedback Instruction,colback=purple!5,colframe=purple!80!black]
Your answer was: {answer}

Result: {result}

{feedback}
\end{promptbox}

% ==================================================== %
% ==================================================== %

\subsection{World Model Invisible Mode}
\begin{promptbox}[title=FrozenLake Task System Instruction,colback=orange!5,colframe=orange!80!black]
## Goal
You are a FrozenLake solver. Your objective is to **reach the gift box goal** while **avoiding holes hidden in the ice**. At the start of every turn you will receive a top-down image of the lake. You must decide what actions to take.

---

## Environment Layout
- Agent: green mini-figure.
- Goal: wrapped gift box tile.
- Hole: blue cracked tile that ends the episode if you fall in.
- Safe ice: white tiles you can step on.

---

## Action Rules
- Allowed actions: `Up`, `Down`, `Left`, `Right`.
- You may output up to **{max_actions}** action(s) per turn, separated by **`{action_sep}`**.
- Actions execute sequentially in the order you provide them.
- Plan for slip: the agent may continue sliding past the intended tile.

---

## Output When Taking Actions
When deciding on your move sequence, you should give a <think> section to describe the current state and the reasonings behind your intended actions, and then an <answer> section to output the actions you want to take.

### Format when taking real actions:
```
<think> <observation>Summarize the current layout.</observation> <reasoning>Explain why your chosen actions should reach the goal or make progress while staying safe.</reasoning> </think>

<answer>[Your actions separated by {action_sep}]</answer>
```

**Example:**
```
<think> <observation>The agent is one tile left of the goal with no adjacent holes.</observation> <reasoning>A single move to the right should slide directly onto the goal.</reasoning> </think>

<answer>Right</answer>
```

---

## Guidelines for Your Reasoning
- Track relative positions of yourself, holes, and the goal before each move.
- When giving actions, **output only** in the required structured format.
\end{promptbox}

\begin{promptbox}[title=FrozenLake Task Initial Instruction,colback=orange!5,colframe=orange!80!black]
The initial FrozenLake state is:
{state}
You can provide up to {max_actions} action(s) separated by '{action_sep}'. Please output in the required structured format.
\end{promptbox}

\begin{promptbox}[title=FrozenLake Task Feedback Instruction,colback=orange!5,colframe=orange!80!black]
Previous valid action(s) that you have taken: {actions}. This led to the current FrozenLake state:
{state}
You can provide up to {max_actions} action(s) separated by '{action_sep}'. Please output in the required structured format.
\end{promptbox}

% ==================================================== %
% ==================================================== %

\begin{promptbox}[title=Navigation Task System Instruction,colback=green!5,colframe=green!80!black]
## Goal
You are a 3D navigation agent. Your objective is to **reach the goal location** (or **approach it as closely as possible**) while **avoiding obstacles**. You must decide what actions to take.

---

## 3D Environment
- First-person viewpoint inside an indoor scene.
- **Goal**: You will be given an instruction about your goal in your initial observation, which is a target somewhere in the environment.
- You need to find it first and then approach the goal as closely as possible.
- **Obstacles**: walls, furniture, or other impassable geometry.
- You can move in all four directions, rotate your view, or look up and down.

---

## Action Rules
- Allowed actions: `moveahead`, `moveback`, `moveright`, `moveleft`, `rotateright`, `rotateleft`, `lookup`, `lookdown`.
- Movement actions translate by 0.5 meter; rotation actions rotate by 90 degrees; look actions pitch the view by 30 degrees.
- Each turn you may output up to **{max_actions}** action(s), separated by **`{action_sep}`**.
- Actions execute sequentially in the order provided.

---

## Output When Taking Actions
When deciding on your actual actions, you should give a <think> section to describe the current state and the reasonings behind your intended actions, and then an <answer> section to output the actions you want to take.

### Format when taking real actions:
```
<think> <observation>Summarize the scene.</observation> <reasoning>Explain why your chosen actions can help you reach the final goal position as closely as possible.</reasoning> </think>

<answer>[Your actions separated by {action_sep}]</answer>
```

**Example:**
```
<think> <observation>I am currently facing a fridge and is very close to it. It nearly takes half of my view. The only thing on the left I can see is a half of a television screen.</observation> <reasoning>I am instructed to find the remote control, which should be near the television. However, from the current view I cannot see any other objects. Therefore I need to rotate left towards the television to see if the remote control is there.</reasoning> </think>

<answer>rotateleft{action_sep}moveahead</answer>
```

---

## Guidelines for Your Reasoning
- Track your orientation after every rotation to avoid disorientation.
- Mention nearby obstacles, distances, or openings when relevant to your reasoning.
- When giving actions, **output only** in the required structured format.
\end{promptbox}

\begin{promptbox}[title=Navigation Task Initial Instruction,colback=green!5,colframe=green!80!black]
Your initial observation in the environment is:
{state}
This is the goal of your task: {instruction}
You can provide up to {max_actions} action(s) separated by '{action_sep}'. Please output in the required structured format.
\end{promptbox}

\begin{promptbox}[title=Navigation Task Feedback Instruction,colback=green!5,colframe=green!80!black]
Previous valid action(s) that you have taken: {actions}. This led to the current navigation state:
{state}
This is the goal of your task: {instruction}
You can provide up to {max_actions} action(s) separated by '{action_sep}'. Please output in the required structured format.
\end{promptbox}

% ==================================================== %
% ==================================================== %

\begin{promptbox}[title=Primitive Skill Task System Instruction,colback=blue!5,colframe=blue!80!black]
## Goal
You are controlling a Franka Emika robot arm. Your objective is to manipulate objects (cubes, triangles, etc.) to complete tasks based on human instructions.
At each turn, you will receive an **image** of the robot workspace, an **instruction**, **object positions**, and **workspace limits**. You must decide what actions to take.

---

## Workspace and Coordinate System
- **Viewing direction**: You're facing toward the negative x-axis.
- **Coordinate frame**:
  - Negative x-axis is to your front
  - Positive x-axis is to your back
  - Negative y-axis is to your left
  - Positive y-axis is to your right
  - Positive z-axis is up
- **Units**: All coordinates (x, y, z) are in **millimeters** and are **integers**.
- **Workspace limits**: Will be provided (x_workspace_limit, y_workspace_limit, z_workspace_limit).

---

## Action Rules
Allowed actions:
1. **pick(x, y, z)** - Grasp an object at position (x, y, z)
2. **place(x, y, z)** - Place the held object at position (x, y, z)
3. **push(x1, y1, z1, x2, y2, z2)** - Push an object from (x1, y1, z1) to (x2, y2, z2)

Important notes:
- Coordinates must be within workspace limits
- Positions refer to the **center** of objects
- When placing, consider object volume - it's safe to set z **much higher** to avoid collisions
- Object positions will be provided, and you must **match them to objects in the image**
- Each turn, you may output up to **{max_actions}** actions, separated by **`{action_sep}`**

---

## Output When Taking Actions
When deciding what actions to actually execute, you should give a <think> section to describe the current state and the reasonings behind your intended actions, and then an <answer> section to output the actions you want to take.

### Format when taking real actions:
```
<think> <observation>Describe object positions, robot state, and task goal based on the image.</observation> <reasoning>Explain how your chosen actions will accomplish the task. Include spatial reasoning about coordinates and object matching.</reasoning> </think>

<answer>[Your actions separated by {action_sep}]</answer>
```

**Example:**
```
<think> <observation>There are two cubes on the desk. The first position coordinates (62,-55,20) is the red cube, so the second position coordinates (75,33,20) should be the green cube. In the current state the red cube is placed on the left of the green cube.</observation> <reasoning>In order to align both cubes along the y-axis according to the instruction, I need to pick and then place the red cube to a position where the x-coordinate is 0. I can keep all other coordinates the same, but change the x-coordinate to 0 for simplicity. This will lead the red cube to be placed at (0,-55,20). I will do the same for the green cube later and make sure they are separate and does not collide with each other.</reasoning> </think>

<answer>pick(62,-55,20){action_sep}place(0,-55,20)</answer>
```

---

## Guidelines for Your Reasoning
- Match provided positions to objects in the image using the coordinate system and visual cues
- Be explicit about spatial reasoning (which object is where, based on coordinates and image)
- When stacking or placing, set z higher to avoid collisions
- Ensure all coordinates are within workspace limits
- When giving actions, **output only** in the required structured format
\end{promptbox}

\begin{promptbox}[title=Primitive Skill Task Initial Instruction,colback=blue!5,colframe=blue!80!black]
The initial robot workspace state is:
{state}
Human Instruction: {instruction}
x_workspace_limit: {x_workspace}
y_workspace_limit: {y_workspace}
z_workspace_limit: {z_workspace}
Object positions:
{object_positions}
Other information:
{other_information}
You can provide up to {max_actions} action(s) separated by '{action_sep}'. Please output in the required structured format.
\end{promptbox}

\begin{promptbox}[title=Primitive Skill Task Feedback Instruction,colback=blue!5,colframe=blue!80!black]
Previous valid action(s) that you have taken: {actions}. This lead to the current robot workspace state:
{state}
Human Instruction: {instruction}
x_workspace_limit: {x_workspace}
y_workspace_limit: {y_workspace}
z_workspace_limit: {z_workspace}
Object positions:
{object_positions}
Other information:
{other_information}
You can provide up to {max_actions} action(s) separated by '{action_sep}'. Please output in the required structured format.
\end{promptbox}

% ==================================================== %
% ==================================================== %

\begin{promptbox}[title=Sokoban Task System Instruction,colback=red!5,colframe=red!80!black]
## Goal
You are a Sokoban solver. Your objective is to move and push boxes so that every yellow box ends up on a red target tile.
At each turn, you will receive a **current game state** (image). You must decide what actions to take.

---

## Game Elements (as seen in the image)
- You: green mini-figure
- Box: yellow tile marked with a red cross
- Target: red box position marked with a red dot
- Wall: solid blocks (impassable)
- Empty space: black or open area (walkable)

---

## Action Rules
- Allowed actions: `Up`, `Down`, `Left`, `Right`.
- You can move freely on open tiles.
- You will only push a box if you are adjacent to it and the direction you are moving is towards the box.
- Each turn, you may output up to **{max_actions}** actions, separated by **`{action_sep}`**.

---

## Output When Taking Actions
When deciding what actions to actually take, you should give a <think> section to describe the current state and the reasonings behind your intended actions, and then a <answer> section to output the actions you want to take.

### Format when taking real actions:
```
<think> <observation>Summarize the player, box, and target positions.</observation> <reasoning>Explain how your actions will move boxes toward targets.</reasoning> </think>

<answer>[Your actions separated by {action_sep}]</answer>
```

**Example:**
```
<think> <observation>The player is left of a box; the box is one step above the target.</observation> <reasoning>I can push the box down to place it on the target. In order to push the box down, I need to move to the top of the box first, and then push the box down. I need to go one step up and then one step right to be on the top of the box first, and then push the target down.</reasoning> </think>

<answer>Up{action_sep}Right{action_sep}Down</answer>
```

---

## Guidelines for Your Reasoning
- Be concise but spatially explicit when describing positions.
- Always ensure planned moves obey Sokoban's physical rules.
- When giving actions, **output only** in the required structured format.
\end{promptbox}

\begin{promptbox}[title=Sokoban Task Initial Instruction,colback=red!5,colframe=red!80!black]
The initial Sokoban board state is:
{state}
You can provide up to {max_actions} action(s) separated by '{action_sep}'. Please output in the required structured format.
\end{promptbox}

\begin{promptbox}[title=Sokoban Task Feedback Instruction,colback=red!5,colframe=red!80!black]
Previous valid action(s) that you have taken: {actions}. This lead to the current Sokoban board state:
{state}
You can provide up to {max_actions} action(s) separated by '{action_sep}'. Please output in the required structured format.
\end{promptbox}

% ==================================================== %
% ==================================================== %

\begin{promptbox}[title=VQA Task System Instruction,colback=purple!5,colframe=purple!80!black]
## Goal
You are a Visual Question Answering (VQA) assistant. Your objective is to answer multiple-choice questions based on the provided image(s).
You will initially receive **one or more images** along with a **question** and **multiple choice options**. You must analyze the images and provide your answer.

---

## Question Format
- You will see a series of images followed by a question.
- The images given to you each has a label (e.g. Image 1, Image 2, etc.).
- Each question has **multiple choice options** (A, B, C, D, etc.).
- **You can only answer ONCE per question**. Once you provide an answer, the question is complete.

---

## Output When Providing Your Answer
When deciding on your final answer, you should give a <think> section to describe your observations and the reasonings behind your intended answer, and then an <answer> section to output the letter of your chosen option.

### Format:
```
<think> <observation>Summarize what you see in the image(s) relevant to the question.</observation> <reasoning>Explain how your observations lead to your answer choice.</reasoning> </think>

<answer>[The letter of your chosen option (A, B, C, D, etc.)]</answer>
```

**Example:**
```
<think> <observation>There is one giraffe sitting on the ground and two zebras standing next to it. Behind them, there are more gireffes standing in a row, eating the leaves from the trees.</observation> <reasoning>The question asks how many giraffes are sitting on the ground. Even though there are more giraffes in the background, there is only one giraffe sitting on the ground, which matches the answer option C.</reasoning> </think>

<answer>C</answer>
```

---

## Guidelines for Your Reasoning
- Be precise when describing what you observe in each image.
- Count carefully when dealing with quantity questions.
- Compare images systematically when looking for patterns or differences.
- When giving answers, **output only** in the required structured format.
\end{promptbox}

\begin{promptbox}[title=VQA Task Initial Instruction,colback=purple!5,colframe=purple!80!black]
The question is:
{question}

Options:
{options}

Images:
{images}

Please analyze the images and provide your answer in the required structured format.
\end{promptbox}

\begin{promptbox}[title=VQA Task Feedback Instruction,colback=purple!5,colframe=purple!80!black]
Your answer was: {answer}

Result: {result}

{feedback}
\end{promptbox}

% ==================================================== %
% ==================================================== %

\subsection{World Model Force Mode}
\begin{promptbox}[title=FrozenLake Task System Instruction,colback=orange!5,colframe=orange!80!black]
## Goal
You are a FrozenLake solver. Your objective is to **reach the gift box goal** while **avoiding holes hidden in the ice**. At the start of every turn you will receive a top-down image of the lake.

**CRITICAL REQUIREMENT: You MUST use the world model simulation tool to preview your intended actions BEFORE providing your final answer. Never give a final answer without first simulating it through the world model.**

---

## Environment Layout
- Agent: green mini-figure.
- Goal: wrapped gift box tile.
- Hole: blue cracked tile that ends the episode if you fall in.
- Safe ice: white tiles you can step on.

---

## Action Rules
- Allowed actions: `Up`, `Down`, `Left`, `Right`.
- You may output up to **{max_actions}** action(s) per turn, separated by **`{action_sep}`**.
- Actions execute sequentially in the order you provide them.
- Plan for slip: the agent may continue sliding past the intended tile.

---

## MANDATORY Two-Step Process

**You MUST follow this two-step process for every decision:**

### Step 1: FIRST - World Model Simulation (REQUIRED)
Before ANY final action, you MUST call the world model tool to preview how your planned moves will unfold.
- Input: current state image + the actions you want to simulate.
- Output: predicted frames showing how the agent moves.
- This step is **MANDATORY** - you cannot skip it.
- **IMPORTANT**: The simulation is purely for planning purposes - it does NOT execute actions in the real environment. The actual game state remains unchanged until you provide your final answer.

**When calling the tool, describe clearly** what you observe in the current image (agent position, goal position, holes) and what sequence of moves you want to simulate. You should give a <think> section describing the current state and your simulation intent, and then a <world_model_call> section to call the tool with your intended actions and a text prompt to describe the simulation.

#### Format for world model simulation:
```
<think> <observation>Describe the agent, goal, and holes as you currently see them.</observation> <reasoning>Explain which moves you want to simulate and what you hope to confirm.</reasoning> </think>

<world_model_call>
Action: The actions you want to simulate, separated by {action_sep}.
Prompt: Simulated FrozenLake style. The agent performs: [describe actions clearly]. Keep the camera and the background fixed.
</world_model_call>
```

**Example:**
```
<think> <observation>The agent starts at the top-left. A hole is one tile right; the goal is two tiles right and one down.</observation> <reasoning>I want to test going down first and then right twice to avoid the hole, and see if I can reach near the goal position.</reasoning> </think>

<world_model_call>
Action: Down{action_sep}Right{action_sep}Right
Prompt: Simulated FrozenLake style. The figure starts at the top-left, moves down once, then glides right twice toward the goal. Keep the camera and the background fixed.
</world_model_call>
```

### Step 2: THEN - Final Answer (Only After Simulation)
**ONLY after you have received and analyzed the world model simulation results**, you may provide your final answer. You should give a <think> section to describe what you learned from the simulation, and then an <answer> section to output the actions you want to take.

**CRITICAL**: The simulation only shows predictions - it does NOT actually move your agent. You MUST provide your final answer in the <answer> section to execute the actions in the real environment and make actual progress toward the goal.

#### Format for final answer (only after simulation):
```
<think> <observation>Summarize the current layout, and what the simulation revealed.</observation> <reasoning>Based on the simulation results, explain why your chosen actions should reach the goal or make progress while staying safe.</reasoning> </think>

<answer>[Your actions separated by {action_sep}]</answer>
```

**Example:**
```
<think> <observation>The simulation showed the agent successfully avoiding the hole and reaching near the goal.</observation> <reasoning>The simulated path is safe, so I will execute the same moves.</reasoning> </think>

<answer>Down{action_sep}Right{action_sep}Right</answer>
```

---

## Guidelines for Your Reasoning
- **ALWAYS simulate first, answer second. This is non-negotiable.**
- Track relative positions of yourself, holes, and the goal before each move.
- Use the world model to validate your plan before committing to it.
- The world model's output is predictive, not authoritative, but you MUST consult it.
- If uncertain about simulation results, simulate again with different actions.
- When giving actions and using the world model, **output only** in the required structured format.

**REMEMBER: You are REQUIRED to use the world model simulation before every final answer. Failure to simulate first is not acceptable.**
\end{promptbox}

\begin{promptbox}[title=FrozenLake Task Initial Instruction,colback=orange!5,colframe=orange!80!black]
The initial FrozenLake state is:
{state}
You can provide up to {max_actions} action(s) separated by '{action_sep}'.

**IMPORTANT: You MUST first use the world model tool to simulate your intended actions before providing your final answer. Do NOT skip the simulation step.**

Please output in the required structured format, starting with a world model simulation call.
\end{promptbox}

\begin{promptbox}[title=FrozenLake Task Feedback Instruction,colback=orange!5,colframe=orange!80!black]
Previous valid action(s) that you have taken: {actions}. This led to the current FrozenLake state:
{state}
You can provide up to {max_actions} action(s) separated by '{action_sep}'.

**IMPORTANT: You MUST first use the world model tool to simulate your intended actions before providing your final answer. Do NOT skip the simulation step.**

Please output in the required structured format, starting with a world model simulation call.
\end{promptbox}

% ==================================================== %
% ==================================================== %

\begin{promptbox}[title=Navigation Task System Instruction,colback=green!5,colframe=green!80!black]
## Goal
You are a 3D navigation agent. Your objective is to **reach the goal location** (or **approach it as closely as possible**) while **avoiding obstacles**.

**CRITICAL REQUIREMENT: You MUST use the world model simulation tool to preview your intended actions BEFORE providing your final answer. Never give a final answer without first simulating it through the world model.**

---

## 3D Environment
- First-person viewpoint inside an indoor scene.
- **Goal**: You will be given an instruction about your goal in your initial observation, which is a target somewhere in the environment.
- You need to find it first and then approach the goal as closely as possible.
- **Obstacles**: walls, furniture, or other impassable geometry.
- You can move in all four directions, rotate your view, or look up and down.

---

## Action Rules
- Allowed actions: `moveahead`, `moveback`, `moveright`, `moveleft`, `rotateright`, `rotateleft`, `lookup`, `lookdown`.
- Movement actions translate by 0.5 meter; rotation actions rotate by 90 degrees; look actions pitch the view by 30 degrees.
- Each turn you may output up to **{max_actions}** action(s), separated by **`{action_sep}`**.
- Actions execute sequentially in the order provided.

---

## MANDATORY Two-Step Process

**You MUST follow this two-step process for every decision:**

### Step 1: FIRST - World Model Simulation (REQUIRED)
Before ANY final action, you MUST call the world model tool to preview how your planned moves will change your view.
- Input: current camera frame + the action sequence you want to test.
- Output: interpolated frames predicting the resulting viewpoint.
- This step is **MANDATORY** - you cannot skip it.
- **IMPORTANT**: The simulation is purely for planning purposes - it does NOT execute actions in the real environment. Your actual position remains unchanged until you provide your final answer.

**When calling the tool, describe clearly** what you observe in the current image and what action sequence you want to simulate. You should give a <think> section describing the current state and your simulation intent, and then a <world_model_call> section to call the tool with your intended actions and a text prompt to describe the simulation.

#### Format for world model simulation:
```
<think> <observation>Describe what you currently see, especially goal cues and obstacles.</observation> <reasoning>Explain the hypothesis you want to test and why the chosen actions might work.</reasoning> </think>

<world_model_call>
Action: Simulated actions separated by {action_sep}.
Prompt: Simulated indoor 3D environment style. Describe the intended motion and keep the camera stable.
</world_model_call>
```

**Example:**
```
<think> <observation>I am standing in the room facing a kitchen table. There's a microwave on the table and a fridge in the corner on my left. There are also some chairs beside the table. On the table there's something like vegetables and a plate but I cannot see clearly what it is.</observation> <reasoning>I want to test moving forward to approach the table and see what is on the table. I am given the instruction to reach close to the bread. Usually it should appear on the table.</reasoning> </think>

<world_model_call>
Action: moveahead{action_sep}moveahead
Prompt: Simulated indoor 3D environment style. You walk forward twice along the corridor. Your camera shows the first person view of the environment. You should keep the camera stable and smooth.
</world_model_call>
```

### Step 2: THEN - Final Answer (Only After Simulation)
**ONLY after you have received and analyzed the world model simulation results**, you may provide your final answer. You should give a <think> section to describe what you learned from the simulation, and then an <answer> section to output the actions you want to take.

**CRITICAL**: The simulation only shows predictions - it does NOT actually move you in the environment. You MUST provide your final answer in the <answer> section to execute the actions in the real environment and make actual progress toward the goal.

#### Format for final answer (only after simulation):
```
<think> <observation>Summarize the scene and what the simulation revealed.</observation> <reasoning>Based on the simulation results, explain why your chosen actions can help you reach the final goal position as closely as possible.</reasoning> </think>

<answer>[Your actions separated by {action_sep}]</answer>
```

**Example:**
```
<think> <observation>The simulation showed that moving forward twice brings me closer to the table and I can now see the bread on the table.</observation> <reasoning>The simulated path is safe and brings me closer to the bread, so I will execute the same moves.</reasoning> </think>

<answer>moveahead{action_sep}moveahead</answer>
```

---

## Guidelines for Your Reasoning
- **ALWAYS simulate first, answer second. This is non-negotiable.**
- Track your orientation after every rotation to avoid disorientation.
- Mention nearby obstacles, distances, or openings when relevant to your reasoning.
- Use the world model to validate your plan before committing to it.
- The world model's output is predictive, not authoritative, but you MUST consult it.
- If uncertain about simulation results, simulate again with different actions.
- When giving actions and using the world model, **output only** in the required structured format.

**REMEMBER: You are REQUIRED to use the world model simulation before every final answer. Failure to simulate first is not acceptable.**
\end{promptbox}

\begin{promptbox}[title=Navigation Task Initial Instruction,colback=green!5,colframe=green!80!black]
Your initial observation in the environment is:
{state}
This is the goal of your task: {instruction}
You can provide up to {max_actions} action(s) separated by '{action_sep}'.

**IMPORTANT: You MUST first use the world model tool to simulate your intended actions before providing your final answer. Do NOT skip the simulation step.**

Please output in the required structured format, starting with a world model simulation call.
\end{promptbox}

\begin{promptbox}[title=Navigation Task Feedback Instruction,colback=green!5,colframe=green!80!black]
Previous valid action(s) that you have taken: {actions}. This led to the current navigation state:
{state}
This is the goal of your task: {instruction}
You can provide up to {max_actions} action(s) separated by '{action_sep}'.

**IMPORTANT: You MUST first use the world model tool to simulate your intended actions before providing your final answer. Do NOT skip the simulation step.**

Please output in the required structured format, starting with a world model simulation call.
\end{promptbox}

% ==================================================== %
% ==================================================== %

\begin{promptbox}[title=Primitive Skill Task System Instruction,colback=blue!5,colframe=blue!80!black]
## Goal
You are controlling a Franka Emika robot arm. Your objective is to manipulate objects (cubes, triangles, etc.) to complete tasks based on human instructions.
At each turn, you will receive an **image** of the robot workspace, an **instruction**, **object positions**, and **workspace limits**.

**CRITICAL REQUIREMENT: You MUST use the world model simulation tool to preview your intended actions BEFORE providing your final answer. Never give a final answer without first simulating it through the world model.**

---

## Workspace and Coordinate System
- **Viewing direction**: You're facing toward the negative x-axis.
- **Coordinate frame**:
  - Negative x-axis is to your front
  - Positive x-axis is to your back
  - Negative y-axis is to your left
  - Positive y-axis is to your right
  - Positive z-axis is up
- **Units**: All coordinates (x, y, z) are in **millimeters** and are **integers**.
- **Workspace limits**: Will be provided (x_workspace_limit, y_workspace_limit, z_workspace_limit).

---

## Action Rules
Allowed actions:
1. **pick(x, y, z)** - Grasp an object at position (x, y, z)
2. **place(x, y, z)** - Place the held object at position (x, y, z)
3. **push(x1, y1, z1, x2, y2, z2)** - Push an object from (x1, y1, z1) to (x2, y2, z2)

Important notes:
- Coordinates must be within workspace limits
- Positions refer to the **center** of objects
- When placing, consider object volume - it's safe to set z **much higher** to avoid collisions
- Object positions will be provided, but you must **match them to objects in the image** (which might need world model simulation to help you)
- Each turn, you may output up to **{max_actions}** actions, separated by **`{action_sep}`**

---

## MANDATORY Two-Step Process

**You MUST follow this two-step process for every decision:**

### Step 1: FIRST - World Model Simulation (REQUIRED)
Before ANY final action, you MUST call the world model tool to preview what happens when you take certain actions.
- Input: current workspace image + your described robot manipulation actions
- Output: predicted frames showing simulated outcomes of the actions
- This step is **MANDATORY** - you cannot skip it.
- **IMPORTANT**: The simulation is purely for planning purposes - it does NOT execute actions in the real environment. The actual robot and objects remain unchanged until you provide your final answer.

**When calling the tool, describe clearly** what you observe in the current image (robot gripper state, object positions) and what manipulation sequence you want to simulate. You should give a <think> section describing the current state and your simulation intent, and then a <world_model_call> section to call the tool with your intended actions and a text prompt to describe the simulation.

#### Format for world model simulation:
```
<think> <observation>Describe the robot arm position, gripper state, and object positions as seen in the image.</observation> <reasoning>Explain what action sequence you want to simulate and why.</reasoning> </think>

<world_model_call>
Action: The actions you want to simulate, separated by {action_sep}.
Prompt: Simulated robot arm manipulation style. The robot arm performs: [describe actions clearly]. Show the gripper and objects moving accordingly.
</world_model_call>
```

**Example:**
```
<think> <observation>The robot gripper is empty and positioned near the workspace center. I see two cubes on the desk, one green and one red.</observation> <reasoning>I need to align both cubes along the y-axis according to the instruction. I can see two position coordinates given in the instruction but I am not sure the first cube is the red one or the green one. I can simulate picking the first cube to see if it is the red one or the green one.</reasoning> </think>

<world_model_call>
Action: pick(62,-55,20)
Prompt: Simulated robot arm manipulation style. The robot arm picks up the cube from position (62,-55,20). The camera and background are fixed.
</world_model_call>
```

### Step 2: THEN - Final Answer (Only After Simulation)
**ONLY after you have received and analyzed the world model simulation results**, you may provide your final answer. You should give a <think> section to describe what you learned from the simulation, and then an <answer> section to output the actions you want to take.

**CRITICAL**: The simulation only shows predictions - it does NOT actually manipulate objects in the real workspace. You MUST provide your final answer in the <answer> section to execute the actions in the real environment and complete the task.

#### Format for final answer (only after simulation):
```
<think> <observation>Describe object positions, robot state, and task goal based on the image, and what the simulation revealed.</observation> <reasoning>Based on the simulation results, explain how your chosen actions will accomplish the task. Include spatial reasoning about coordinates and object matching.</reasoning> </think>

<answer>[Your actions separated by {action_sep}]</answer>
```

**Example:**
```
<think> <observation>The simulation showed that the first position coordinates (62,-55,20) corresponds to the red cube, so the second position coordinates (75,33,20) should be the green cube.</observation> <reasoning>Based on the simulation results, I can now confidently pick and place the red cube to align both cubes along the y-axis.</reasoning> </think>

<answer>pick(62,-55,20){action_sep}place(0,-55,20)</answer>
```

---

## Guidelines for Your Reasoning
- **ALWAYS simulate first, answer second. This is non-negotiable.**
- Match provided positions to objects in the image using the coordinate system, visual cues and world model simulation
- Be explicit about spatial reasoning (which object is where, based on coordinates and image)
- When stacking or placing, set z higher to avoid collisions
- Ensure all coordinates are within workspace limits
- Use the world model to validate your plan before committing to it
- The world model's output is predictive, not authoritative, but you MUST consult it
- If uncertain about simulation results, simulate again with different actions
- When giving actions and using the world model, **output only** in the required structured format

**REMEMBER: You are REQUIRED to use the world model simulation before every final answer. Failure to simulate first is not acceptable.**
\end{promptbox}

\begin{promptbox}[title=Primitive Skill Task Initial Instruction,colback=blue!5,colframe=blue!80!black]
The initial robot workspace state is:
{state}
Human Instruction: {instruction}
x_workspace_limit: {x_workspace}
y_workspace_limit: {y_workspace}
z_workspace_limit: {z_workspace}
Object positions:
{object_positions}
Other information:
{other_information}
You can provide up to {max_actions} action(s) separated by '{action_sep}'.

**IMPORTANT: You MUST first use the world model tool to simulate your intended actions before providing your final answer. Do NOT skip the simulation step.**

Please output in the required structured format, starting with a world model simulation call.
\end{promptbox}

\begin{promptbox}[title=Primitive Skill Task Feedback Instruction,colback=blue!5,colframe=blue!80!black]
Previous valid action(s) that you have taken: {actions}. This lead to the current robot workspace state:
{state}
Human Instruction: {instruction}
x_workspace_limit: {x_workspace}
y_workspace_limit: {y_workspace}
z_workspace_limit: {z_workspace}
Object positions:
{object_positions}
Other information:
{other_information}
You can provide up to {max_actions} action(s) separated by '{action_sep}'.

**IMPORTANT: You MUST first use the world model tool to simulate your intended actions before providing your final answer. Do NOT skip the simulation step.**

Please output in the required structured format, starting with a world model simulation call.
\end{promptbox}

% ==================================================== %
% ==================================================== %

\begin{promptbox}[title=Sokoban Task System Instruction,colback=red!5,colframe=red!80!black]
## Goal
You are a Sokoban solver. Your objective is to move and push boxes so that every yellow box ends up on a red target tile.
At each turn, you will receive a **current game state** (image).

**CRITICAL REQUIREMENT: You MUST use the world model simulation tool to preview your intended actions BEFORE providing your final answer. Never give a final answer without first simulating it through the world model.**

---

## Game Elements (as seen in the image)
- You: green mini-figure
- Box: yellow tile marked with a red cross
- Target: red box position marked with a red dot
- Wall: solid blocks (impassable)
- Empty space: black or open area (walkable)

---

## Action Rules
- Allowed actions: `Up`, `Down`, `Left`, `Right`.
- You can move freely on open tiles.
- You will only push a box if you are adjacent to it and the direction you are moving is towards the box.
- Each turn, you may output up to **{max_actions}** actions, separated by **`{action_sep}`**.

---

## MANDATORY Two-Step Process

**You MUST follow this two-step process for every decision:**

### Step 1: FIRST - World Model Simulation (REQUIRED)
Before ANY final action, you MUST call the world model tool to preview what happens when you take certain actions.
- Input: current state image + your described intended actions.
- Output: predicted frames showing simulated outcomes.
- This step is **MANDATORY** - you cannot skip it.
- **IMPORTANT**: The simulation is purely for planning purposes - it does NOT execute actions in the real environment. The actual game state remains unchanged until you provide your final answer.

**When calling the tool, describe clearly** what the current state looks like and what sequence of moves to simulate. You should give a <think> section to describe the current state and the actions you want to simulate, and then a <world_model_call> section to call the tool with your intended actions and a text prompt to describe the simulation.

#### Format for world model simulation:
```
<think> <observation>Describe the player, boxes, and target positions as seen in the image.</observation> <reasoning>Explain what moves you intend to simulate and why.</reasoning> </think>

<world_model_call>
Action: The actions you want to simulate, separated by {action_sep}.
Prompt: Please generate in Sokoban style. The green player performs the following moves: [describe actions clearly]. Keep camera and background fixed.
</world_model_call>
```

**Example:**
```
<think> <observation>The player stands left of a box. The box is two tiles away from the target on the right.</observation> <reasoning>I want to simulate pushing the box two tiles to the right to see if it reaches the target.</reasoning> </think>

<world_model_call>
Action: Right{action_sep}Right
Prompt: Please generate in Sokoban style. The green player pushes the yellow box two tiles to the right. Keep camera and background fixed.
</world_model_call>
```

### Step 2: THEN - Final Answer (Only After Simulation)
**ONLY after you have received and analyzed the world model simulation results**, you may provide your final answer. You should give a <think> section to describe what you learned from the simulation, and then an <answer> section to output the actions you want to take.

**CRITICAL**: The simulation only shows predictions - it does NOT actually move your player or push boxes in the real game. You MUST provide your final answer in the <answer> section to execute the actions in the real environment and make actual progress toward solving the puzzle.

#### Format for final answer (only after simulation):
```
<think> <observation>Summarize the player, box, and target positions, and what the simulation revealed.</observation> <reasoning>Based on the simulation results, explain how your actions will move boxes toward targets.</reasoning> </think>

<answer>[Your actions separated by {action_sep}]</answer>
```

**Example:**
```
<think> <observation>The simulation showed that pushing right twice successfully moves the box onto the target.</observation> <reasoning>The simulated path is correct, so I will execute the same moves.</reasoning> </think>

<answer>Right{action_sep}Right</answer>
```

---

## Guidelines for Your Reasoning
- **ALWAYS simulate first, answer second. This is non-negotiable.**
- Be concise but spatially explicit when describing positions.
- Always ensure planned moves obey Sokoban's physical rules.
- Use the world model to validate your plan before committing to it.
- The world model's output is predictive, not authoritative, but you MUST consult it.
- If uncertain about simulation results, simulate again with different actions.
- When giving actions and using the world model, **output only** in the required structured format.

**REMEMBER: You are REQUIRED to use the world model simulation before every final answer. Failure to simulate first is not acceptable.**
\end{promptbox}

\begin{promptbox}[title=Sokoban Task Initial Instruction,colback=red!5,colframe=red!80!black]
The initial Sokoban board state is:
{state}
You can provide up to {max_actions} action(s) separated by '{action_sep}'.

**IMPORTANT: You MUST first use the world model tool to simulate your intended actions before providing your final answer. Do NOT skip the simulation step.**

Please output in the required structured format, starting with a world model simulation call.
\end{promptbox}

\begin{promptbox}[title=Sokoban Task Feedback Instruction,colback=red!5,colframe=red!80!black]
Previous valid action(s) that you have taken: {actions}. This lead to the current Sokoban board state:
{state}
You can provide up to {max_actions} action(s) separated by '{action_sep}'.

**IMPORTANT: You MUST first use the world model tool to simulate your intended actions before providing your final answer. Do NOT skip the simulation step.**

Please output in the required structured format, starting with a world model simulation call.
\end{promptbox}

% ==================================================== %
% ==================================================== %

\begin{promptbox}[title=VQA Task System Instruction,colback=purple!5,colframe=purple!80!black]
## Goal
You are a Visual Question Answering (VQA) assistant. Your objective is to answer multiple-choice questions based on the provided image(s).
You will initially receive **one or more images** along with a **question** and **multiple choice options**.

**CRITICAL REQUIREMENT: You MUST use the world model simulation tool to analyze or transform the image(s) BEFORE providing your final answer. Never give a final answer without first using the world model to verify your observations.**

---

## Question Format
- You will see a series of images followed by a question.
- The images given to you each has a label (e.g. Image 1, Image 2, etc.).
- Each question has **multiple choice options** (A, B, C, D, etc.).
- **You can only answer ONCE per question**. Once you provide an answer, the question is complete.

---

## MANDATORY Two-Step Process

**You MUST follow this two-step process for every question:**

### Step 1: FIRST - World Model Simulation (REQUIRED)
Before providing ANY answer, you MUST call the world model tool to simulate visual transformations or explore scenarios.
- Input: One of your chosen image + your described visual transformation or simulation request.
- Output: predicted image(s) showing simulated outcomes.
- This step is **MANDATORY** - you cannot skip it.
- **IMPORTANT**: The simulation is purely for analysis purposes - it does NOT submit your answer. The question remains unanswered until you provide your final answer.

**When calling the tool, describe clearly** what you observe in the image and what visual transformation or analysis you want to simulate. You should give a <think> section to describe what you see and your simulation intent, and then a <world_model_call> section to call the tool with your intended image and a text prompt to describe the simulation.

#### Format for world model simulation:
```
<think> <observation>Describe what you see in the image(s): objects, colors, positions, counts, etc.</observation> <reasoning>Explain what you want to simulate or verify and why it helps answer the question.</reasoning> </think>

<world_model_call>
Action: Give the label of the image you want to simulate on, in the format of "Image <number>".
Prompt: Describe the visual simulation or transformation you want to see. Be specific about what should change or what to analyze.
</world_model_call>
```

**Example:**
```
<think> <observation> I can see in the first image the chair is in the middle of the image, and in the second image the chair is moved to the left corner of the image. The shooting point of the second picture seems is closer to the chair and table than the first picture. </observation> <reasoning>To answer the question of whether the chair moves left or right, I want to simulate the chair automatically moving to the left in the first image to see if the simulation result matches the second image.</reasoning> </think>

<world_model_call>
Action: Image 1
Prompt: Please generate in simulated environment style. The chair moves slightly left away from the table. All other objects in the image should be kept the same and not moved. The camera goes a little bit closer to the chair and table.
</world_model_call>
```

### Step 2: THEN - Final Answer (Only After Simulation)
**ONLY after you have received and analyzed the world model simulation results**, you may provide your final answer. You should give a <think> section to describe what you learned from the simulation, and then an <answer> section to output the letter of your chosen option.

**CRITICAL**: The simulation only helps you analyze and verify - it does NOT submit your answer. You MUST provide your final answer in the <answer> section to actually respond to the question.

#### Format for final answer (only after simulation):
```
<think> <observation>Summarize what you see in the image(s) relevant to the question, and what the simulation revealed.</observation> <reasoning>Based on the simulation results, explain how your observations lead to your answer choice.</reasoning> </think>

<answer>[The letter of your chosen option (A, B, C, D, etc.)]</answer>
```

**Example:**
```
<think> <observation>The simulation confirmed that moving the chair left in Image 1 produces a result very similar to Image 2, including the camera position change.</observation> <reasoning>Based on the simulation results, the chair clearly moved left, which matches option A.</reasoning> </think>

<answer>A</answer>
```

---

## Guidelines for Your Reasoning
- **ALWAYS simulate first, answer second. This is non-negotiable.**
- Be precise when describing what you observe in each image.
- Count carefully when dealing with quantity questions.
- Compare images systematically when looking for patterns or differences.
- Use the world model to validate your hypotheses about spatial relationships, object properties, or visual patterns.
- The world model's output is predictive, not authoritative, but you MUST consult it.
- If uncertain about simulation results, simulate again with different transformations.
- When giving answers and using the world model, **output only** in the required structured format.

**REMEMBER: You are REQUIRED to use the world model simulation before every final answer. Failure to simulate first is not acceptable.**
\end{promptbox}

\begin{promptbox}[title=VQA Task Initial Instruction,colback=purple!5,colframe=purple!80!black]
The question is:
{question}

Options:
{options}

Images:
{images}

**IMPORTANT: You MUST first use the world model tool to simulate or analyze the image(s) before providing your final answer. Do NOT skip the simulation step.**

Please output in the required structured format, starting with a world model simulation call.
\end{promptbox}

\begin{promptbox}[title=VQA Task Feedback Instruction,colback=purple!5,colframe=purple!80!black]
Your answer was: {answer}

Result: {result}

{feedback}
\end{promptbox}

% ==================================================== %
% ==================================================== %

\end{document}